%% file: arxiv.tex
\pdfoutput=1
\documentclass[]{fairmeta}

\input{math_commands.tex}

\usepackage{amsmath,amssymb}
\usepackage{enumitem}
\usepackage{capt-of}
\usepackage{float}
\usepackage{listings}
\usepackage{tabularx}
\usepackage{url}
\usepackage[normalem]{ulem}

\definecolor{draftnotecolor}{HTML}{0072B2}

\definecolor{imodraftnotecolor}{HTML}{D55E00}

\newcommand{\gitswarm}{\texttt{GitSwarm}}

\title{GitSwarm: Decentralized Compounding Inference}

\author[1,2,3]{Vedant Shah}
\author[1,4]{Ankur Samanta}
\author[1]{Paras Dahal}
\author[1]{Mikhail  Plekhanov}
\author[1]{Carole-Jean Wu}
\author[1]{Scott Yih}
\author[1]{Remi Munos}
\author[1]{Rob Fergus}
\author[1]{Jakob Foerster}
\author[5,*]{Ruslan Salakhutdinov}
\author[6,*]{Sanjeev Arora}
\author[1] {Jason Weston}
\author[2,3]{Aaron Courville}
\author[1]{Anirudh Goyal}

\affiliation[1]{Meta Superintelligence Labs}
\affiliation[2]{Mila - Qu\'{e}bec AI Institute}
\affiliation[3]{Universit\'{e} de Montr\'{e}al}
\affiliation[4]{Columbia University}
\affiliation[5]{Carnegie Mellon University}
\affiliation[6]{Princeton University}

\contribution[*]{Work done at Meta}

\abstract{\lipsum[1]}

\correspondence{Vedant Shah (\email{veds12@meta.com}), Anirudh Goyal (\email{agi@meta.com})
}

\abstract{
Long-horizon problem solving and scientific research require computation to accumulate across successive attempts. Partial solutions, experimental findings, and even unsuccessful approaches can become useful to later work, yet most inference-time computation is organized around individual trajectories or candidates rather than a persistent body of reusable work. We call this paradigm \emph{compounding inference}: organizing inference-time computation so that intermediate work persists and can be inspected, extended, combined, or challenged by subsequent computation. We instantiate compounding inference in \gitswarm{}, an asynchronous system in which homogeneous agents independently decide how to advance a task while collaborating through structured persistent memory. Agents explore, experiment, verify, refine, and synthesize previous work in a shared, branchable Git repository. Atomic commits preserve intermediate artifacts, while explicit semantic dependencies record how later contributions build on work across branches. We evaluate \gitswarm{} on long-horizon problem solving and sustained GPU-backed experimental research. On IMOProofBench-Advanced, \gitswarm{} solves all 30 problems in one run using GPT-5.5. On ProgramBench, it achieves a 79.4\% mean score, compared with 65.1\% for the strongest reported baseline under the stated budget. Across three neural architecture research tasks: Residual Matrix Transformer, Looped Transformer, and NanoChat, \gitswarm{} improves upon the respective starting architectures through successive experimentation. Beyond final performance, we measure whether computation actually accumulates: on ProgramBench, 94.7\% of contributions are subsequently built upon, while the selected solution's transitive ancestry covers 82--93\% of the contribution graph. Together, these results show that inference-time computation can accumulate across otherwise independent episodes, forming an evolving body of work that subsequent inference can reuse.

}

\begin{document}
\raggedbottom
\maketitle

\begin{figure}[h!]
    \centering
    \includegraphics[width=\linewidth]
        {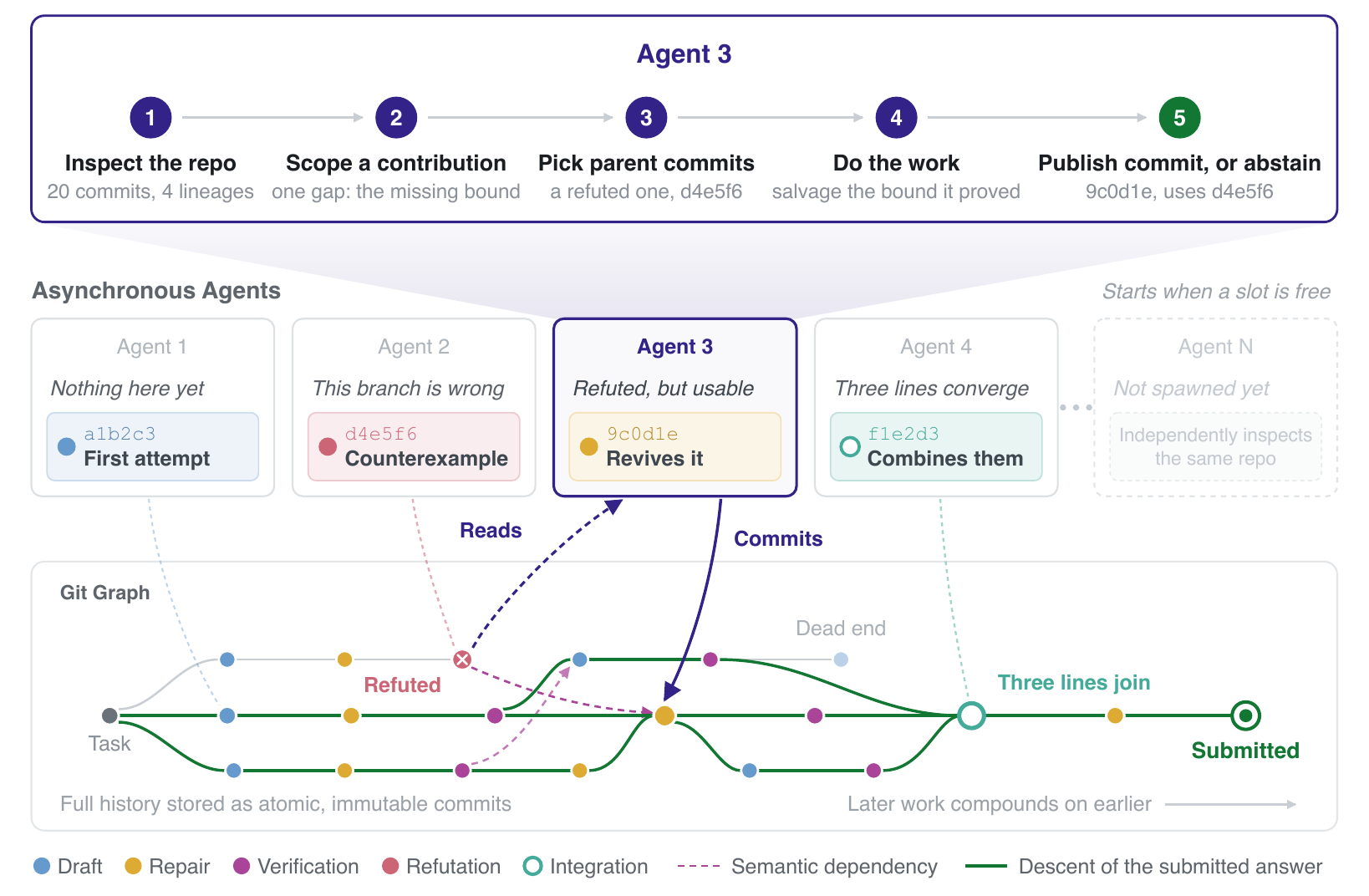}
    \caption{
        \textbf{Overview of \gitswarm{} (Schematic).}
        Homogeneous workers independently inspect,
        extend, and contribute to a shared,
        branchable repository. Contributions may
        include exploration, verification, repair,
        experimentation, or synthesis. Atomic
        commits preserve intermediate artifacts,
        while semantic dependencies link
        work across branches. The
        highlighted ancestry illustrates how
        successive contributions can accumulate
        into a final solution.
    }
    \label{fig:gitswarm-overview}
\end{figure}

\section{Introduction}
\label{sec:introduction}

Scaling inference-time computation improves language
model performance through repeated sampling, longer
reasoning trajectories, search, and multi-agent
collaboration~\citep{bestofn,testtimescaling,multiagent}.
These approaches largely ask how additional
computation should be spent. Long-horizon problem
solving and scientific research pose a complementary
challenge: useful computation must also accumulate
across attempts.

The value of intermediate work is often unknown
when it is produced. An unsuccessful experiment may
reveal a promising direction; a partial solution may
establish an ingredient used much later; competing
approaches may contain ideas that become useful only
when combined. If such work disappears when an
episode terminates, later computation may need to
rediscover it. Preserving it instead allows subsequent
agents to build on an accumulated body of prior work.
We call this paradigm \emph{compounding inference}:
organizing inference-time computation into a
persistent, evolving body of reusable work.

Existing approaches support different parts of this
process. Independent sampling explores diverse
candidates, longer reasoning trajectories retain work
within an episode, and multi-agent systems enable
collaboration, often through predefined roles,
centralized planning, or shared conversations
\citep{hong2024metagpt,chen2024agentverse,
wang2025moa,wu2023autogen}.
Other approaches distill earlier attempts into
summaries that guide subsequent inference
\citep{rw_pdr2025,venkatraman2025rsa,rw_agentic_scaling2026}.
We investigate a different organization: independent
workers preserve competing lines of work and their
dependencies, while deciding for themselves which
previous contributions to extend.

We instantiate compounding inference in \gitswarm{},
which uses a shared, branchable Git repository as
structured persistent memory. Homogeneous workers
operate asynchronously, inspect previous
contributions, and independently decide whether to
explore, experiment, verify, refine, or combine
existing work. Each contribution is published as an
atomic commit. Importantly, \gitswarm{} distinguishes
the repository state from which a worker begins from
the additional prior contributions that its work
depends on. Explicit semantic dependencies therefore
allow a worker to extend one branch while drawing on
discoveries from others. This preserves competing
approaches without losing the connections between
them. No central intellectual planner assigns tasks
or roles; an execution harness manages scheduling
and validation, while workers nominate candidates
for final selection.

Our evaluation spans two settings. The first is
\emph{long-horizon problem solving}, where tasks
have well-defined objectives and verifiable outcomes.
We study mathematical proof and program
reconstruction. On IMOProofBench-Advanced
\citep{luong2025robust}, \gitswarm{} solves all
30 problems in one run using Codex (GPT-5.5-High)
workers and slightly under one million output tokens.
On ProgramBench~\citep{yang2026programbench}, with Codex (GPT-5.5-High) workers, 
it achieves a 79.39\% mean score, as compared to
65.12\% for Forced Codex - a single Codex (GPT-5.5-High) agent
repeatedly prompted to continue, at a comparable
compute budget. Mean performance improves with
inference-time computation over the tested ranges.

The second setting is \emph{sustained experimental
research}, where agents must decide which experiments
to run and learn from their outcomes. We evaluate
\gitswarm{} on Residual Matrix
Transformer~\citep{mak2025rmt}, Looped
Transformer~\citep{geiping2025recurrent},
and NanoChat. \gitswarm{} improves upon the
respective starting points in all three tasks.
On Residual Matrix Transformer, two configurations
achieve held-out losses of 2.890 and 2.894,
compared with 3.147 for the reference, using
the same 1.3B-token training budget.
The resulting research histories show workers
combining architectural changes developed on
separate branches and revisiting initially
unsuccessful experiments.

Beyond final performance, we examine whether
earlier computation actually accumulates.
Across the analyzed ProgramBench runs, 94.7\%
of contributions are subsequently referenced,
and the selected solution's transitive ancestry
covers 81.6--92.6\% of the contribution graph
on average across budgets, largely through
declared cross-branch dependencies.
Execution traces provide separate evidence that
workers consume earlier artifacts.
Mathematical proof exhibits a different pattern:
workers frequently reuse previous work to verify
and compare candidates even when that work does
not enter the selected proof's ancestry.
Together, these results show that persistent
memory can support both the synthesis of earlier
contributions and the accumulation of evidence
that guides subsequent computation.

Our contributions are therefore threefold:
we formulate \emph{compounding inference} as a
paradigm for accumulating reusable inference-time
computation across episodes; we introduce
\gitswarm{} as a concrete realization based on
persistent, branchable contributions with explicit
cross-branch dependencies; and we evaluate both
task performance and the accumulation process
itself across problem solving and sustained
experimental research.

\section{\texorpdfstring{\gitswarm{}}{GitSwarm}}
\label{sec:method}

\gitswarm{} implements compounding inference through
asynchronous, homogeneous workers collaborating over
a shared Git repository. Workers independently decide
how to advance the task, while an execution harness
manages concurrency, isolation, validation, and
publication. The repository preserves intermediate
work as a branchable computation graph.

\subsection{Design Principles}
\label{sec:desiderata}

We seek to accumulate computation without prescribing
how agents should organize their work. This requires
three properties of the shared memory:
\emph{persistence}, so intermediate discoveries
survive their authors; \emph{branchability}, so
competing approaches can develop without prematurely
committing to a single solution; and
\emph{attribution}, so subsequent workers can
identify dependencies between contributions,
including across branches. These properties support \emph{adaptive, decentralized
control}: workers independently choose what to
explore, verify, refine, or combine based on the
evolving shared state. The harness coordinates
execution but does not assign intellectual roles
or prescribe a sequence of problem-solving stages.

\subsection{Shared Computation Graph}
\label{sec:method-formulation}
\label{sec:provenance}

Consider a task $q$, a fixed language model
$p_\theta$, and a budget of $B$ charged worker
episodes. For each task, \gitswarm{} initializes
a Git repository containing the task specification
and required artifacts, committed as an immutable
root $v_0$. Accepted contributions form an evolving graph
$G_t=(V_t,E_t)$, where
$V_t=\{v_0,\ldots,v_t\}$ contains the root and
the first $t$ published contributions. Each
contribution $v_i$ is an atomic Git commit
with a physical parent $p_i\in V_{i-1}$.
This parent identifies the state from which
the worker began, but need not capture all
the information used in its contribution.

To record additional dependencies, each worker
declares a set of semantic sources
$U_i\subseteq V_{i-1}$ in a mandatory
\texttt{informed\_by=} field. The physical parent
must be included in $U_i$. The graph contains
both Git-parent edges and additional
semantic-dependency edges:
\begin{equation}
\begin{aligned}
E_t ={}&
\{(p_i,v_i):1\leq i\leq t\}\\
&{}\cup
\{(u,v_i):
u\in U_i\setminus\{p_i\},
\ 1\leq i\leq t\}.
\end{aligned}
\end{equation}

Edges point from an earlier contribution to
one that depends on it. This representation
allows a worker to inherit one branch while
using discoveries from others, without
requiring  merges. Semantic dependencies
are declared by workers, not
independently established as causal influence.

\subsection{Asynchronous Worker Protocol}
\label{sec:worker-execution}
\label{sec:async-coordination}

A run has a budget of $B$ charged worker episodes
and maximum concurrency $C$. The harness maintains
up to $C$ active workers, launching replacements
as slots become available. All workers use the
same model and behavioral contract, without
predefined roles.

Each worker operates in an isolated Git worktree.
It inspects the repository, selects a starting
commit, and independently decides how to advance
the task. Each episode ends with one of three actions:
\begin{enumerate}[leftmargin=*,nosep]
    \item \textbf{Contribute:} Publish one validated,
    atomic commit containing new work, such as a
    partial solution, experiment, test, or refinement.
    Validation enforces task-specific publication
    requirements, not solution correctness.

    \item \textbf{Nominate:} Select an existing
    candidate and publish a separate nomination
    record with a rationale, without modifying
    the candidate.

    \item \textbf{Abstain:} Terminate without
    publication when no useful additional work
    is identified or equivalent work is already
    underway.
\end{enumerate}

Voluntary abstentions trigger replacement without
consuming $B$, although they incur inference costs;
other episode outcomes consume the budget. Workers
may therefore allocate effort adaptively among
exploration, refinement, experimentation,
verification, integration, and selection.

\subsection{Live Swarm State: Dashboard and In-Flight Activity}
\label{sec:swarm-state}

\begin{figure}[t]
    \centering
    \includegraphics[width=\linewidth]{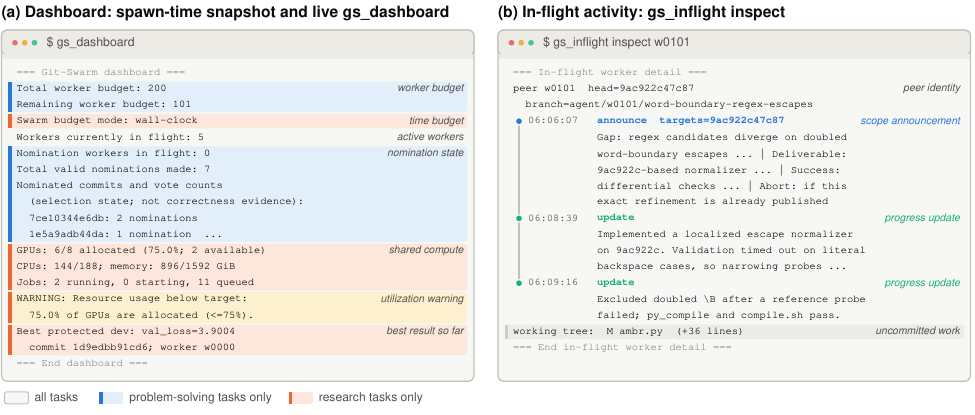}
    \caption{\textbf{Live swarm state.} (a) A dashboard
    combining fields from a ProgramBench worker
    ($B=200$) and a Residual Matrix Transformer worker;
    shading marks fields shown only for one task type.
    (b) A ProgramBench worker inspecting a peer's
    activity log. Text is taken from worker traces and
    abridged.}
    \label{fig:swarm-state}
\end{figure}

Workers also see the current state of the run and
what their peers are doing, through two interfaces
that report state but do not assign work
(Figure~\ref{fig:swarm-state}).

\noindent\textbf{Dashboard.}
Every worker receives a dashboard snapshot at
spawn time and can query the live state with
\texttt{gs\_dashboard}. It reports the remaining
worker budget, the workers in flight, and the
nomination tally, noting that votes are not
evidence of correctness. For research tasks, which
use a wall-clock budget and select the final model
by validation score, the budget and nomination
fields are replaced by shared GPU, CPU, and job
usage, a warning when GPU utilization
is low, and the best validation score 
with its commit.

\noindent\textbf{In-flight activity.}
Since uncommitted work is invisible in the
repository, each worker keeps a public activity log
via \texttt{gs\_inflight}. It announces its scope
(gap, deliverable, success and abort criteria)
before editing and posts progress updates. Peers can
list active workers or inspect a worker's log and
uncommitted changes, but cannot modify them. This is meant to help workers avoid taking up work which is already being undertaken by their in-flight peers.

\subsection{Decentralized Readout}
\label{sec:async-readout}

While the agents work asynchronously, eventually a consensus needs to form on which exact commit to submit. We keep this process distributed across workers rather than delegating to privileged final judge, referring to the process as \textit{decentralized readout}. Let $c_i$ be the candidate identified
in nomination record $i$. After the
worker budget is exhausted, the
harness selects the candidate with
the most nominations:
\begin{equation}
v^\star =
\arg\max_{v\in V_{\mathrm{cand}}}
\sum_i \mathbb{I}[c_i=v],
\end{equation}
with deterministic tie-breaking. Nomination records remain separate
from the computation graph. Workers
can therefore evaluate existing
candidates while others continue
constructing new ones, without
changing the ancestry of the
eventually selected solution.

\textbf{Summary.} Together, persistent commits, cross-branch reuse,
and decentralized worker decisions allow computation
to accumulate across otherwise independent episodes.
We next evaluate this approach on long-horizon problem
solving and sustained experimental research, and
examine how intermediate work is reused.

\section{Experiments}
\label{sec:experiments}

We evaluate \gitswarm{} in two settings: long-horizon problem solving and
sustained experimental research. We first measure benchmark performance and
inference-time scaling on mathematical proof and program reconstruction. We then
study three GPU-backed architecture research tasks. Finally, we examine reuse,
transitive accumulation, and adaptive coordination. Controlled synthetic tests
(Appendix~\ref{app:synthetic-task-details}) and architectural comparisons probe
individual system properties.

\subsection{Experimental Setup}
\label{sec:experimental-setup}

\noindent\textbf{Problem-solving tasks.}
IMOProofBench-Advanced~\citep{luong2025robust}
contains 30 mathematical proof problems.
ProgramBench~\citep{yang2026programbench}
asks agents to reconstruct programs from
documentation and an execute-only reference.
We evaluate all 30 proof problems and a
fixed subset of 50 ProgramBench tasks,
stratified by language, difficulty, test-suite
size, and runtime
(Appendix~\ref{app:programbench-subsets}).
We use nine separate programming tasks
and six proof problems during development;
the latter are included in the full proof
evaluation. Scoring rules and task selection
are detailed in Appendix~\ref{app:setup-benchmarks}.

\noindent\textbf{Models and baselines.}
For IMOProofBench-Adavnced, we use \gitswarm{} with Gemini-3.1-Pro and
GPT-5.5-high workers. We compare against
single agents using bash, mini-SWE-agent~\citep{lieret2025minisweagent}, or Codex~\citep{openai2025codex} (as appropriate to the model),  repeatedly prompted to continue
after stopping.
For ProgramBench, we use Codex (GPT-5.5-high)
workers and GPT-5.5 workers at the default
reasoning effort, and compare against a
repeatedly continued Codex (GPT-5.5-high) agent. We always use a worker concurrency of 5 for \gitswarm{} in the case of problem-solving tasks. 
We call these continued single-agent
baselines \emph{Forced} and evaluate
their intermediate natural stopping checkpoints.

\noindent\textbf{Compute accounting.}
\gitswarm{}'s budget $B$ counts charged worker
episodes. Voluntary abstentions are replaced
without consuming the budget, although they
still incur inference costs. Programming
runs use budgets of 100, 200, 300, and
500 episodes; proof runs use 20, 40, 80,
120, and 160. These measurements account for differences in
the amount of work performed within an
episode.

We report performance scaling curves with respect to input and output token counts in the case of problem-solving tasks. We count tokens per model request and sum them over all requests made by a worker or agent: input tokens count the full prompt sent with each request, including all previously accumulated context. Output tokens count only the tokens newly generated in that request, including hidden reasoning tokens. Reported values are per-problem totals averaged across problems. record input and output tokens. The research experiments use
separate GPU and time budgets, introduced
alongside each task. Full prompts and
compute protocols appear in the appendix.

\subsection{Benchmark Performance and Inference-Time Scaling}
\label{sec:performance}

We first test whether additional inference-time
compute improves performance on tasks with
well-defined answers. We evaluate program
reconstruction and mathematical proof, measuring
both final performance and how it changes as
workers receive larger budgets.

\begin{figure}[t]
    \centering
    \includegraphics[width=\linewidth]
        {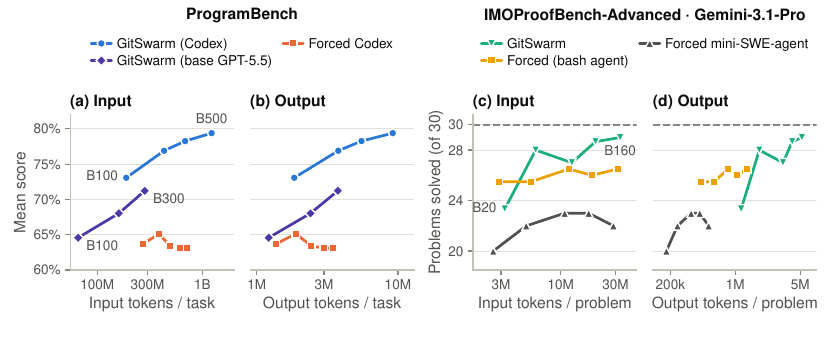}
    \caption{
        \textbf{Inference-time scaling.}
        Performance against input and output
        tokens on ProgramBench (a, b) and
        IMOProofBench-Advanced with
        Gemini-3.1-Pro (c, d). Each \gitswarm{}
        point represents one run at the indicated worker budget.
    }
    \label{fig:programbench-scaling}
\end{figure}

\noindent\textbf{ProgramBench.}
Figure~\ref{fig:programbench-scaling} examines
scaling on the 50 tasks shared by all plotted
runs. With Codex workers, \gitswarm{} improves
from 73.1\% with 100 workers to 79.4\%
with 500. With GPT-5.5  workers,
it improves from 64.6\% with 100 workers
to 71.2\% with 300. In comparison, the
sequentially continued Codex baseline
remains between 63.1\% and 65.1\%
across its plotted checkpoints. \gitswarm{}'s
gains diminish at larger budgets, while
input-token costs grow as workers inspect
an expanding repository.

\noindent\textbf{IMOProofBench-Advanced.}
We next test whether scaling also helps on
mathematical proof problems. With Gemini-3.1-Pro, the mean number
of fully correct proofs increases from
23.3 at a budget of 20 workers to
29.0 at 160. The curve is not monotonic:
performance decreases locally at a
budget of 80. The forced
mini-SWE-agent baseline reaches at
most 23 fully correct proofs on
its plotted curve. The continued bash agent baseline solved 26.5 problems at similar input token budget. With GPT-5.5-high
workers, \gitswarm{} solves 28 of 30
problems at a budget of 20 workers
and all 30 at budgets of 40 and 80. Note that Best-of-N baselines (with oracle verifier) also reach the benchmark ceiling at lower token consumptions,
so saturating this benchmark does not establish a unique advantage for GitSwarmFull proof results and
compute breakdowns appear in
Appendix~\ref{app:benchmark-results}.

These benchmarks show how performance
changes with additional inference-time
compute. Both, however, have fixed targets
and established evaluation procedures.
We next consider open-ended research,
where workers must also decide which
experiments to run and how to build
on their results.

\subsection{Sustained Experimental Research}
\label{sec:autoresearch}

We next move from fixed-answer problems to open-ended
research, where workers must decide what to investigate,
run experiments, and interpret their results. We study
three neural architecture tasks: Residual Matrix
Transformer (RMT)~\citep{mak2025rmt}, Looped Transformer~\citep{geiping2025recurrent}, and NanoChat~\citep{nanochat}.
Workers share code and experimental findings through
the repository and submit GPU jobs to a shared resource allocation through SLURM.
Each task has its own research objective and compute
budget. We strip \gitswarm{} of the nomination based decentralized readout for these experiments since all tasks discussed here support a validation performance based candidate selection (Appendix~\ref{app:autoresearch-setup}).

\noindent\textbf{Residual Matrix Transformer.}
Starting from RMT~\citep{mak2025rmt}, workers seek
architectural improvements while preserving its
matrix-valued residual state and remaining below
350M parameters. We run two 12 Codex (GPT-5.6-sol; xHigh effort) worker swarms with for 72 hours on an 8 GPU H200
node, allocating either eight or four GPUs collectively to the swarm for running experiments.

\begin{figure}[t]
    \centering
    \begin{minipage}[c]{0.327\linewidth}
        \centering
        \includegraphics[
            width=\linewidth,
            height=2in,
            keepaspectratio
        ]{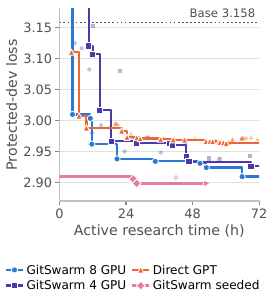}
        \par\scriptsize (a) Development loss
    \end{minipage}\hfill
    \begin{minipage}[c]{0.645\linewidth}
        \centering
        \includegraphics[
            width=\linewidth,
            height=2in,
            keepaspectratio
        ]{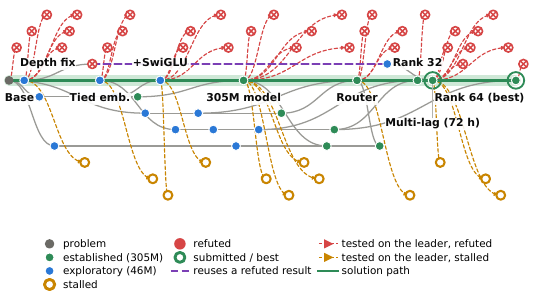}
        \par\scriptsize (b) Research dependencies
    \end{minipage}
    \caption{
        \textbf{RMT research.}
        (a) Best-so-far development loss over the
        72-hour research budget; the reference
        loss is 3.158.
        (b) The 8-GPU swarm's research graph,
        showing competing branches, integration
        into the final architecture, and recovery
        of an earlier failed idea.
    }
    \label{fig:autoresearch_hillclimb}
\end{figure}

The two swarms achieve held-out losses of 2.890
and 2.894, compared with 3.147 for the reference.
All three models train on 1.3B tokens. Forced GPT-5.6-sol (xHigh effort) 
reaches 2.927 with 3.0B training tokens, while
Codex reaches 2.690 with 10.1B. Thus, both swarms
improve the reference architecture at its original
training-token budget. Figure~\ref{fig:autoresearch_hillclimb} shows how
the 8-GPU swarm arrived at its final architecture.
Four of its six changes originated on other branches. Notably, one of them began as a failed attempt to use the previous
token's memory. Another worker revisited the result
49.5 hours later, extended the idea to multiple
preceding tokens, and developed a temporal mixer
that entered the final model. The best architecture's declared
dependencies span 24 commits by 24 workers. The changes made by codex on the other hand involved simply increasing the width of the attention layer and scaling the training budget. Thus, \gitswarm{} came up with more interesting architecture modifications.

The repository also supports reuse across completed
searches. In a separate run, we seeded \gitswarm{} with the winning commits from three prior independent \gitswarm{} runs. Workers combined their discoveries and
reduced the inherited best development loss from
2.909 to 2.898 in approximately 28 hours. This run
inherited prior research, 
so its progress is not directly comparable to an
independent run. Full trajectories,  experimental
histories and a more detailed discussion of the findings appear in Appendix~\ref{app:rmt}.

\noindent\textbf{Looped Transformer.}
We next study Huginn~\citep{geiping2025recurrent},
a 308M-parameter transformer with weight-tied
recurrence. Workers must preserve recurrence and
remain below 350M unique parameters. A 12-worker
swarm uses Codex (GPT-5.5; High effort) workers with a 48-hour budget
on one 8 GPU H200 node.

After 26 hours, the swarm reaches a protected-development loss of 3.187, compared with
3.265 for the reference, 3.274 for direct GPT,
and 3.479 for Codex. The winning architecture
changes the recurrent input adapter from a linear layer to a 
gated feed-forward network. A parameter-matched
control reaches 3.212, retaining
two thirds of the improvement. Its research history tells a different story
from RMT. The winning adapter was proposed early
but rejected after evaluation on 128 examples subset of the full development set.
Other workers re-evaluated it on 512 examples
and confirmed the improvement on the full
development set. Subsequent analysis found that
the smaller evaluation misranked five of twelve
model pairs, compared with one at 512 examples.

\begin{figure}[t]
    \centering
    \begin{minipage}[c]{0.327\linewidth}
        \centering
        \includegraphics[
            width=\linewidth,
            height=2in,
            keepaspectratio
        ]{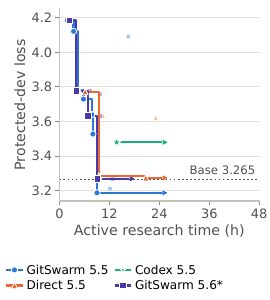}
        \par\scriptsize (a) Development loss
    \end{minipage}\hfill
    \begin{minipage}[c]{0.645\linewidth}
        \centering
        \includegraphics[
            width=\linewidth,
            height=2in,
            keepaspectratio
        ]{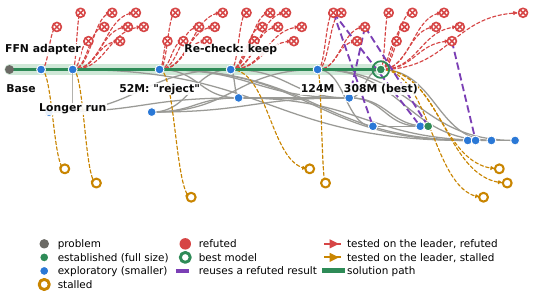}
        \par\scriptsize (b) Research dependencies
    \end{minipage}
    \caption{
        \textbf{Looped Transformer research.}
        (a) Best-so-far development loss at
        eight recurrences. Arrows mark ongoing
        runs; the two-node GPT-5.6 run is
        not compute-matched.
        (b) The winning adapter's development
        path and its wider evidence dependencies,
        including reversal of an early rejection
        and reuse of negative results.
    }
    \label{fig:autoresearch_loop}
\end{figure}

As Figure~\ref{fig:autoresearch_loop} shows,
the winning code follows a relatively short
development path, but its declared dependencies
span 89 commits. Much of this
additional history consists of experimental
evidence rather than transferred code. During
the first 26 hours, workers conducted 395
architectural experiments across twelve idea
families; 74\% of negative architectural
results were subsequently cited. Full controls
and research histories appear in
Appendix~\ref{app:loop}.

\noindent\textbf{NanoChat.}
Finally, we study joint architecture and
training-system improvement on NanoChat. Workers
start from a 50.3M-parameter language model and
may change its architecture, optimizer, and
training implementation. The data, tokenizer,
and evaluator remain fixed. Each candidate is
trained from scratch under a five-minute GPU
training clock and evaluated using validation
bits per byte (BPB; lower is better). We compare \gitswarm{} with single-agent Codex
and an Atlas-style evolutionary-search
baseline~\citep{hambardzumyan2026aira2},
using 16 GPUs per run over the first 24 active
research hours on H200 or GB300 hardware.

In the GPT-6-Astra + GB300 run, a 20-worker swarm
reduces validation BPB from 0.993 to 0.905
through 1,465 successful evaluations.
Its incumbent combines several changes
developed on separate branches, including
shared hashed bigram and trigram embeddings.
Unlike the preceding tasks, NanoChat does not
constrain model capacity: the incumbent grows
to approximately 1.72B parameters, predominantly
in its lexical table. The improvement therefore
reflects joint search over architecture,
capacity, optimization, and training systems.

\begin{figure}[t]
    \centering
    \begin{minipage}[c]{0.43\linewidth}
        \centering
        \includegraphics[
            width=\linewidth,
            height=2in,
            keepaspectratio
        ]{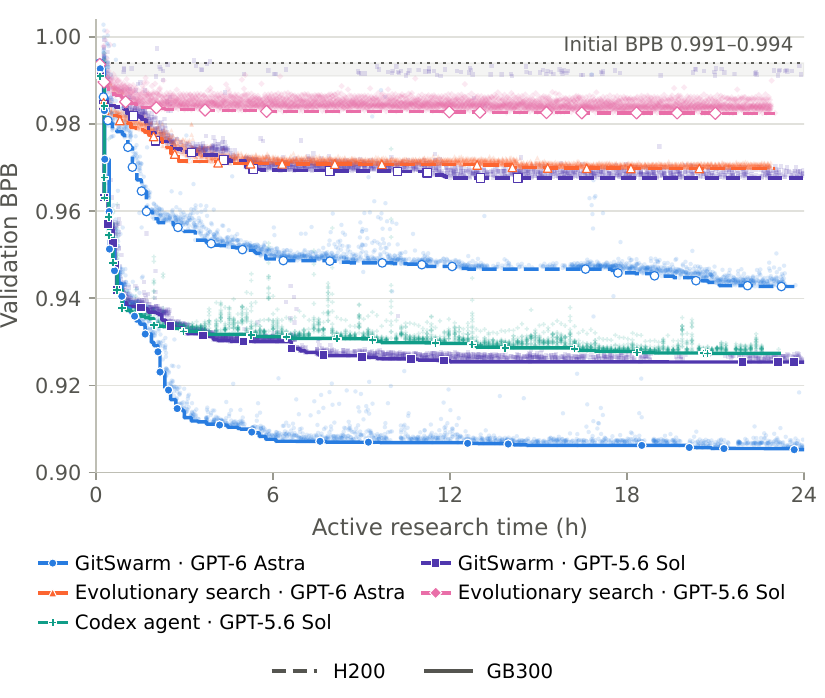}
        \par\scriptsize (a) Validation BPB
    \end{minipage}\hfill
    \begin{minipage}[c]{0.545\linewidth}
        \centering
        \includegraphics[
            width=\linewidth,
            height=2in,
            keepaspectratio
        ]{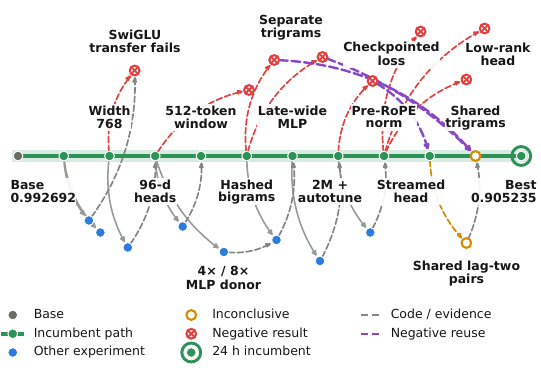}
        \par\scriptsize (b) Research dependencies
    \end{minipage}
    \caption{
        \textbf{NanoChat research over the first
        24 active hours.}
        (a) Validation BPB across methods and
        hardware configurations.
        (b) Selected dependencies in the
        Astra/GB300 swarm. Green marks incumbent
        ancestry, gray dashed edges mark code
        or evidence reuse, and purple marks
        reuse of negative results. Crossed red
        nodes are negative results; hollow amber
        nodes are inconclusive. The graph
        contracts omitted ancestry and is
        not a time axis.
    }
    \label{fig:autoresearch_nanochat}
\end{figure}

The research graph in
Figure~\ref{fig:autoresearch_nanochat} shows
how an early failure also informs the final
design. Separate trigram tables initially
perform poorly. Approximately 21 active hours
later, another worker reuses their hashing
scheme in a shared-table design. A subsequent
worker enlarges the table, producing the
24-hour incumbent. More broadly, 70.1\% of
negative-labelled research commits are cited
by later accepted contributions. Complete
source comparisons and protocol details appear
in Appendix~\ref{app:nanochat}.

These histories reveal distinct forms of
compounding research. RMT integrates architectural
changes from separate branches. Looped Transformer
accumulates evidence that overturns an early
experimental verdict. NanoChat explores a broader
space of model and training changes, including
ideas recovered from unsuccessful experiments.
We next examine how frequently workers reuse
earlier contributions across the benchmark runs.

\subsection{Evidence for Compounding Computation}
\label{sec:worker-reuse}
\label{sec:transitive-accumulation}

The research histories show how earlier discoveries
can shape later work. We now examine how frequently
this happens across the benchmark runs. We measure
three aspects of reuse: declared dependencies,
artifact consumption observed in execution traces,
and ancestry of selected solutions. We conduct this analysis on the ProgramBench and IMOProofBench-Advanced evaluation runs reported in Section~\ref{sec:performance}.

\noindent\textbf{Declared reuse.}
Workers cite earlier contributions in nearly every
eligible episode: 99.9\% on ProgramBench and
96.2\% on IMOProofBench-Advanced. Later workers
build upon 94.7\% and 44.4\% of published
contributions, respectively, by naming them
as Git parents or \texttt{informed\_by=} sources.
These measurements pool 200 ProgramBench runs
and 150 proof runs across the evaluated budgets.

\begin{figure}[b!]
    \centering
    \begin{minipage}[c]{0.55\linewidth}
        \centering
        \includegraphics[width=0.8\linewidth]{
            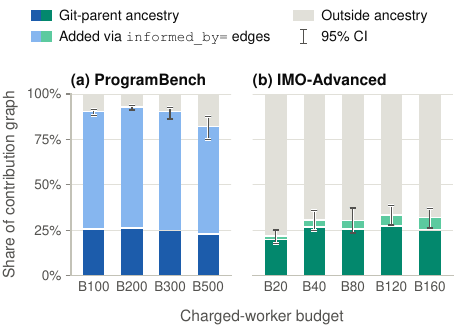
        }
    \end{minipage}\hfill
    \begin{minipage}[c]{0.4\linewidth}
        \captionof{figure}{
            \textbf{Selected-solution ancestry.}
            Share of contributions reached through
    Git-parent ancestry, additional
    \texttt{informed\_by=} dependencies, or neither,
    across worker budgets. ProgramBench
    solutions incorporate much of the
    contribution graph, primarily through
    cross-branch dependencies. In contrast,
    selected IMOProofBench proofs have
    shallower ancestry: later work often
    evaluates and compares existing candidates
    rather than extending the selected proof.
        }
        \label{fig:winner-ancestry-coverage}
    \end{minipage}
\end{figure}

\begin{figure}[t]
    \centering
    \includegraphics[width=0.8\linewidth]{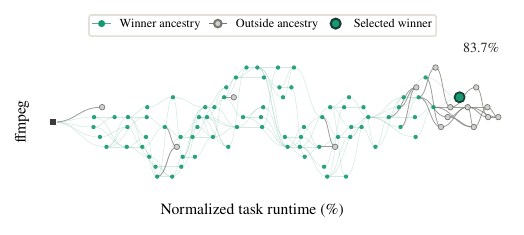}
    \caption{\textbf{Selected-solution ancestry in a ProgramBench run
    (FFmpeg B100).} Commits over normalized task runtime. The selected
    winner's transitive ancestry (green), reached through Git parents and
    \texttt{informed\_by=} edges, covers 83.7\% of the contribution graph;
    only a few commits, mostly late in the run, lie outside it (gray).}
    \label{fig:ffmpeg-ancestry}
\end{figure}

\begin{figure}[b!]
    \centering
    \includegraphics[width=\linewidth]{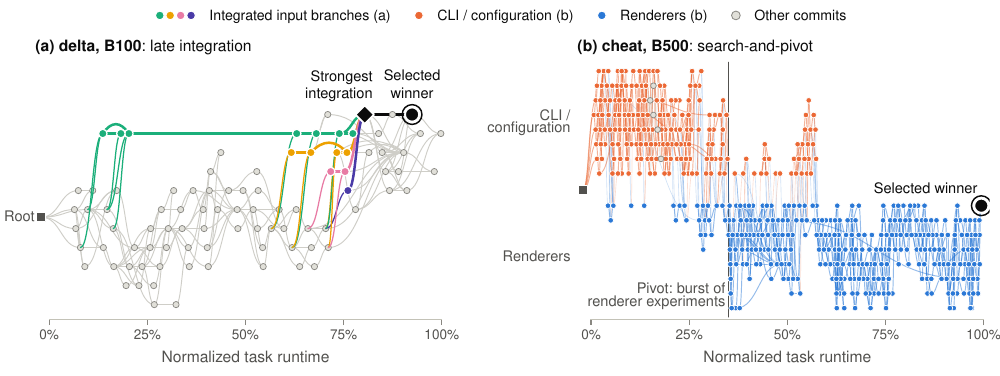}
    \caption{\textbf{Collaboration workflows in two ProgramBench runs.}
    Commits over normalized task runtime. (a) Separately developed branches
    (colored) converge late on the strongest integration (diamond), which the
    selected winner builds on. (b) Work pivots from CLI and configuration
    (orange) to renderers (blue), which become the selected lineage.}
    \label{fig:workflow-examples}
\end{figure}

\noindent\textbf{Selected-solution ancestry.}
The two tasks differ in how earlier work enters
the final solution. On ProgramBench, the selected  
solution's transitive ancestry covers
81.6--92.6\% of the contribution graph on
average across budgets. Git-parent ancestry
accounts for only about one quarter of the graph;
declared cross-branch dependencies connect most
of the remainder
(Figures~\ref{fig:winner-ancestry-coverage}
and~\ref{fig:ffmpeg-ancestry}). On IMOProofBench-Advanced, selected proofs reach
only 21.3--33.0\% of the graph. This can be explained by the fact that final solutions appear relatively early in the graph in the case of IMOProofBench. Later workers
often evaluate and nominate existing candidates
rather than extending them. Thus, programming
typically accumulates contributions into a
final implementation, while proof solving 
uses earlier work to compare and select candidates.
Contribution graphs also reveal interesting 
collaboration workflows, such as late integration
of separately developed branches and a gradual
pivot between subsystems
(Figure~\ref{fig:workflow-examples}).

\noindent\textbf{Artifact consumption.} We ask workers to leave informative and reusable artifacts in their commits, meant to be consumed by future workers which build upon their work. Declared dependencies do not by themselves establish that workers consume earlier artifacts. We thus analyse artifact consumption by workers from execution traces. 
On ProgramBench, 79.9--82.5\% of eligible
voluntary artifact bundles are actively reused,
often across Git branches.  On IMOProofBench-Advanced, workers open 583
of 759 computational probe bundles. Later
workers execute or adapt 172 of these while
citing their producers; 52 feed contributions
in the selected solution's ancestry. Required
reasoning and evaluation artifacts are also
accessed in approximately 98--99\% of eligible
revisions across both domains. Together, these measurements show that
persistent memory supports both synthesis and
candidate evaluation. They establish observed
reuse, not the causal contribution of individual
dependencies to performance. Measurement
definitions, more detailed artifact breakdowns and discussion on the workflow examples appear in
Appendices~\ref{app:reuse-details},
\ref{app:artifact-analysis}, and
\ref{app:workflow-examples}.

\subsection{Adaptive Coordination and Design Tradeoffs}
\label{sec:work-over-time}
\label{sec:coordination-dynamics}
\label{sec:ablations}

\begin{figure}[t]
    \centering
    \includegraphics[width=\linewidth]{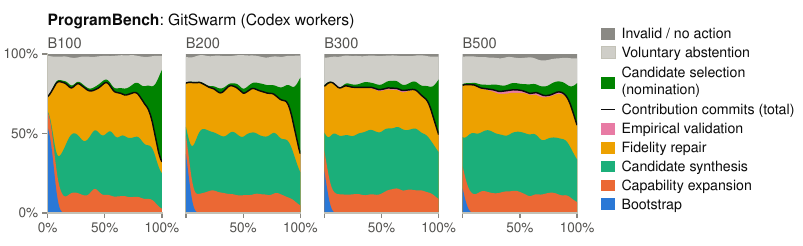}
    \caption{\textbf{ProgramBench worker activity over inference.}
    Share of worker episodes by activity category
    (Table~\ref{tab:worker-activity-taxonomies}) against
    worker-spawn progress, per worker budget; the black line
    tops all contribution commits. After a short bootstrap wave,
    workers keep expanding, repairing, and synthesizing
    candidates until termination, and nomination rises only
    near the end. The IMOProofBench-Advanced counterpart
    appears in Figure~\ref{fig:imo-work-mix}.}
    \label{fig:programbench-work-mix}
\end{figure}

The different ancestry patterns are reflected
in how workers allocate their effort. The 
workers in \gitswarm{} have no fixed assigned roles. New workers dynamically decide the kind of contribution to make depending on the work committed so far. We analyse the work done by workers over the systems' progress.  We label each
worker episode after the run with one primary
activity, using a bottom-up taxonomy derived
separately for each benchmark
(Appendix~\ref{app:worker-activity-taxonomies}).
On ProgramBench
(Figure~\ref{fig:programbench-work-mix}),
a short wave creates initial candidates, after
which capability expansion, fidelity repair,
and candidate synthesis persist through
termination. Nomination becomes more prominent
near the end - eating into, yet not completely replacing implementation
work. Thus additional workers find several ways to
improve the candidate rather than only repeat
comparisons. Empirical validation remains
uncommon and becomes visible mainly at larger
budgets. Consistent with this sustained
construction, the selected candidate first
appears at a median of 93.1--98.0\% of run
progress across budgets. On IMOProofBench-Advanced,
by contrast, workers shift early toward selecting
among existing proofs
(Appendix Figure~\ref{fig:imo-work-mix}):
selected proofs appear at 8.6--17.5\% of run
progress, and the share of episodes devoted to
selection rises from 44.2\% to 69.7\% as the
budget grows from 20 to 160. These patterns
emerge without assigned worker roles.

Workers also avoid some redundant work
and inspect the repository  (Appendix~\ref{app:inspection-strategies}).
About one in six ProgramBench workers
abstains after identifying equivalent work,
compared with fewer than 2.7\% of proof
workers. Abstention avoids duplicate
publication but cannot recover computation
already spent. Programming workers typically
inspect five to seven candidate commits,
and proof workers about three, rather than
reading the entire growing graph. Detailed
activity and coordination analyses appear in
Appendices~\ref{app:worker-activity-taxonomies},
\ref{app:voting-dynamics} and \ref{app:abstentions}.

\noindent\textbf{Design tradeoffs.}
We compare asynchronous execution and
decentralized readout on nine ProgramBench
development tasks, using GPT-5.5 workers,
a budget of 50 workers, five concurrent
workers, and three seeds. Asynchronous execution takes 38.0 minutes
on average, versus 49.3 minutes for
synchronized waves. Their observed scores
are $62.06\pm2.44\%$ and
$63.11\pm0.58\%$, respectively.

For readout, decentralized nomination
achieves $62.06\pm2.44\%$ in 38.0 minutes,
compared with $64.40\pm1.49\%$ in
34.9 minutes for a central selector.
A central aggregation agent achieves
$58.19\pm4.53\%$ in 43.9 minutes.
These small development-set experiments
characterize latency and readout tradeoffs;
they do not establish statistical equivalence
or isolate the benefits of persistent memory.
Full results appear in
Appendix~\ref{app:design-tradeoffs}.

\section{Discussion and Limitations}
\label{sec:limitations}

Our results reveal different forms of compounding 
inference. In program reconstruction, workers repeatedly
integrate discoveries from different branches into
increasingly capable solutions. In mathematical proof,
workers often use previous work to verify and compare
candidates without extending the selected proof's
lineage. The architecture research experiments illustrate
both cross-branch synthesis and the value of revisiting
earlier failures: the RMT architecture incorporates
mechanisms developed on separate branches, while the
Looped Transformer run recovers an initially rejected
idea through further evaluation. Useful computation
therefore need not appear directly in the final
candidate's ancestry; it may also inform which
directions are pursued, rejected, or selected.

\noindent\textbf{Causal attribution.}
Our dependency and execution-trace analyses establish
that workers inspect, reuse, and accumulate previous
work, but do not isolate the resulting performance
gains. The mandatory \texttt{informed\_by=} field records
worker-declared dependencies, while observed artifact
consumption provides evidence of use rather than
causal influence. Conversely, selected-solution
ancestry omits potentially useful verification and
candidate comparison. A direct intervention would
resume matched runs from the same candidate, with
and without access to its preceding research history,
under equal remaining compute. Separate interventions
would be needed to isolate the effects of persistence,
branchability, and cross-branch dependencies. Our
synthetic experiments demonstrate their operational
benefits but do not establish their individual
contributions to benchmark performance.

\noindent\textbf{Evaluation scope.}
Our results are conditional on the evaluated tasks
and budgets. ProgramBench performance is measured
on a fixed 50-task subset out of a total of 200 tasks. The perfect IMOProofBench-Advanced result is a
single run; some independent-sampling and
Forced baselines also reach the benchmark
ceiling, and execution conditions differ across
some proof comparisons. In AutoResearch, the
RMT swarm models and reference share the same
1.3B-token training budget, whereas the compared
single-agent models use more training tokens.
The seeded RMT run additionally inherits earlier
research. The recorded Looped Transformer runs
were still in progress, and NanoChat comparisons
vary in starting checkpoint and GPU allocation.
\gitswarm{} improves upon the starting points across
all three research tasks, but the results do not
establish superiority over every single-agent
configuration.

\noindent\textbf{Computational overhead.}
Persistent collaboration incurs costs as the
repository grows. Workers consume input tokens
inspecting previous contributions, while
redundant and unsuccessful episodes can spend
compute without producing accepted work.
GPU-backed research adds resource contention
and scheduling overhead. Although performance
improves with additional inference-time compute
over the tested ranges, understanding when the
benefits of accumulated work outweigh these
costs requires longer-horizon evaluations
with matched total compute and wall-clock budgets.

\section{Conclusion}
\label{sec:conclusion}

We introduced \emph{compounding  inference}, in which
intermediate work persists and can be extended across
independent agent episodes. \gitswarm{} provides a
substrate for this process through asynchronous workers,
a shared branchable repository, and explicit cross-branch
dependencies, without a central intellectual planner.
Across problem solving and GPU-backed research, we
observe extensive reuse of earlier work, including
the synthesis of independent discoveries and the
recovery of initially unsuccessful ideas. We view this substrate as a \emph{foundation for long-horizon
discovery}. It supports a continuing loop in which
agents propose ideas, run experiments, preserve their
findings, and use earlier evidence to guide new research.
A natural next step is to train agents with reinforcement
learning across diverse tasks within this environment.
Rather than rewarding only the outcome of an individual
episode, such training could incentivize contributions
whose value emerges through subsequent work, enabling
agents to learn not just to solve problems, but to
advance a \emph{collective process of discovery}.

\section*{AI Usage Statement}
Large language model based digital agents are the object of study in this work: every agent is a frontier model accessed through a provider API. An LLM is also used as a measurement instrument. Experiments combine semantic judgments by an LLM analyzer, and also to design synthetic tasks. LLMs based coding agents were used to assist in implementing code, which was reviewed by one author, and for generating draft versions of scientific figures. LLMs and generative AI tools were not used to develop theoretical models or conceptual frameworks, formulate mathematical claims, provide critical ingredients for proving mathematical claims, assist in the writing of proofs, propose or refine hypotheses, to support qualitative and thematic data analysis, design or provide feedback on research  methodology or experiments, assist with translation, clean and reformat dataset, interpret results. We have reviewed all AI-assisted work and take responsibility for the final content of this paper, including text, claims, and artifacts produced with the aid of generative AI.

\section*{Ethics Statement}
This work involves no human subjects, crowdsourced annotation, or personal data; all experiments use public benchmarks (IMOProofBench-Advanced and ProgramBench) and research tasks based on publicly available research papers. Because \gitswarm{} lets LLM agents execute code and launch GPU jobs autonomously, workers operated in isolated Git worktrees and isolated environments with no internet access throughout our experiments. Candidates were scored by fixed evaluators on protected development and separate held-out sets. 
Two capabilities studied here are dual-use: reconstructing programs from an execute-only reference and its documentation could facilitate unauthorized re-implementation of proprietary software, and automating architecture research could
accelerate AI development, including of systems with potential for misuse. We therefore view \gitswarm{} as a tool for human-supervised research; as our analyses show, agents' intermediate conclusions can be wrong (small proxy evaluations mis-ranked
candidate models), so results should be independently verified before they are relied upon. With recents events involving swarms of agents showing increasingly mis-aligned behavior and escaping sandboxes, we believe that extensive behavioral analysis, including the kind discussed in this paper, is warranted pre-deployment. 

\section*{Acknowledgements}

The authors would like to thank Sneha Silwal, Harman Singh, Vineet Jain, Nanda H Krishna, Sinong Wang, Hanjun Dai, Xinyun Chen, Xin Wang, Tianhe Yu, Liang Tan, Roshan James and Trapit Bansal for helpful discussions and feedback.

\bibliographystyle{assets/plainnat}
\bibliography{iclr2027_conference}

\clearpage
\beginappendix
\section*{Contents}

\begingroup
\small
\renewcommand{\arraystretch}{1.08}
\noindent\begin{tabularx}{\linewidth}{@{}lXlr@{}}
\toprule
& \textbf{Contents} & \textbf{Supports} & \textbf{Page} \\
\midrule
\textbf{\ref{sec:ext_related-work}} & \textbf{Related Work} &  & \pageref{sec:ext_related-work} \\
\textbf{\ref{app:setup-details}} & \textbf{Experimental Setup Details} & Sec.~\ref{sec:experimental-setup} & \pageref{app:setup-details} \\
\quad\ref{app:setup-benchmarks} & Benchmarks, Scoring, and Models & Sec.~\ref{sec:experimental-setup} & \pageref{app:setup-benchmarks} \\
\quad\ref{app:programbench-subsets} & ProgramBench Task Subsets & Sec.~\ref{sec:experimental-setup} & \pageref{app:programbench-subsets} \\
\quad\ref{app:setup-compute} & Baseline Continuation Protocol & Sec.~\ref{sec:experimental-setup} & \pageref{app:setup-compute} \\
\quad\ref{app:setup-analysis-corpus} & Run Corpus for the Trace Analyses & Secs.~\ref{sec:performance}, \ref{sec:worker-reuse}, \ref{sec:coordination-dynamics} & \pageref{app:setup-analysis-corpus} \\
\addlinespace
\textbf{\ref{app:synthetic-task-details}} & \textbf{Controlled Tests of Substrate Properties} & Sec.~\ref{sec:experiments} & \pageref{app:synthetic-task-details} \\
\quad\ref{app:hidden-rulebook-details} & Persistence: Hidden Rulebook & Sec.~\ref{sec:experimental-setup} & \pageref{app:hidden-rulebook-details} \\
\quad\ref{app:hidden-maze-details} & Branchability: Hidden Maze & Sec.~\ref{sec:experimental-setup} & \pageref{app:hidden-maze-details} \\
\quad\ref{app:retracted-evidence-details} & Attribution: Retracted Evidence & Sec.~\ref{sec:experimental-setup} & \pageref{app:retracted-evidence-details} \\
\quad\ref{app:synthetic-scope} & Scope of the Controlled Tests & Sec.~\ref{sec:experimental-setup} & \pageref{app:synthetic-scope} \\
\addlinespace
\textbf{\ref{app:benchmark-results}} & \textbf{Benchmark Performance Details} & Sec.~\ref{sec:performance} & \pageref{app:benchmark-results} \\
\addlinespace
\textbf{\ref{app:autoresearch}} & \textbf{AutoResearch: Additional Details} & Sec.~\ref{sec:autoresearch} & \pageref{app:autoresearch} \\
\quad\ref{app:autoresearch-setup} & Common Setup & Sec.~\ref{sec:autoresearch} & \pageref{app:autoresearch-setup} \\
\quad\ref{app:rmt} & Residual Matrix Transformer & Sec.~\ref{sec:autoresearch} & \pageref{app:rmt} \\
\quad\ref{app:loop} & Looped Transformer & Sec.~\ref{sec:autoresearch} & \pageref{app:loop} \\
\quad\ref{app:nanochat} & NanoChat & Sec.~\ref{sec:autoresearch} & \pageref{app:nanochat} \\
\quad\ref{app:autoresearch-findings} & Findings Across Research Tasks & Secs.~\ref{sec:autoresearch}, \ref{sec:limitations} & \pageref{app:autoresearch-findings} \\
\addlinespace
\textbf{\ref{app:reuse-accumulation}} & \textbf{Reuse, Artifacts, and Accumulation} & Sec.~\ref{sec:worker-reuse} & \pageref{app:reuse-accumulation} \\
\quad\ref{app:reuse-details} & Declared Reuse and Selected-Solution Ancestry & Sec.~\ref{sec:worker-reuse} & \pageref{app:reuse-details} \\
\quad\ref{app:artifact-analysis} & Persistent Artifact Production and Consumption & Sec.~\ref{sec:worker-reuse} & \pageref{app:artifact-analysis} \\
\quad\ref{app:workflow-examples} & Collaboration Workflows & Sec.~\ref{sec:worker-reuse} & \pageref{app:workflow-examples} \\
\addlinespace
\textbf{\ref{app:additional-coordination}} & \textbf{Worker Activity, Coordination, and Design Tradeoffs} & Sec.~\ref{sec:coordination-dynamics} & \pageref{app:additional-coordination} \\
\quad\ref{app:worker-activity-taxonomies} & Worker-Activity Taxonomies, Activity, and Coordination-Tool Use over Time & Sec.~\ref{sec:coordination-dynamics} & \pageref{app:worker-activity-taxonomies} \\
\quad\ref{app:system-health} & Worker Episode Outcomes and System Health & Secs.~\ref{sec:coordination-dynamics}, \ref{sec:limitations} & \pageref{app:system-health} \\
\quad\ref{app:voting-dynamics} & Voting Dynamics and Selection Timing & Sec.~\ref{sec:coordination-dynamics} & \pageref{app:voting-dynamics} \\
\quad\ref{app:abstentions} & Voluntary Abstention and Redundancy & Sec.~\ref{sec:coordination-dynamics} & \pageref{app:abstentions} \\
\quad\ref{app:inspection-strategies} & Selective Inspection of Shared History & Sec.~\ref{sec:coordination-dynamics} & \pageref{app:inspection-strategies} \\
\quad\ref{app:design-tradeoffs} & Architectural Tradeoffs & Sec.~\ref{sec:ablations} & \pageref{app:design-tradeoffs} \\
\addlinespace
\textbf{\ref{app:gitswarm-system-prompt}} & \textbf{\gitswarm{} Worker Prompts} & Secs.~\ref{sec:method}, \ref{sec:experimental-setup} & \pageref{app:gitswarm-system-prompt} \\
\quad\ref{app:prompts-system} & System Prompt & Sec.~\ref{sec:method} & \pageref{app:prompts-system} \\
\quad\ref{app:prompts-user} & User Prompts & Sec.~\ref{sec:experimental-setup} & \pageref{app:prompts-user} \\
\bottomrule
\end{tabularx}
\endgroup

\clearpage

\section{Related Work}
\label{sec:ext_related-work}

Our work lies at the intersection of inference-time
search, agent memory, multi-agent coordination, and
automated research. These approaches differ in how
they preserve intermediate work, support competing
lines of exploration, record dependencies, and
decide what to investigate next.
Table~\ref{tab:related-work-design} compares
representative systems along these dimensions.

\noindent\textbf{Inference-time search and reflection.}
Repeated sampling and compute-aware inference
allocate additional computation to candidate
solutions~\citep{bestofn,testtimescaling}.
Tree of Thoughts explores and evaluates
alternative reasoning paths~\citep{rw_tot2023},
while Language Agent Tree Search combines
tree search with environmental
feedback~\citep{rw_lats2024}.
These approaches preserve intermediate states
within a search. In contrast, \gitswarm{}
preserves partial solutions, executable artifacts,
and negative results across independently
scheduled worker episodes.

Other methods retain information across attempts.
Reflexion preserves verbal feedback for
subsequent trials~\citep{rw_reflexion2023};
ExpeL extracts and retrieves insights from
previous tasks~\citep{rw_expel2024}; and
Voyager accumulates reusable executable
skills~\citep{rw_voyager2023}.
MemGPT manages long-term context
\citep{rw_memgpt2023}, while A-MEM links
and updates memories as new information
arrives~\citep{rw_amem2025}.
These systems establish the value of external
memory. Our focus is a shared, versioned
contribution history in which independent
workers can pursue competing approaches
and explicitly declare dependencies across them. 

PDR~\citep{rw_pdr2025} study inference
as an iterative improvement process. Their
Parallel-Distill-Refine method generates
diverse drafts in parallel, distills them into
a bounded textual workspace, and refines a
candidate over successive rounds. Similarly, Recursive Self-Aggregation (RSA)~\citep{venkatraman2025rsa} involves sampling multiple ($N$) outputs in parallel, 
curating new contexts by randomly sub-sampling and collecting $K$ outputs, and \textit{aggregating} them using an LLM operator. This workflow is then repeated over successive rounds. PDR and RSA allow
inference to accumulate information without
continually expanding the model's context. SqueezeEvolve~\citep{maheswaran2026squeeze} follows a similar orchestration. \gitswarm{} shares the goal of building on earlier
computation but uses a different memory model.
Rather than repeatedly compressing drafts into
a single workspace, it preserves individual
contributions in a shared, branchable repository.
Independent workers choose which previous results
to inspect or extend, while each commit explicitly
records its dependencies across branches. \gitswarm{}
also operates entirely at inference time, whereas
Madaan et al.\ additionally investigate training
models to use their iterative improvement operator.

Kim et al.~\citep{rw_agentic_scaling2026}
extend inference-time scaling to long-horizon
coding agents. They summarize each agent
trajectory into a structured record of its
hypotheses, progress, and failures. These
summaries support recursive tournament voting
over parallel attempts and sequential refinement
through Parallel-Distill-Refine (PDR). Like \gitswarm{}, their method allows later
computation to benefit from earlier attempts.
The difference lies in how that work is
preserved and organized. Their approach
distills trajectories into compact contexts
for selection and subsequent rollouts.
\gitswarm{} instead preserves individual
contributions and artifacts in a shared,
branchable repository. Workers independently
choose which contributions to extend, while
each commit explicitly records its dependencies
across branches.

\noindent\textbf{Multi-agent coordination.}
Multi-agent debate improves candidate answers
through interaction between model
instances~\citep{multiagent}, while
Mixture-of-Agents aggregates responses across
successive model layers~\citep{wang2025moa}.
AutoGen supports configurable patterns of
agent conversation~\citep{wu2023autogen}.
MetaGPT assigns specialized roles within
a prescribed software-development
workflow~\citep{hong2024metagpt}, and
AgentVerse studies configurable teams and
collaboration patterns~\citep{chen2024agentverse}.
Magentic-One uses an orchestrator to assign
work and revise plans~\citep{rw_magenticone2024}.

GPTSwarm represents agent operations and
communication as an optimizable computation
graph~\citep{rw_gptswarm2024}.
Its graph specifies how a multi-agent program
executes, whereas ours records how published
contributions accumulate over time.
Rather than assigning roles or optimizing a
communication topology, \gitswarm{} lets
homogeneous workers independently choose
whether to explore, repair, verify, integrate,
or nominate existing work. Its harness still
centrally manages execution and applies
deterministic plurality readout.

\noindent\textbf{Version control for collaborating agents.}
AgentGit introduces version-control operations
such as committing, branching, and rollback
to agent workflows, supporting recovery and
alternative execution paths~\citep{rw_agentgit2025}.
CAID uses parallel Git worktrees and
branch-and-merge integration for asynchronous
software engineering, with a central manager
that plans and delegates
subtasks~\citep{rw_caid2026}.
SwarmResearch also runs coding agents on
separate Git branches, but uses a Shepherd
Agent to direct their search and adjust
parallelism~\citep{rw_swarmresearch2026}.

These systems demonstrate how version control
can organize parallel agent work. In \gitswarm{},
the repository instead serves as a persistent
substrate for self-directed contributions.
Each accepted commit has a physical Git parent
and must declare its semantic dependencies
through \texttt{informed\_by=}, including references
to work on other branches. Nomination records
remain separate from the contribution graph,
so evaluating an existing candidate does
not alter its ancestry.

\noindent\textbf{Concurrent work: Agora.}
Agora~\citep{rw_agora2026}, developed in
parallel with our work, also investigates
compounding research by self-directed agents
over a persistent Git-backed contribution
graph. Its workers operate without assigned
tasks or a central planner. Contributions
retain their dependencies through Git
parent relationships, including multiple
parents, while searchable frontier views
and diversity-aware recommendations help
workers navigate the research history.

\gitswarm{} studies a closely related design
space with a different contribution and
selection protocol. It separates each
commit's physical Git parent from mandatory
semantic \texttt{informed\_by=} references and
selects final candidates through worker
nominations and deterministic plurality
readout. Our evaluation also examines
fixed-answer benchmarks, controlled
substrate properties, and sustained
GPU-backed research, with separate
measurements of declared dependencies,
artifact consumption, and selected-solution
ancestry. Agora's extended weight-transfer
study instead emphasizes research diversity
and independent verification. The two
studies provide complementary evidence
about repository-mediated collective
research rather than a direct performance
comparison.

\noindent\textbf{Blackboards and global workspaces.}
GitSwarm belongs to a long lineage of architectures that coordinate
distributed computation through shared state.
Classical blackboard systems such as HEARSAY-II allow independent knowledge
sources to opportunistically read from and contribute to a common
workspace~\citep{erman1980hearsay,nii1986blackboard}.
Global Workspace Theory similarly organizes cognition around multiple
specialist processes whose information can gain access to a shared broadcast
workspace~\citep{baars1988cognitive}.
Shanahan developed computational realizations of this principle in cognitive
and neural architectures based on competition and global broadcast
\citep{shanahan2006cognitive,shanahan2008spiking}, and
\citet{goyal2022workspace} brought the idea to modern neural architectures,
allowing learned specialist modules to communicate through a capacity-limited
shared workspace.

GitSwarm inherits the shared-workspace principle but changes the temporal and
structural role of the workspace.
Rather than maintaining a transient common state for coordinating computation
within a cognitive cycle or neural forward process, GitSwarm preserves an
explicit history of contributions across agent episodes.
The workspace is durable and branchable, and dependencies between
contributions are recorded explicitly, allowing competing lines of work to
persist and later recombine.
Our focus is therefore not only coordination through shared state, but the
\emph{accumulation of reusable computation} over long horizons.

\noindent\textbf{Autonomous experimentation
and evolutionary search.}
Karpathy's AutoResearch repeatedly modifies,
trains, and evaluates a NanoChat model in a
single-agent loop~\citep{rw_autoresearch2026}.
The AI Scientist automates stages from idea
generation to paper writing
\citep{rw_aiscientist2024}, while AI Scientist-v2
uses progressive agentic tree search directed
by an experiment manager
\citep{rw_aiscientistv22025}.
Agent Laboratory organizes literature review,
experimentation, and writing into a staged
research workflow~\citep{rw_agentlab2025}.

AlphaEvolve maintains an evaluated population
of programs and uses evolutionary selection
for algorithm discovery
\citep{rw_alphaevolve2025}.
AIRA$_2$ combines agentic experimentation,
consistent evaluation, and asynchronous
multi-GPU execution
\citep{hambardzumyan2026aira2}.
These systems demonstrate the value of
experimental feedback, persistent candidate
pools, and parallel research.

Our question is how independent research
episodes can additionally preserve hypotheses,
failed experiments, code, and supporting
evidence in an auditable contribution graph,
without a central agent selecting each next
experiment. The research histories in
Section~\ref{sec:autoresearch} illustrate
cross-branch architectural synthesis and
the later reconsideration of initially
rejected results. These observations
demonstrate reuse, but do not by themselves
establish its causal effect on performance.

Earlier work on autonomous research establishes
complementary ways to organize experimental search.
AIRA~\citep{rw_aira2025} studies how search policies
and operators interact in machine learning research,
while AI Co-Scientist~\citep{rw_coscientist2025}
uses asynchronous agents and tournament-based
refinement to develop scientific hypotheses.
The Darwin G\"odel Machine~\citep{rw_dgm2025}
maintains an expanding archive of coding agents,
empirically validating modifications that improve
the agents themselves.
ARTS~\citep{juneja2026learning} uses a central reasoning
model to inspect the logs of earlier experiments and
choose which node of a search tree to expand, separating
weak hypotheses from flawed implementations rather than
ranking nodes by score. Because this scientist carries the
entire search, ARTS condenses the growing history into
short written insights and, in a variant, into the model's
weights via test-time training, and it returns the best
validated node. \gitswarm{} has no central scientist:
independent workers, each starting with a fresh context,
retrieve what they need from a shared version-controlled
persistent repository that is never summarized or pruned,
can combine separate branches into a single contribution,
and select the final solution through decentralized
nomination when no reliable validation score exists.
\gitswarm{} instead keeps
the worker model fixed and accumulates contributions
to an external research task. Its repository
preserves competing experiments, including
unsuccessful ones, with explicit dependencies
between contributions.

Meta-Harness~\citep{lee2026meta} searches over
the code of LLM harnesses with a single coding-agent
proposer that reads the source code, scores, and
execution traces of all prior candidates from a
filesystem. Like \gitswarm{}, it exposes the full,
uncompressed history for selective retrieval rather
than summarizing it. Its search, however, proceeds in
sequential iterations, and every stored candidate is a
complete harness, evaluated on a fixed search set and
logged in a predefined per-candidate directory.
\gitswarm{} contributions can instead be bounded partial
work that later workers extend, providing a higher
branching factor and thus allowing more exploration.
Branches and explicit dependencies let many asynchronous
workers build on one another concurrently under version
control, and beyond a small core layout, workers choose
which artifacts to leave. \gitswarm{} is also applied
across program reconstruction, proof search, and
architecture research rather than to a single class of
artifact.

\noindent\textbf{Evaluating autonomous research.}
MLAgentBench~\citep{rw_mlagentbench2024},
MLGym~\citep{rw_mlgym2025},
ResearchGym~\citep{rw_researchgym2026},
and AIRS-Bench~\citep{rw_airsbench2026}
provide environments and benchmarks for
end-to-end machine learning research.
MLGym additionally provides an environment
for developing reinforcement learning
algorithms for research agents.
These works motivate evaluation beyond
fixed-answer coding tasks. Our experiments
instead examine how independently scheduled
workers accumulate code and experimental
evidence through shared persistent memory.

\begin{table}[t]
    \centering
    \caption{
        \textbf{How related systems organize
        intermediate work.}
        Representative mechanisms for retaining
        state, exploring alternatives, recording
        dependencies, and choosing subsequent
        work. This is a comparison of system
        designs, not a performance ranking.
    }
    \label{tab:related-work-design}
    \scriptsize
    \setlength{\tabcolsep}{3pt}
    \renewcommand{\arraystretch}{1.12}
    \begin{tabularx}{\linewidth}{@{}
        >{\raggedright\arraybackslash}p{0.16\linewidth}
        >{\raggedright\arraybackslash}p{0.19\linewidth}
        >{\raggedright\arraybackslash}p{0.17\linewidth}
        >{\raggedright\arraybackslash}p{0.20\linewidth}
        >{\raggedright\arraybackslash}X@{}
    }
        \toprule
        System
        & Retained state
        & Alternative paths
        & Recorded dependencies
        & Next-work control \\
        \midrule

        ToT; LATS
        \citep{rw_tot2023,rw_lats2024}
        & Search states
        & Reasoning tree
        & Search edges
        & Search policy \\

        Reflexion; ExpeL
        \citep{rw_reflexion2023,rw_expel2024}
        & Reflections and experiences
        & Successive attempts
        & Episode or insight memory
        & Individual agent \\

        MetaGPT; AutoGen
        \citep{hong2024metagpt,wu2023autogen}
        & Messages and task artifacts
        & Configured interactions
        & Workflow and messages
        & Roles or configured dialogue \\

        GPTSwarm
        \citep{rw_gptswarm2024}
        & Executable agent graph
        & Graph variants
        & Operation-level edges
        & Graph optimizer \\

        AgentGit
        \citep{rw_agentgit2025}
        & Versioned workflow states
        & Rollback and branches
        & State lineage
        & Application workflow \\

        CAID
        \citep{rw_caid2026}
        & Git worktrees and commits
        & Parallel branches and merges
        & Code integration lineage
        & Central delegator \\

        SwarmResearch
        \citep{rw_swarmresearch2026}
        & Git branch states
        & Parallel search agents
        & Branch-level history
        & Shepherd agent \\

        AI Scientist-v2
        \citep{rw_aiscientistv22025}
        & Research experiments
        & Progressive search tree
        & Search and experiment lineage
        & Experiment manager \\

        AlphaEvolve
        \citep{rw_alphaevolve2025}
        & Candidate database
        & Evolutionary population
        & Candidate variations and scores
        & Evolutionary selection \\

        AIRA$_2$
        \citep{hambardzumyan2026aira2}
        & Experiment records
        & Asynchronous GPU workers
        & Experimental feedback
        & Research agents and harness \\

        Agora
        \citep{rw_agora2026}
        & Append-only Git contributions
        & Shared research DAG
        & Git-parent dependencies
        & Independent workers and
          frontier suggestions \\

        \textbf{\gitswarm{} (ours)}
        & Atomic Git contributions
        & Persistent Git branches
        & Git parent and required
          cross-branch \texttt{informed\_by=}
        & Independent workers;
          plurality readout \\

        \bottomrule
    \end{tabularx}
\end{table}

\paragraph{Comparison by design principles.}
Table~\ref{tab:related-work-desiderata}
compares systems along seven properties relevant
to compounding inference: shared memory that
persists across agent episodes; durable branches
for competing approaches; self-directed work
selection without a central task planner;
absence of predefined specialist roles;
asynchronous workers; explicit dependencies
across branches; and dependency records that distinguish
the physical starting state from other sources
used by a contribution.

These properties are not independent, nor are they
a performance ranking. In particular, systems
can provide several of them through different
mechanisms. We compare the protocols described
in the cited papers, rather than every configuration
their underlying frameworks could support.

\begin{table}[t]
    \centering
    \caption{
        \textbf{Design properties of related systems.}
        Shared memory persists across worker episodes;
        branching preserves competing contributions;
        self-direction means workers choose their
        next task rather than receive assignments.
        Cross-branch dependencies explicitly connect
        contributions from different branches.
        Separate sources means the physical starting
        state is recorded apart from additional semantic sources.
        $\checkmark$: part of the reported design;
        $\sim$: partial or configuration-dependent;
        --: not part of the reported protocol.
    }
    \label{tab:related-work-desiderata}
    \scriptsize
    \setlength{\tabcolsep}{2.5pt}
    \renewcommand{\arraystretch}{1.18}
    \begin{tabular}{@{}lccccccc@{}}
        \toprule
        System
        & \shortstack{Shared\\memory}
        & \shortstack{Durable\\branches}
        & \shortstack{Self-\\directed}
        & \shortstack{No fixed\\roles}
        & \shortstack{Async\\workers}
        & \shortstack{Cross-branch\\deps.}
        & \shortstack{Separate\\sources} \\
        \midrule

        MetaGPT~\citep{hong2024metagpt}
        & -- & -- & -- & -- & -- & -- & -- \\

        AutoGen~\citep{wu2023autogen}
        & $\sim$ & $\sim$ & $\sim$
        & $\sim$ & $\sim$ & -- & -- \\

        AgentGit~\citep{rw_agentgit2025}
        & $\sim$ & $\checkmark$ & $\sim$
        & $\sim$ & $\sim$ & -- & -- \\

        CAID~\citep{rw_caid2026}
        & $\checkmark$ & $\checkmark$ & --
        & -- & $\checkmark$ & $\sim$ & -- \\

        SwarmResearch~\citep{rw_swarmresearch2026}
        & $\checkmark$ & $\checkmark$ & --
        & -- & $\checkmark$ & -- & -- \\

        \midrule

        \textbf{\gitswarm{}}
        & $\checkmark$ & $\checkmark$
        & $\checkmark$ & $\checkmark$
        & $\checkmark$ & $\checkmark$
        & $\checkmark$ \\

        \bottomrule
    \end{tabular}
\end{table}

\paragraph{Relation to repository-based collaboration.}
CAID and SwarmResearch use Git branches to
support parallel work, but organize that work
through a central manager or Shepherd Agent.
The former delegates dependent software
engineering subtasks; the latter chooses
starting points and steers exploration.
\gitswarm{} instead lets each worker inspect
the repository and independently choose
what contribution to make.

Agora is closer in its control model. Its
workers also choose their own research
directions without assigned roles or a
central planner. Both systems preserve
intermediate research in a Git-backed
contribution graph, including alternative
approaches and unsuccessful experiments. Their dependency tracking and selection mechanisms
differ. Agora represents dependencies through
Git parent relationships, including multiple
parents, and provides evidence-aware and
diversity-aware views of the research frontier.
\gitswarm{} separates the physical Git parent
from additional mandatory \texttt{informed\_by=}
references, allowing a worker to build on
one branch while explicitly citing discoveries
from others. Workers nominate candidate
solutions, and the harness applies
deterministic plurality readout.

These differences motivate distinct analyses.
We measure declared cross-branch dependencies,
trace-visible artifact consumption, and the
transitive ancestry of selected solutions
across programming and proof benchmarks.
Our sustained research experiments additionally
illustrate architectural synthesis and the
later reuse of negative experimental results.

\section{Experimental Setup Details}
\label{app:setup-details}

This appendix complements Section~\ref{sec:experimental-setup} with
benchmark, scoring, and model details
(Appendix~\ref{app:setup-benchmarks}), the ProgramBench task subsets
(Appendix~\ref{app:programbench-subsets}), the baseline continuation protocol
(Appendix~\ref{app:setup-compute}), and the run corpus
underlying the trace analyses (Appendix~\ref{app:setup-analysis-corpus}).

\subsection{Benchmarks, Scoring, and Models}
\label{app:setup-benchmarks}

\paragraph{IMOProofBench-Advanced.}
The 30 proof-style problems of IMOProofBench-Advanced~\citep{luong2025robust}
include problems sourced from high-school level mathematical competitions as
well as novel problems designed by mathematicians. All 30 problems are
evaluated. Candidate proofs are graded using the benchmark's proof-scoring
protocol~\citep{luong2025robust}; each solution receives a score in
$\{0,1,6,7\}$. In the scaling analysis
(Figure~\ref{fig:imoproofbench-gemini-scaling}), a proof counts as fully
correct when it receives the full score of 7.

With Codex (GPT-5.5-high) workers, \gitswarm{} is compared with a single
Forced Codex (GPT-5.5-high) agent; with Gemini-3.1-Pro workers, it is
compared with Gemini-3.1-Pro wrapped in the mini-SWE-agent
harness~\citep{lieret2025minisweagent}. The proof baselines also include
synchronized independent best-of-five trajectories
(Section~\ref{sec:experimental-setup}).

\paragraph{ProgramBench.}
Each ProgramBench~\citep{yang2026programbench} task provides documentation
and an execute-only reference program; the solver must construct a
self-contained codebase whose compiled executable matches the reference on
hidden tests. The benchmark score is the fraction of hidden cases on which
the candidate's behavior matches the reference. The \gitswarm{} Codex
workers use GPT-5.5 at high reasoning effort, the \gitswarm{} GPT-5.5 workers
use the serving endpoint's default reasoning effort, and the Forced
Codex agent uses GPT-5.5 at high reasoning effort. The development and
evaluation task subsets are described in Appendix~\ref{app:programbench-subsets}.

\paragraph{Architecture research tasks.}
For each research task of Section~\ref{sec:autoresearch} we report task-specific validation metrics, research wall-clock budgets, worker
and GPU allocations, and comparison conditions. The common research setting
and reporting protocol are given in Appendix~\ref{app:autoresearch-setup},
and the task specifications in Appendix~\ref{app:autoresearch}.

\subsection{ProgramBench Task Subsets}
\label{app:programbench-subsets}

\paragraph{Development tasks.}
We use nine ProgramBench tasks during development, including for the
design-tradeoff comparisons (Table~\ref{tab:design-tradeoffs}):
\texttt{wfxr\_\_csview},
\texttt{johanneskaufmann\_\_html-to-markdown},
\texttt{altdesktop\_\_i3-style},
\texttt{elkowar\_\_pipr},
\texttt{tomnomnom\_\_gron},
\texttt{gabotechs\_\_dep-tree},
\texttt{rvben\_\_rumdl},
\texttt{sigoden\_\_argc}, and
\texttt{johnkerl\_\_miller}.
None of them is part of the evaluation subset.

\paragraph{Evaluation subset.}
The 50 evaluation tasks are drawn from the tasks that remain after
excluding the nine development tasks and six further tasks. The subset is
selected by stratifying jointly on several dimensions, including
implementation language (26 Rust, 11 Go, 8 C, 3 C++, 1 Haskell, and 1 Java
task), benchmark difficulty labels (7 easy, 30 medium, 4 hard, and 9
unlabeled), test-suite size (10 tasks from each quintile), and runtime
(balanced across quartiles). Remaining ties are broken deterministically
with a fixed hash. Table~\ref{tab:programbench-eval-tasks} lists the 50 tasks.

\begin{table}[t]
  \centering
  \footnotesize
  \renewcommand{\arraystretch}{1.2}
  \caption{\textbf{ProgramBench evaluation subset.} The 50 evaluation tasks,
  grouped by implementation language. Identifiers have the form
  \texttt{owner\_\_repository.commit}.}
  \label{tab:programbench-eval-tasks}
  \begin{tabularx}{\linewidth}{@{}l>{\raggedright\arraybackslash}X@{}}
    \toprule
    Language & Tasks \\
    \midrule
    C (8) & \texttt{abishekvashok\_\_cmatrix.5c082c6}, \texttt{arq5x\_\_bedtools2.dd57059}, \texttt{chirlu\_\_sox.42b3557}, \texttt{ffmpeg\_\_ffmpeg.360a402}, \texttt{jqlang\_\_jq.b33a763}, \texttt{lua\_\_lua.c6b4848}, \texttt{sqlite\_\_sqlite.839433d}, \texttt{universal-ctags\_\_ctags.243595e} \\
    \addlinespace
    C{+}{+} (3) & \texttt{gromacs\_\_gromacs.665ea4c}, \texttt{halitechallenge\_\_halite.822cfb6}, \texttt{osgeo\_\_gdal.0847f12} \\
    \addlinespace
    Go (11) & \texttt{ariga\_\_atlas.6d81150}, \texttt{cheat\_\_cheat.b8098dc}, \texttt{dundee\_\_gdu.ede21d2}, \texttt{incu6us\_\_goimports-reviser.81bd549}, \texttt{mgechev\_\_revive.201451e}, \texttt{multiprocessio\_\_dsq.c3ae0ba}, \texttt{noborus\_\_ov.b96c2ba}, \texttt{noborus\_\_trdsql.d8c5ff6}, \texttt{psampaz\_\_go-mod-outdated.bb79367}, \texttt{segmentio\_\_chamber.5f93f5f}, \texttt{sheepla\_\_pingu.926d475} \\
    \addlinespace
    Haskell (1) & \texttt{jgm\_\_pandoc.5caad90} \\
    \addlinespace
    Java (1) & \texttt{stathissideris\_\_ditaa.f2286c4} \\
    \addlinespace
    Rust (26) & \texttt{agourlay\_\_zip-password-finder.704700d}, \texttt{alexpovel\_\_srgn.89f943b}, \texttt{bensadeh\_\_tailspin.6278437}, \texttt{brocode\_\_fblog.3b54330}, \texttt{burntsushi\_\_ripgrep.3b7fd44}, \texttt{burntsushi\_\_xsv.f430466}, \texttt{cordx56\_\_rustowl.655bc5c}, \texttt{crowdagger\_\_crowbook.ea214d7}, \texttt{dalance\_\_amber.69a0f52}, \texttt{dandavison\_\_delta.acd758f}, \texttt{isona\_\_dirble.e2dea9f}, \texttt{jrnxf\_\_thokr.09375ef}, \texttt{nachoparker\_\_dutree.44e877d}, \texttt{nikolassv\_\_bartib.6b9b5ce}, \texttt{o2sh\_\_onefetch.e5958ce}, \texttt{oppiliappan\_\_statix.e9df54c}, \texttt{parcel-bundler\_\_lightningcss.aa2ed1e}, \texttt{rhysd\_\_kiro-editor.4157485}, \texttt{shashwatah\_\_jot.a92aad8}, \texttt{simeg\_\_eureka.df3796c}, \texttt{sirwart\_\_ripsecrets.34c9e03}, \texttt{sitkevij\_\_hex.61ae69b}, \texttt{trasta298\_\_keifu.3331426}, \texttt{wfxr\_\_code-minimap.0ddeea5}, \texttt{wgunderwood\_\_tex-fmt.3f1aef6}, \texttt{xampprocky\_\_tokei.505d648} \\
    \bottomrule
  \end{tabularx}
\end{table}

\subsection{Baseline Continuation Protocol}
\label{app:setup-compute}

\paragraph{Natural-stop continuation for Forced baselines.}
For the Forced baselines, we use natural stopping as the
budget knob: every time the agent stops naturally, we resume the session by
prompting it with \texttt{Keep working...}. Intermediate
checkpoints of the Forced agent are graded to obtain a compute-scaling curve.

\subsection{Run Corpus for the Trace Analyses}
\label{app:setup-analysis-corpus}

\paragraph{Proof runs.}
Because GPT-5.5 saturates the benchmark at relatively small compute budgets
(Section~\ref{sec:performance}), the cross-budget proof scaling discussion
emphasizes the Gemini-3.1-Pro runs. The IMOProofBench-Advanced trace
analyses of dependencies, artifact access, worker activity, and coordination
(Sections~\ref{sec:worker-reuse}--\ref{sec:coordination-dynamics};
Appendices~\ref{app:reuse-accumulation} and~\ref{app:additional-coordination})
use the Gemini-3.1-Pro \gitswarm{} proof runs.

\paragraph{ProgramBench runs.}
The ProgramBench trace analyses cover the four scaling budgets
$B=100$--$500$ on the 50-task evaluation subset.

\section{Controlled Tests of Substrate Properties}
\label{app:synthetic-task-details}

This appendix gives the full task definitions, interfaces, treatments,
settings, and results for the three controlled tasks summarized in
Section~\ref{sec:experimental-setup}. Each synthetic task isolates a
setting in which one property of the shared substrate (persistence,
branchability, or attribution; Section~\ref{sec:desiderata}) should reduce a
particular form of repeated work. All experiments use Gemini-3.1-Pro and
three random seeds; Appendix~\ref{app:synthetic-scope} discusses the scope of
these tests. Symbols such as $B$, $C$, $D$, and $k$ are local to the
synthetic tasks; in particular, $B$ denotes the number of workers in a
synthetic run, and $C$ in the Retracted Evidence task counts
candidate conclusions, not the concurrency limit of
Section~\ref{sec:worker-execution}. The worked examples below use short
readable IDs; experimental instances use random opaque IDs and randomize
their display order for each worker.

\subsection{Persistence: Hidden Rulebook}
\label{app:hidden-rulebook-details}

\paragraph{Task generation.}
The hidden rulebook is a collection of records with opaque IDs. A record is
either a fact containing an integer, or a unary rule containing the ID of one
earlier record and an addition or multiplication to apply modulo $997$. For
each of the $P$ requested targets, the generator creates an independent chain
of $D$ rules ending in one fact. Solving all targets therefore requires
exactly $Q=P(D+1)$ relevant records. The remainder of the rulebook consists of
disconnected distractor facts and rules. A worker initially sees only the
target IDs. Inspecting a rule reveals its predecessor's ID but not that
predecessor's contents, so each chain must be discovered one record at a time
(Figure~\ref{fig:app-hidden-rulebook-example}).

\begin{figure}[tbp]
  \centering
  \begin{minipage}{0.94\linewidth}
\begin{lstlisting}[
  basicstyle=\fontfamily{lmtt}\selectfont\scriptsize,
  frame=single,
  columns=fullflexible,
  keepspaces=true
]
Visible initially: target t7

inspect(t7) -> multiply predecessor r4 by 3
inspect(r4) -> add 5 to predecessor f2
inspect(f2) -> fact 7

Evaluation: f2 = 7 -> r4 = 12 -> t7 = 36

Independent:  worker 0: t7, r4     worker 1: t7, r4  (repeats)
Persistent:   worker 0: t7, r4     worker 1: f2      (continues)
\end{lstlisting}
  \end{minipage}
  \caption{\textbf{Hidden Rulebook example.} A worker cannot query \texttt{f2}
  until an earlier query reveals its ID\@. Shared observations let a later
  worker continue the chain instead of rediscovering its prefix.}
  \label{fig:app-hidden-rulebook-example}
\end{figure}

\paragraph{Interface and treatments.}
\texttt{inspect(record\_id)} costs one query and returns either a fact value
or a rule's predecessor, operation, operand, and modulus. The harness appends
each successful result to the current worker's \texttt{observations.jsonl}.
\texttt{bash(command)} lets a worker inspect filesystem memory and write
notes, and \texttt{submit\_answers(answers)} records its target--value map and
ends the episode. We compare shared persistent memory against two independent
conditions. In the shared persistent treatment, workers run sequentially and
each new worker can read the completed directories of all previous workers,
which are mounted read-only. In the independent treatments, every worker
receives an isolated directory. Target order is independently permuted for
every worker in all treatments, preventing a fixed presentation order from
determining which chains are explored first.

\paragraph{Experimental settings.}
We sweep the number of targets $P$ (with $D=4$) and the chain depth $D$ (with
$P=4$). The shared persistent condition uses $k=10$ queries per worker and
enough sequential workers to provide approximately $Q$ queries in total. We
include two deliberately over-budget independent baselines: two relatively
long workers ($B=2$), and many short workers with $k=4$ queries each. Their
aggregate budget is approximately $4Q/3>Q$, so their lower coverage cannot be
explained by having too few total queries.
Table~\ref{tab:hidden-rulebook-settings} gives the exact settings.

\begin{table}[tbp]
  \centering
  \small
  \caption{\textbf{Hidden Rulebook settings.} Entries in the final three
  columns give $(B,k)$: the number of workers and queries per worker.}
  \label{tab:hidden-rulebook-settings}
  \begin{tabular}{@{}llccc@{}}
    \toprule
    Sweep & Value & Shared persistent & Independent, $B=2$ & Independent, $k=4$ \\
    \midrule
    $P$ ($D=4$) & 4  & $(2,10)$  & $(2,14)$  & $(7,4)$  \\
                & 8  & $(4,10)$  & $(2,27)$  & $(14,4)$ \\
                & 12 & $(6,10)$  & $(2,40)$  & $(20,4)$ \\
                & 16 & $(8,10)$  & $(2,54)$  & $(27,4)$ \\
                & 32 & $(16,10)$ & $(2,107)$ & $(54,4)$ \\
    \midrule
    $D$ ($P=4$) & 4  & $(2,10)$  & $(2,14)$  & $(7,4)$  \\
                & 8  & $(4,10)$  & $(2,24)$  & $(12,4)$ \\
                & 16 & $(7,10)$  & $(2,46)$  & $(23,4)$ \\
                & 32 & $(14,10)$ & $(2,88)$  & $(44,4)$ \\
    \bottomrule
  \end{tabular}
\end{table}

\paragraph{Metric.}
We report \emph{relevant-record coverage}: the number of distinct queried
records that belong to a requested target chain, divided by $Q$. The oracle
computes this quantity from the union of all workers' observations;
correctness of the workers' final submissions is not required for the
coverage metric.

\begin{figure}[tbp]
    \centering
    \includegraphics[width=\linewidth]{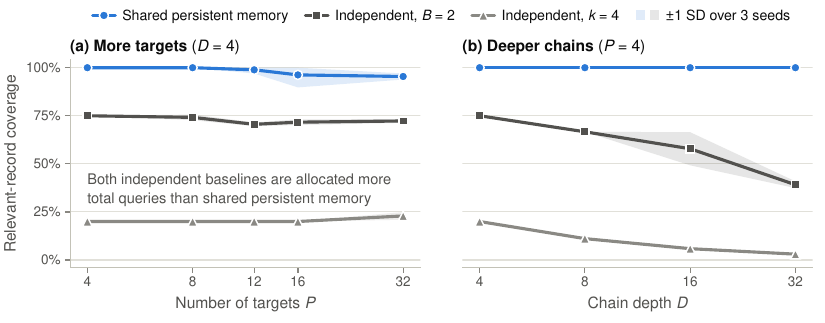}
    \caption{\textbf{Persistence reduces rediscovery.} Relevant-record coverage in Hidden Rulebook as (a) the number of targets $P$ and (b) the chain depth $D$ grow, for shared persistent memory and two independent baselines with larger total query budgets. Mean over three seeds; bands: $\pm1$~SD\@.}
    \label{fig:hidden-rulebook-synthetic}
\end{figure}

\paragraph{Results.}
Figure~\ref{fig:hidden-rulebook-synthetic} shows that shared persistent memory
maintains high relevant-record coverage as the number of targets or the
dependency depth increases. The independent baselines' coverage stays roughly flat but well below shared
memory as the number of targets grows (about 70--75\% for $B=2$ and 20--23\% for
$k=4$) and drops with dependency depth, despite their larger aggregate query
budget. Persistence
therefore reduces the rediscovery cost of extending sequential
investigations.

\subsection{Branchability: Hidden Maze}
\label{app:hidden-maze-details}

\begin{figure}[tbp]
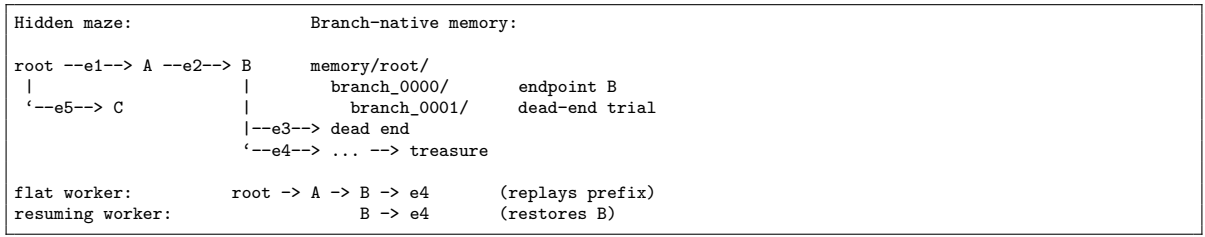

  \centering
  \begin{minipage}{0.94\linewidth}
\begin{lstlisting}[
  basicstyle=\fontfamily{lmtt}\selectfont\scriptsize,
  frame=single,
  columns=fullflexible,
  keepspaces=true
]
Hidden maze:                  Branch-native memory:

root --e1--> A --e2--> B      memory/root/
 |                     |        branch_0000/       endpoint B
 `--e5--> C            |          branch_0001/     dead-end trial
                       |--e3--> dead end
                       `--e4--> ... --> treasure

flat worker:          root -> A -> B -> e4       (replays prefix)
resuming worker:                   B -> e4       (restores B)
\end{lstlisting}
  \end{minipage}
  \caption{\textbf{Hidden Maze example.} A restorable endpoint lets a new
  worker explore another exit from \texttt{B} without paying again for the
  path from the entrance. Nested directories record which endpoint each
  trajectory extended.}
  \label{fig:app-hidden-maze-example}
\end{figure}

\paragraph{Task generation.}
The maze is a randomly generated rooted tree of maximum depth $D$. During
breadth-first generation, each room at depth less than $D$ samples between
zero and $N$ children; we retain trees that reach depth $D$ and place the
treasure at a uniformly sampled deepest leaf. Room and exit IDs are opaque. A
worker never sees the tree globally: its current state reveals only the
current room, its visible exits, the exit returning to the preceding room, and
whether the room is a dead end or contains the treasure. A move through any
visible exit costs one step, including moves that retrace previously explored
routes. Thus a worker may backtrack physically, but doing so also spends its
limited budget (Figure~\ref{fig:app-hidden-maze-example}).

\paragraph{Interface and memory treatments.}
Every worker has a restricted \texttt{bash(command)} tool and a writable
\texttt{current\_state.json}. The call \texttt{move(exit\_id, rationale)}
moves from the room stored in this file, costs one step, and replaces it with
the new room state. \texttt{yield\_worker(summary)} ends the episode. Where a
treatment makes saved states restorable, a checkpoint can be copied to
\texttt{current\_state.json} only before the first move; afterward, the worker
follows one continuous physical trajectory. The harness validates every
restored checkpoint and commits the final state after the episode, so workers
cannot fabricate states or modify previous work. We compare three filesystem
memory organizations:
\begin{itemize}[leftmargin=*,itemsep=2pt]
  \item \textbf{Flat persistent memory} stores all earlier moves in
  \texttt{memory/paths.jsonl}, but every worker starts at the entrance and no
  saved state is restorable, so a worker must physically retrace any known
  prefix.
  \item \textbf{Unstructured resumable memory} stores endpoint files under
  \texttt{memory/nodes/}; a worker can copy one to
  \texttt{current\_state.json}, but its prompt does not mention this
  possibility.
  \item \textbf{Branch-native memory} stores every completed trajectory as a
  child directory beneath the endpoint it resumed, forming a tree of
  checkpoints. Its prompt explicitly asks the worker to inspect the tree,
  select an endpoint, and restore its \texttt{checkpoint.json} before
  continuing exploration. Selecting a leaf continues a branch, while selecting
  an endpoint with existing children creates a sibling branch. Branch-native
  memory thus makes both earlier endpoints and their relationships directly
  available for subsequent exploration.
\end{itemize}
Previous memory is read-only in all treatments.

\paragraph{Experimental settings.}
We sweep maze depth and maximum branching factor; $k$ denotes the move budget
per worker. In the depth sweep, we fix $N=4$ and use $D\in\{4,6,8,10\}$, with
$(B,k)$ equal to $(2,6)$, $(3,9)$, $(4,12)$, and $(4,15)$, respectively. In the
branching-factor sweep, we fix $D=6$, use $N\in\{2,4,6,8\}$, and set
$(B,k)=(3,9)$. In every condition, $k>D$, so an individual worker can reach
the treasure by choosing the correct route; no treatment is made impossible by
its step limit.

\paragraph{Metrics.}
We record treasure discovery, total steps to the treasure, checkpoint
restorations, and the \emph{repeated-move fraction}: the fraction of steps
whose directed room--exit pair was already traversed by an earlier step.

\begin{figure}[tbp]
    \centering
    \includegraphics[width=\linewidth]{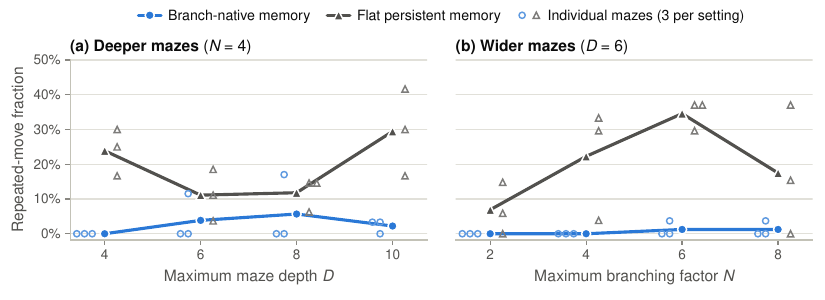}
    \caption{\textbf{Branch-native memory reduces repeated traversal.} Repeated-move fraction in Hidden Maze for branch-native and flat persistent memory as (a) maze depth $D$ and (b) branching factor $N$ grow. Lines: mean over three mazes; open markers: individual mazes.}
    \label{fig:hidden-maze-synthetic}
\end{figure}

\paragraph{Results.}
Figure~\ref{fig:hidden-maze-synthetic} shows that branch-native memory
substantially reduces repeated traversal relative to flat persistent memory
across the tested depth and branching-factor regimes; results for unstructured
resumable memory are not shown. Restoring a previously
reached endpoint allows subsequent workers to explore another branch without
paying again for its known prefix.

\paragraph{Caveat.}
This experiment compares practical memory interfaces rather than changing
branchability alone: the branch-native treatment also makes checkpoint
restoration explicit in the worker prompt.

\subsection{Attribution: Retracted Evidence}
\label{app:retracted-evidence-details}

\paragraph{Task generation.}
The archive contains source notes, derived evidence notes, and candidate
conclusions. A source note states a numerical claim such as ``specimen Iona
has baseline resonance 22.'' A derived note names a specimen and metric,
applies a signed numerical adjustment to one earlier claim, and reports the
result. Its true parent is the unique preceding note whose specimen, metric,
and value make this calculation correct. A conclusion states that one
candidate answers one question and names the claim required to support it.
An artifact is valid only if its producer and every artifact in its ancestry
remain valid. We choose one intermediate artifact to be authored by a
retracted laboratory and attach multiple downstream branches to it; all
conclusions below it become invalid. For each of the $P$ questions, exactly
one of the $C$ candidate conclusions has a fully clean lineage
(Figure~\ref{fig:app-retracted-evidence-example}).

\begin{figure}[tbp]
  \centering
  \begin{minipage}{0.94\linewidth}
\begin{lstlisting}[
  basicstyle=\fontfamily{lmtt}\selectfont\scriptsize,
  frame=single,
  columns=fullflexible,
  keepspaces=true
]
s0: Lab Umber reports baseline resonance = 22
 |
 |-- s1: Lab Amber applies +6, obtaining spectral index = 28
 |    `-- ... --> c1: spectral answer A               [valid]
 |
 `-- p0: Lab Cedar applies +4, obtaining value = 26
      `-- r1: Lab Saffron applies -11, obtaining value = 15
           |             Lab Saffron is RETRACTED
           |-- ... --> c2: spectral answer B          [invalid]
           `-- ... --> c3: binding answer X           [invalid]

Decoys report nearby values (19, 25, 31, ...), but only one value
makes each stated numerical transformation correct.
\end{lstlisting}
  \end{minipage}
  \caption{\textbf{Retracted Evidence example.} Retraction propagates through
  ancestry: every conclusion depending on \texttt{r1} is invalid even when its
  own author is clean.}
  \label{fig:app-retracted-evidence-example}
\end{figure}

\paragraph{Interface and treatments.}
The initial catalog always shows each artifact's ID, type, and title;
conclusion rows also show their question ID and candidate answer. In the
attribution treatment, each row additionally contains its producer and
immediate parent IDs. In the no-attribution treatment, these fields are absent
and workers must recover the lineage from the numerical statements. The prompt
does not name the treatment.

\texttt{read\_artifacts(artifact\_ids, rationale)} returns the full text of the
requested artifacts and charges one read per artifact.
\texttt{submit\_answers(answers)} submits one candidate for every question,
and \texttt{yield\_worker(summary)} ends an episode. Four workers run
sequentially; all artifact texts read by earlier workers and all earlier
submissions remain visible. Thus both treatments have persistent
observations: they differ in whether explicit attribution (dependency)
metadata is available, not in whether their accumulated observations
persist.

\paragraph{Difficulty controls and settings.}
Lineage depth $L$ is the number of ancestry levels ending at a conclusion.
Semantic ambiguity $A$ is the number of plausible parent notes with the
required specimen and metric at a parent-selection step; only one of them has
the required value. Fan-out $F$ is the number of downstream branches below the
retracted intermediate artifact. All conditions use $P=4$ questions, $C=2$
candidates per question, four workers, and three seeds. We sweep one factor
at a time (Table~\ref{tab:retracted-evidence-settings}). The per-worker read
budget is $k=\lceil|\mathcal{A}|/4\rceil$, where $|\mathcal{A}|$ is the
generated archive size, so the combined budget of the four workers is
sufficient to read the complete archive in principle.

\begin{table}[tbp]
  \centering
  \small
  \caption{\textbf{Retracted Evidence settings.} One factor is varied at a
  time. The per-worker artifact-read budget $k$ is set to
  $\lceil |\mathcal{A}|/4\rceil$, where $|\mathcal{A}|$ is the generated
  archive size, so the four workers can read the complete archive in
  principle.}
  \label{tab:retracted-evidence-settings}
  \begin{tabular}{@{}lllll@{}}
    \toprule
    Sweep & Varied values & Fixed values & Archive sizes & $k$ \\
    \midrule
    Depth & $L=3,4,5,6$ & $A=4,\ F=6$ & $54,94,134,174$ & $14,24,34,44$ \\
    Ambiguity & $A=2,4,6,8$ & $L=5,\ F=6$ & $72,134,196,258$ & $18,34,49,65$ \\
    Fan-out & $F=4,6,8,10$ & $L=5,\ A=4$ & $116,134,152,170$ & $29,34,38,43$ \\
    \bottomrule
  \end{tabular}
\end{table}

\paragraph{Metric.}
The primary metric is the fraction of unique archive artifacts read before the
first exact submission. It measures the information-gathering cost of
identifying valid conclusions, and distinguishes the ability to follow
recorded dependencies from the cost of reconstructing those dependencies from
content.

\begin{figure}[tbp]
    \centering
    \includegraphics[width=\linewidth]{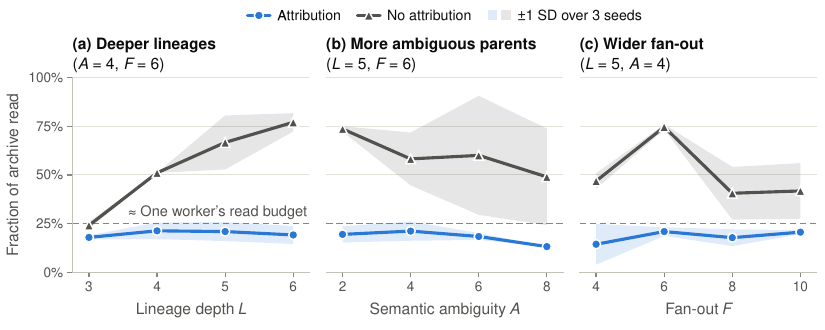}
    \caption{\textbf{Attribution reduces lineage reconstruction.} Fraction of the archive read before the first exact submission in Retracted Evidence, with and without attribution, as (a) lineage depth $L$, (b) semantic ambiguity $A$, and (c) fan-out $F$ grow. Mean over three seeds; bands: $\pm1$~SD; dashed line: about one worker's read budget.}
    \label{fig:retracted-evidence-synthetic}
\end{figure}

\paragraph{Results.}
Figure~\ref{fig:retracted-evidence-synthetic} shows that explicit attribution
reduces the fraction of the archive workers must inspect at every tested
lineage depth, semantic ambiguity, and downstream fan-out level, with the gap
widening with lineage depth. Recorded dependency
metadata allows workers to follow and invalidate affected lineages without
reconstructing every dependency from artifact contents.

\subsection{Scope of the Controlled Tests}
\label{app:synthetic-scope}

These experiments examine three distinct costs: rediscovering prior
observations, retraversing previously explored states, and reconstructing
dependencies between artifacts. They provide controlled illustrations of why
persistence, branchability, and attribution may be useful for compounding 
inference. They do not establish that any individual property causes
\gitswarm{}'s benchmark improvements (Section~\ref{sec:limitations}). Such a
claim would require interventions on the corresponding property within the
full system while controlling for changes in compute, worker behavior, and the
remaining substrate capabilities.

\section{Benchmark Performance Details}
\label{app:benchmark-results}

\begin{figure}[tbp]
    \centering
    \includegraphics[width=\linewidth]{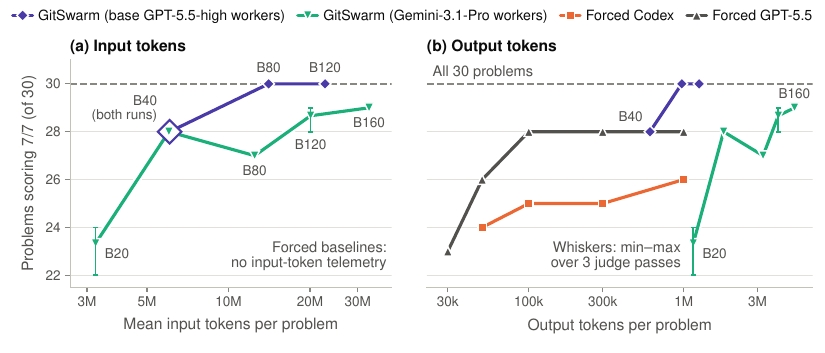}
    \caption{\textbf{IMOProofBench-Advanced proof scaling.} Problems solved (7/7) against mean input (a) and output (b) tokens per problem for \gitswarm{} with Gemini-3.1-Pro and base (tool-calling) GPT-5.5-high workers, and for Forced Codex and Forced GPT-5.5, shown only in (b) for lack of input telemetry. Gemini points: mean of three judge passes.}
    \label{fig:imoproofbench-gemini-scaling}
\end{figure}

This appendix complements Section~\ref{sec:performance} with the
IMOProofBench-Advanced scaling figure
(Figure~\ref{fig:imoproofbench-gemini-scaling}) and additional detail on both
benchmarks. We evaluate both final performance and how performance changes as
inference compute increases. The worker budget $B$ is the controlled
system-level knob, while recorded input and output tokens provide model-level
measures of the computation that was actually consumed. Benchmark protocols,
scoring, and the baseline continuation protocol are described in
Appendices~\ref{app:setup-benchmarks} and~\ref{app:setup-compute}.

\paragraph{IMOProofBench-Advanced.}
Figure~\ref{fig:imoproofbench-gemini-scaling} shows the proof scaling curves
for the plotted configurations. Proof construction follows the same broad
trend as ProgramBench, but with more run-to-run variation and earlier
saturation, whereas program reconstruction continues to benefit from larger
budgets. The
configuration differences among the proof runs, which make these cross-budget
curves descriptive rather than a controlled estimate of the effect of worker
count (Section~\ref{sec:performance}), are listed in
Appendix~\ref{app:setup-analysis-corpus}.

\clearpage
\section{AutoResearch: Additional Details}
\label{app:autoresearch}

This appendix complements Section~\ref{sec:autoresearch} with the common
research setting (Appendix~\ref{app:autoresearch-setup}), the precise task
specifications and compared configurations, full-size hill-climbing figures,
and the detailed research records of the Residual Matrix Transformer
(Appendix~\ref{app:rmt}) and Looped Transformer (Appendix~\ref{app:loop}) runs,
including the Looped Transformer's semantic research graph. It also gives the
NanoChat comparison caveats (Appendix~\ref{app:nanochat}) and observations that
hold across the three tasks (Appendix~\ref{app:autoresearch-findings}).

\subsection{Common Setup}
\label{app:autoresearch-setup}

\paragraph{Research setting.}
In each of the three architecture-improvement tasks, the system is given a
description of an initial transformer architecture and a fixed number of GPUs,
and is instructed to make architectural modifications and hill-climb on a given
metric. The focus is on meaningfully interesting architectural modifications
rather than minor tweaks and hyperparameter changes. Workers start from an
existing implementation and a task-specific research objective, independently
propose modifications, implement candidate designs, run experiments, and use
previous results to guide subsequent work.

\paragraph{Resource coordination.}
Unlike mathematical proofs or program reconstruction, these tasks require
coordinating expensive experiments under limited computational resources.
Workers must coordinate not only on architecture design but also on resource
management: which worker uses how many GPUs to run its experiments, and which
experiments to prioritize when resources are constrained. Workers submit
training and evaluation jobs through a SLURM scheduler and share a fixed GPU
allocation (Section~\ref{sec:autoresearch}), resembling a typical research group
pursuing architecture research. Their code, experimental results, and research
notes persist in the repository, allowing other workers to inspect and extend
previous work.

\paragraph{Evaluation and reporting.}
We evaluate progress through the best objective value attained over research
time and, where available, held-out evaluations of the resulting architectures.
Because the objective, starting architecture, research budget, and comparison
conditions differ across tasks, we report each task separately rather than
aggregating the outcomes into a single research performance score.

\subsection{Residual Matrix Transformer}
\label{app:rmt}

\subsubsection{Task Details}
\label{app:rmt-task}

\paragraph{Architecture.}
The Residual Matrix Transformer (RMT;~\citealp{mak2025rmt}) replaces the
transformer's residual vector with an outer-product memory matrix: each layer
retrieves its input from the matrix through learned read keys and stores its
output back as outer products of learned write keys and values. This lets the
residual stream be scaled independently of parameters and compute.

\paragraph{Task specification.}
Starting from the authors' JAX implementation, the swarm must develop an
architecturally improved RMT that retains a matrix-valued residual stream and has
at most 350M parameters, trained from scratch on a supplied 6.5B-token
OpenWebText stream. The search metric is next-token cross-entropy on the
4,873-example protected development set (512 context tokens, one withheld target
token); the reference is a 305M RMT trained on 1.3B tokens (development loss
3.158). Final held-out losses, such as the reference's 3.147, come from a
separate held-out evaluation.
The task explicitly prioritizes fundamental architectural changes over
hyperparameter or configuration tuning.

\paragraph{Compared configurations.}
Besides the two 12-worker \gitswarm{} configurations with 8 or 4 GPUs, the
single-agent direct GPT and Codex baselines use the same model (GPT-5.6-sol at
xHigh reasoning) and the same 72-hour budget on one eight-H200 node. We also
examine a \gitswarm{} run seeded with earlier incumbents
(Appendix~\ref{app:rmt_path}).
Figure~\ref{fig:autoresearch_hillclimb}(a) shows the research trajectories.

\subsubsection{Final Architectures and Idea Families}
\label{app:rmt_final}

\paragraph{Final architectures.}
Both \gitswarm{} runs improve on the reference within its parameter cap and
training budget. The 8-GPU run's final model combines six mechanisms
(Table~\ref{tab:rmt_mechanisms}): three general transformer changes (removing a
depth-dependent attention scaling, tied embeddings, and a SwiGLU feed-forward
layer) and three that exploit the matrix memory (routing across attention heads,
wider attention, and mixing in the memory states of recent tokens). The 72-hour
submission (dev 2.909, held-out 2.890) uses rank-32 attention; the best model
after the extension (dev 2.894) widens it to rank 64
(Appendix~\ref{app:rmt_path}). The 4-GPU run converged on an overlapping but
simpler design, tied embeddings, SwiGLU, and a deeper 32-layer stack, with a
held-out loss of 2.894.

\paragraph{How the baselines' gains differ.}
The baselines' gains were of a different kind. Direct GPT trained longer (3.0B
tokens) and added an auxiliary output decoder, and Codex reached 2.690 mainly by
training for 10.1B tokens, from which about three quarters of its gain came.
\gitswarm{}'s gains, by contrast, come entirely from architectural changes at the
reference training budget, half of them specific to the RMT's matrix
memory, which is the kind of change the task asks for.

\begin{table}[tbp]
\centering
\small
\caption{Architectural mechanisms in the best \gitswarm{} RMT model (8-GPU run).}
\label{tab:rmt_mechanisms}
\begin{tabularx}{\linewidth}{lX}
\toprule
Mechanism & What it does \\
\midrule
Depth-scaling fix & Removes the reference's division of attention scores by the layer index. \\
Tied embeddings & Shares input and output token embeddings; freed parameters widen the feed-forward layers. \\
SwiGLU feed-forward & Replaces the GELU feed-forward layer with a gated one of equal size. \\
Cross-head router & A zero-initialized learned mixing of attention scores across heads before the softmax. \\
Wider attention & Decouples the number of attention heads from the memory rank and widens it (16$\rightarrow$32$\rightarrow$64). \\
Temporal memory mixer & Adds gated copies of the memory states of the 1, 2, 4 and 8 previous tokens. \\
\bottomrule
\end{tabularx}
\end{table}

\paragraph{Ideas tried.}
Table~\ref{tab:rmt_families} lists the 19 idea families explored by the
\gitswarm{} RMT run over the 72-hour run and its extension, with their outcomes.
The families range from standard transformer components (gated MLPs, embedding
tying, attention normalization, positional schemes, differential attention) to
ideas specific to the matrix memory (routing along its axes, normalizing it,
changing how layers read from and write to it, mixing it over time). Six
families were incorporated into the final model, one led the frontier and was
superseded, two produced gains that did not survive combination with the leader,
two remained inconclusive at full size, seven were refuted, and one consisted of
controls. The single-agent baselines narrowed much earlier: Codex opened no new
idea threads after 26.5 hours, and direct GPT spent 68\% of its GPU time on a
single family.

\paragraph{Negative results.}
The 69 negative commits (58\% of the 120) are ideas that their authors published
as not beating their control or the current best model. They were not
discarded: the roughly 30\% of them later cited as evidence by other workers
compares with 84\% of positive commits.

\begin{table}[tbp]
\centering
\small
\caption{Idea families explored by the \gitswarm{} RMT run: number of commits,
positive/negative commits as labelled by their authors, and outcome. Family
names follow the run logs: the ``cross-rank attention router'' is the cross-head
router of Table~\ref{tab:rmt_mechanisms}, and ``rank widening'' corresponds to
wider attention. The family assignment covers 119 of the 120 commits, and its positive/negative
split (48/71) differs slightly from the 69 negative commits reported in
Section~\ref{sec:autoresearch}.}
\label{tab:rmt_families}
\begin{tabularx}{\linewidth}{@{}>{\raggedright\arraybackslash}p{3.9cm}cc>{\raggedright\arraybackslash}p{2.4cm}>{\raggedright\arraybackslash}X@{}}
\toprule
Family & Commits & +/$-$ & Outcome & Note \\
\midrule
Attention depth-scaling fix & 4 & 3/1 & In final model & adopted in final model \\
Tied embeddings / token interface & 7 & 3/4 & In final model & adopted in final model \\
Gated MLP (SwiGLU) variants & 7 & 4/3 & In final model & adopted in final model \\
Cross-rank attention router & 6 & 5/1 & In final model & adopted in final model \\
Attention (and MLP) rank widening & 9 & 7/2 & In final model & adopted; widened to rank 64 \\
Multi-lag causal matrix mixing & 13 & 2/11 & In final model & adopted; 9 later variants rejected \\
Differential attention & 8 & 3/5 & Led, then superseded & led twice, then superseded \\
Matrix-state normalization & 8 & 5/3 & Stalled (lost when combined) & 305M gain lost on leader \\
Matrix-state capacity & 4 & 2/2 & Stalled (lost when combined) & 305M gain lost on leader \\
Post-softmax / two-stage routing & 5 & 3/2 & Inconclusive & missed 305M leader by 0.0006 \\
Depth reallocation & 5 & 3/2 & Inconclusive & no 305M result yet \\
Matrix-axis routing & 8 & 3/5 & Refuted & failed 305M transfer twice \\
Memory read/write addressing & 7 & 1/6 & Refuted & gain did not transfer to 305M \\
Attention-value routing & 5 & 1/4 & Refuted & gain did not transfer to 305M \\
Q/K normalization and metrics & 7 & 1/6 & Refuted & 46M gain reversed at 305M \\
Residual-write gating and scaling & 4 & 1/3 & Refuted & 305M run dominated, cancelled \\
Positional biases and writes & 6 & 1/5 & Refuted & no gain on depth-fixed base \\
Matrix-update form & 5 & 0/5 & Refuted & no gain at 46M (0/5) \\
Init, horizon and seed controls & 1 & 0/1 & Control & seed spread $\sim$0.010 nats \\
\bottomrule
\end{tabularx}
\end{table}

\subsubsection{Reuse and Cross-Lineage Inheritance}
\label{app:rmt_reuse}

\paragraph{Concentrated reuse.}
Reuse was highly concentrated. The first durable improvement, a correction to
the reference code found within the first hour, is an ancestor of 102 of the 120
commits, and ten commits receive half of all citations.

\paragraph{Reuse of a failure.}
The most consequential reuse was of a failure. A worker tried letting each token
see the previous token's memory state, and it failed. 49.5 hours later, another
worker cited this failure as evidence that longer look-backs were still
unexplored and built a mixer over several earlier tokens that succeeded; a third
worker carried it into the 72-hour submission.

\paragraph{Cross-lineage inheritance.}
Although the final model's code descends through a single chain of eight commits,
only two of its six mechanisms were first written on that chain. The other four
(SwiGLU, the head router, wider attention, and the temporal mixer) were developed
on side branches by other workers and ported into the chain
(Table~\ref{tab:rmt_inheritance}). Other branches contributed evidence without
code, for example controls used to measure a change, or a competing design that a
later step had to beat. Counting this cited evidence yields the 24-commit,
24-worker dependency set reported in Section~\ref{sec:autoresearch}.

\begin{table}[tbp]
\centering
\small
\caption{Where the final RMT model's mechanisms were first written, and how they
reached the solution path.}
\label{tab:rmt_inheritance}
\begin{tabularx}{\linewidth}{lX}
\toprule
Mechanism & Origin \\
\midrule
Depth-scaling fix & Written on the solution path (0.6\,h). \\
Tied embeddings & Written on the solution path (1.8\,h). \\
SwiGLU feed-forward & Side branch (0.8\,h); ported into the path at 5.9\,h. \\
Cross-head router & Side branch (5.2\,h), validated at full size at 15.1\,h; ported at 34.6\,h. \\
Wider attention & Side branch (8.2\,h), validated at full size at 23.6\,h; ported at 53.1\,h. \\
Temporal memory mixer & Side branch (50.8\,h), generalizing a refuted single-token version (1.2\,h); ported at 65.9\,h. \\
\bottomrule
\end{tabularx}
\end{table}

\subsubsection{Solution Path and Cross-Pollination Run}
\label{app:rmt_path}

\paragraph{Semantic research graph.}
Figure~\ref{fig:autoresearch_hillclimb}(b) condenses the run's commit history into
a semantic graph of 69 nodes. Research steps are ordered by time. The thick green
line is the lineage of the best model; blue results held only on small
($\sim$46M) proxies and green results at full size (305M)\@. Red and orange nodes
are groups of attempts tested on top of the leader of the time and refuted or
stalled, respectively; grey edges show reused code or evidence, and the purple
edge marks a commit that generalizes an earlier failure. Ringed nodes are the
72-hour submission and the best model after the extension.

\paragraph{How the final architecture emerged.}
Early workers screened many ideas in parallel on small proxy models. The eight
commits on the lineage of the best model were:
\begin{itemize}
\item \textbf{Depth fix (0.6\,h, small proxy).} Removed a scaling quirk in the
reference code. Nine attempts made while it was the leader were refuted
(positional schemes, attention normalization, memory addressing and routing).
\item \textbf{Tied embeddings (1.8\,h, proxy).} Freed parameters for the
feed-forward layers. Write gating and alternative update rules tried while it
led failed.
\item \textbf{+ SwiGLU (5.9\,h, proxy).} Ported from a parallel branch; memory
normalization and depth reallocation tried on it stalled.
\item \textbf{First full-size model (11.8\,h, loss 2.962).} The combination
validated at 305M\@. It attracted the largest wave of experiments: ten families
were tried on it and failed or stalled, and a side branch using differential
attention briefly overtook it.
\item \textbf{Head router (34.6\,h, 2.934).} Developed on a side branch; it overtook
the differential-attention line, which led once more (50.7\,h) before being
superseded.
\item \textbf{Wider attention (53.1\,h, 2.924).} Also ported from a side branch.
\item \textbf{Temporal memory mixer (65.9\,h, 2.909; held-out 2.890).} The
72-hour submission, built from the reused single-token failure described above.
Eight variations of the mixer tried on top of it were all rejected.
\item \textbf{Even wider attention (98.7\,h, 2.894).} The best model after the extension.
\end{itemize}

\paragraph{Cross-pollination run.}
The seeded run of Section~\ref{sec:autoresearch} is a new 72-hour \gitswarm{} run
started from the best models of three completed runs (about 220 prior
research-hours in total): the 8-GPU run above (2.909, with the temporal mixer),
the 4-GPU run (2.926, with a deeper stack), and a GPT-5.5 High run (2.989). Its
gains came from combining the seeds: after 26.7 hours it added the 8-GPU run's
head router to the 4-GPU model (2.905), and an hour later it moved the 4-GPU
run's deeper stack onto the 8-GPU model (2.898). The GPT-5.5 seed was used only
indirectly, through a negative result that cited it. The run was heavily
rate-limited by a shared API quota, so its pace is not comparable to the original
runs.

\subsubsection{Findings about the Research Process}
\label{app:rmt_findings}
\begin{itemize}
\item \emph{Small-scale screening was informative early and misleading late.}
During the 72-hour run, 15 of 17 ideas promoted from the small proxy to full
size helped; in the extension, only 1 of 8 did, and several ideas reversed sign
at full size.
\item \emph{Gains measured in isolation often vanished when combined with the
leader.} For example, a memory normalization that helped the plain model hurt
the best model.
\item \emph{Noise matters.} A seed replication moved loss by 0.010, comparable to
several of the later steps on the solution path.
\end{itemize}

\subsection{Looped Transformer}
\label{app:loop}

\subsubsection{Task Details}
\label{app:loop-task}

\paragraph{Architecture.}
The Looped Transformer task uses Huginn~\citep{geiping2025recurrent}, a
depth-recurrent language model that splits the network into a prelude, a
weight-tied recurrent core, and a coda. The prelude embeds the input, the core is
iterated a variable number of times with the prelude output re-injected at every
iteration---an adapter mixes the current state with the encoded input---and the
coda decodes the final latent state, so test-time compute can be scaled by
unrolling more iterations instead of generating more tokens.

\paragraph{Task specification.}
Starting from the authors' official code, scaled to a $\sim$308M-parameter
reference (1,024-token context, eight recurrences), the swarm must develop an
architecturally improved model that keeps genuine weight-tied recurrent-depth
computation, has at most 350M unique parameters, and is trained from scratch with
the reference's token budget on the same 6.5B-token OpenWebText stream as the RMT
task (Appendix~\ref{app:rmt-task}). The metric is the withheld next-token loss on
the 4,873 protected-development 1,024-token contexts, evaluated at exactly
eight recurrences (reference 3.265); we do not yet report a held-out
evaluation for this task.

\paragraph{Protocol and compared configurations.}
Candidates are developed on a ladder of smaller models (52M, 124M, 188M) before
full-size training. \gitswarm{} (12 workers), direct GPT and Codex all use
GPT-5.5 High with a 48-hour budget on one eight-H200 node; the runs were still in
progress at the time of writing (26 of 48 hours).
Figure~\ref{fig:autoresearch_loop}(a) shows the research trajectories.

\subsubsection{Final Architecture and Idea Families}

\paragraph{Final architecture.}
The best model differs from the reference in one place: the linear adapter that
re-injects the input at every iteration is replaced by a small gated feed-forward
network. Because this adapter runs at every recurrence, it is the component most
specific to looped models, and the change adds 27M parameters (8.7\%). The
parameter-matched control of Section~\ref{sec:autoresearch}, trained with the
reference's size, checks that the gain is not simply extra capacity. The
baselines instead added signals to the recurrent state (a learned per-iteration
embedding for direct GPT; input anchors and step signals for Codex), which helped
direct GPT on smaller models but not at full size; neither baseline has yet
beaten the reference.

\paragraph{Ideas tried.}
Table~\ref{tab:loop_families} lists the twelve idea families of the 395
architectural experiments made by the GPT-5.5 \gitswarm{} run in its first 26
hours, with their outcomes: variants of the adapter, a gate on the recurrent
state, attention to the input, parallel specialist feed-forward branches,
iteration-dependent control, extra memory, and changes to the core, readout and
objective. \gitswarm{} used 1,222 GPU jobs and all 64 permitted protected
evaluations, against 324 and 222 jobs for direct GPT and Codex.

\paragraph{Negative results.}
The 170 negative commits amount to 43\% of the architectural ones. Two refuted
results, a narrower adapter and the recurrent-state gate at 188M, were reused to
build the parameter-matched controls.

\begin{table}[tbp]
\centering
\small
\caption{Idea families explored by the GPT-5.5 \gitswarm{} Looped Transformer run
in its first 26 hours: architectural commits, commits published as negative by
their authors, protected evaluations (128, 512 or 4,873 examples), and outcome.
Two of the 64 protected evaluations are not assigned to an architectural family.}
\label{tab:loop_families}
\begin{tabularx}{\linewidth}{@{}>{\raggedright\arraybackslash}p{4.6cm}ccc>{\raggedright\arraybackslash}X@{}}
\toprule
Family & Commits & Neg. & Evals & Outcome \\
\midrule
Feed-forward re-injection adapter & 13 & 0 & 20 & In final model \\
Adapter capacity and placement variants & 37 & 27 & 0 & Refuted \\
Gate on the recurrent state & 32 & 15 & 10 & Early leader; gain vanished at 124M/188M \\
Variants of the re-injected signal & 43 & 23 & 1 & Refuted \\
Attention to the input / first core block & 58 & 18 & 11 & Early versions refuted; later ones in progress \\
Parallel specialist feed-forward branches & 66 & 17 & 13 & Beat the adapter at 52M; stalled at larger sizes \\
Iteration-dependent control & 35 & 15 & 2 & Refuted (128-example win lost at 512) \\
Extra memory or recurrent state & 38 & 14 & 1 & Refuted \\
Core topology and damping & 26 & 14 & 4 & Refuted \\
Readout, coda and feedback & 18 & 13 & 0 & Refuted \\
State initialization and normalization & 19 & 9 & 0 & Refuted \\
Objective, sampling and backpropagation depth & 10 & 5 & 0 & Refuted \\
\bottomrule
\end{tabularx}
\end{table}

\subsubsection{Steps on the Solution Path}
\label{app:loop_path}

\paragraph{How the final architecture emerged.}
Workers first screened all twelve families in parallel on small models; within
two hours the recurrent-state gate and the feed-forward adapter were the two
leading ideas. After the adapter's initial rejection was reversed (see the steps
below), the two were carried up the ladder side by side: the gate's advantage
vanished at 124M and 188M, while the adapter won at every rung and reached 3.187
at full size after 9 hours (Figure~\ref{fig:autoresearch_loop}(b)).
Workers then ran capacity controls and built second-generation mechanisms on top
of the adapter, such as attention to the input and parallel feed-forward
branches, several of which beat it on small models but not yet at full size.
Nothing has improved on the best model since 9 hours; the later shift to
calibration and diagnostics is described in Appendix~\ref{app:loop_findings}.

\paragraph{Cross-lineage inheritance.}
Unlike in the RMT run, the final model's code comes almost entirely from one
lineage of six commits by six workers; only the first changes code, and the only
code imported from elsewhere is a loss-computation fix written for another
branch. Other lineages contributed evidence instead: the rival gate's full-set
scores defined what the adapter had to beat, and a shared tool for fast proxy
measurements (the second most-cited commit) was used to compare candidates.
Counting the commits the best model cites as evidence through \texttt{informed\_by=}
edges, its declared dependencies span 89 commits by 89 distinct workers (98 commits, made
between 0.18 and 9.10 hours, if mistyped commit hashes are recovered by unique
prefix).

\paragraph{Steps.}
The six commits on the lineage of the best model were:
\begin{itemize}
\item \textbf{Feed-forward adapter (0.4\,h, 52M, short run).} Written within the
first 25 minutes, it replaces the linear re-injection adapter with a gated
feed-forward network and imports a loss fix from another branch.
\item \textbf{Longer run (0.7\,h).} The same model trained four times longer.
\item \textbf{Exact 52M rung (2.1\,h).} A quick 128-example check scored it
slightly worse than the reference; its author wrote that it ``should not be
promoted'', and the idea was nearly dropped.
\item \textbf{Re-check (3.3\,h).} About an hour later, another worker re-evaluated
the same checkpoint on 512 examples and found it clearly better ($-0.067$ against
the reference); a third confirmed the gain on the full development set and showed
that it also beat the leading alternative, the recurrent-state gate.
\item \textbf{124M rung (5.8\,h, $-0.055$ against the reference).}
\item \textbf{308M, full size (9.1\,h, 3.187; reference 3.265).} Its commit also
cites a side-branch 188M result ($-0.111$) that finished while it was training. A
parameter-matched control trained later reached 3.212.
\end{itemize}

\subsubsection{Findings about the Research Process}
\label{app:loop_findings}
\begin{itemize}
\item \emph{Small evaluations misranked models.} 128-example evaluations disagreed
with the full set on 5 of 12 comparable pairs, and with 512-example evaluations on
15 of 50; a worker measured this itself and the swarm adopted a
128$\rightarrow$512$\rightarrow$4,873 evaluation funnel.
\item \emph{Gains often failed to transfer across scale.} The recurrent-state gate,
the specialist branches and iteration-dependent control all lost their small-scale
advantage at larger rungs or against a parameter-matched control.
\item \emph{Workers checked each other.} Besides the reversal of the adapter's
initial rejection (Appendix~\ref{app:loop_path}), a later worker overturned a
128-example win for iteration-dependent control, and four workers independently
ran parameter-matched controls.
\item \emph{Shared protocols emerged.} A fixed set of proxy contexts was used by 257
commits and parameter matching was discussed in 187; commits that built measurement
tools and controls were among the most cited.
\item \emph{Work converged, then shifted to diagnostics.} The share of architectural
commits building on the adapter rose from 17\% in the first three hours to about
90\% after nine, which one worker flagged as over-concentration. After the
evaluation budget ran out (20.8 hours), most commits were calibrations and
diagnostics, and 128
of 136 abstentions cited duplicate or already-running work.
\item \emph{No off-task drift} was observed in this run.
\end{itemize}

\subsection{NanoChat}
\label{app:nanochat}

\paragraph{Task and protocol.}
Starting from the original NanoChat/autoresearch implementation, workers
optimize validation bits per byte (BPB; lower is better), training each
candidate from scratch on one GPU under the supplied five-minute training
clock and fixed evaluation. We analyze \gitswarm{} job using
GPT-6 Astra at high reasoning with 20 live workers sharing 16 GB300 GPUs.
Unlike the RMT task, this task does not fix parameter count or training-token
exposure: architecture, model capacity, optimization, and training throughput
are all research variables. We distinguish the first 24 hours of active
allocation from the longer record, which spans 66.84 allocated hours and
78.48 elapsed wall-clock hours because of requeues. Active time includes
startup, compilation, and API waiting while allocated; it excludes gaps
between allocations.

\paragraph{Results.}
Within the first 24 active hours, the run improves from 0.992692 to 0.905235
BPB (8.81\%), completing 1,465 successful evaluations of 1,219 distinct
\texttt{train.py} solutions, plus 175 failed evaluations. The incumbent
was measured at 23.67 active hours, or 25.15 elapsed wall hours; a strict
24-wall-hour cutoff instead gives 0.905487. The complete record contains
3,630 successful and 453 failed evaluations, with 2,784 distinct successful
solutions. Its best observation is 0.898671 (9.47\% below the reference),
but this is an extension result, not a 24-hour result or a separately
held-out test score.

\paragraph{Architecture at the 24-hour cutoff.}
The incumbent retains eight transformer layers but widens the residual
stream from 512 to 768 dimensions, uses eight 96-dimensional attention heads,
and shortens local attention windows from 1,024 to 512 tokens while retaining
full-context layers. It reallocates feed-forward capacity from uniform
fourfold expansion to fourfold expansion in the first four layers and
eightfold expansion in the last four. A new 2,097,152-row table supplies
hashed causal bigram and trigram embeddings, sharing storage between the
two feature types. Query/key normalization moves before rotary embedding.
These changes are accompanied by activation scaling, optimizer changes and
a custom training-head implementation. The resulting model has 1,723.9M
parameters versus 50.3M initially; 1,610.6M belong to the lexical table.
It trains on 547.9M reported tokens versus the reference's 728.2M.
Thus this is substantive architecture and capacity search, but not an
architecture-only, equal-parameter or equal-token comparison.

\paragraph{Ideas tried.}
The record contains measured experiments in width/depth allocation,
attention head geometry and grouped-query attention, local/global context,
gated feed-forward layers, layerwise capacity allocation, normalization and
positional schemes, residual topology, hashed lexical representations,
embedding sharing, value-injection routing, output bottlenecks and auxiliary
supervision. Workers also investigate compiler choices, streamed
cross-entropy gradients, memory ownership, optimizer state and mixed
precision. This inventory is a selected semantic review rather than an
exhaustive partition of every commit. Two controlled comparisons illustrate
its architectural content. First, a SwiGLU change that helped at width 512
failed when transferred to width 768: 0.958091 versus 0.951289 for the
parameter-matched squared-ReLU parent. Second, at equal total MLP parameter
count, early-fourfold/late-eightfold expansion scored 0.932288 versus
0.934069/0.934041 for uniform sixfold expansion; reversing the allocation
scored 0.935398. These comparisons retain initialization, optimizer-shape
and fixed-time-throughput confounds, but go beyond scalar hyperparameter
tuning.

\paragraph{Negative results.}
In the first 24 active hours, worker traces confirm 1,258 accepted research
commit actions. Of these, 709 (56.4\%) are explicitly labelled negative by
their authors, and 497 (70.1\%) are cited by a later accepted research commit
within the same window. Across the complete record, 1,285 of 2,646 accepted
research commit actions (48.6\%) are negative-labelled, and 982 (76.4\%) are
subsequently cited. These are published negative
conclusions, not failed GPU evaluations; their complement includes
inconclusive results and controls, not only positive findings. A citation
records declared code or evidence reuse, not necessarily implementation
inheritance or causal importance.

\paragraph{Reuse of failure.}
The most interpretable chain concerns trigram features. At 2.18 active hours,
worker w0273 added a separate 32K-row trigram table and obtained 0.929923,
worse than both bigram-only controls (0.929131/0.929700). Enlarging the
separate table also failed: 0.924527 versus 0.919483/0.920211. At 23.08
hours, worker w2985 explicitly cited both failures and reused their causal
hash while borrowing a shared-table idea from a lagged-pair branch. This
avoided an additional large table and its optimizer state. The first
shared-table result, 0.905954, did not establish a gain over its strong
parent. Nevertheless, worker w3052 enlarged that shared table to produce
the 24-hour incumbent, 0.905235. After the cutoff, a lexical cooldown/decay
combination reached 0.903842. This is documented reconsideration of a failed
design, not proof that trigram features alone caused the eventual gain:
the surrounding model, capacity and optimizer also changed.

\paragraph{Cross-lineage inheritance.}
The 24-hour incumbent's physical lineage contains 32 research-result
commits, but several mechanisms were developed off that chain. The
96-dimensional head setting and 512-token window came from side branches.
The back-loaded MLP was introduced by w0220 and transferred through a
smaller-table experiment before w0309 combined it with a 524K-row bigram
model. Pre-RoPE query/key normalization came from w0580 and was incorporated
by w0678 into an activation-scaled model. The trigram hash and shared-table
idea came from separate experiments described above. Source-level checks
confirm exact MLP/Block and attention-class transfers for the selected
ports, rather than relying solely on matching prose descriptions.

\paragraph{Replication revised a design decision.}
The following episodes document experimental judgment within the first
24 active hours, including contributions that did not improve the incumbent.
An explicit query/key-normalization epsilon initially scored 0.911027 BPB;
worker w0555 repeated the identical source and obtained 0.911918, revising
the apparent advantage to an unresolved near-tie with a preference for the
simpler inherited normalization. Worker w0580 explicitly cited that failed
replication, retained the original epsilon, and instead moved query/key
normalization before RoPE, obtaining 0.910114. That ordering subsequently
entered the combined architecture. Thus, the earlier evidence informed
both a revised conclusion and a later implementation decision; it did not
establish that the epsilon change was intrinsically harmful.

\paragraph{Experimental-order verification.}
An earlier comparison tested uniform attention-query scaling first
(0.906862 BPB), then depth-dependent scaling (0.907103). Worker w1407
investigated whether order could explain this difference by predeclaring
the reverse sequence: depth-dependent scaling first (0.908089), then
uniform scaling (0.907224), using the unchanged sources. Uniform scaling
again won the pair. The worker retained the unfavorable observation and
distinguished these paired comparisons from historical aggregate means,
which still marginally favored depth-dependent scaling. This investigated
an alternative explanation for prior results rather than merely repeating
the preferred configuration. It was not a randomized crossover study:
training-token exposure, compiler state and measurement time remained
uncontrolled, so it did not identify an order effect or establish expected
architectural superiority.

\paragraph{Verification supplied reusable controls.}
The same verification worker assembled ten measurements of the
byte-identical ordinary, uniform-scaling source, without introducing a new
architecture. Worker w1456 explicitly reused this comparison set when
evaluating a streamed output-head implementation. Its 0.906886 BPB beat
nine observations but not the best, 0.906862; the worker therefore did not
claim a new quality record or a repeatable BPB gain. Instead, it identified
the clearer observed benefit as a reduction in whole-run peak allocation
from 88,331.6 to 84,233.8 reported MB (4.64\%). This was not an isolated
training-only memory profile. The verification contribution supplied
experimental context for interpreting a later systems change, rather than
code or a new incumbent.

\paragraph{Consolidation included simplification.}
Worker w0642 questioned an inherited MLP multiplier, 0.56776,
whose rationale assumed an activation distribution rather than measuring
the learned one. Replacing it with 0.5 scored 0.909838 BPB, which the worker
described as an effective near-tie, not established superiority. Worker
w0678 retained the simpler coefficient and combined it with the separately
developed pre-RoPE normalization, obtaining 0.908277. The half-amplitude
coefficient remained in the 24-hour incumbent. This documents measured
consolidation rather than treating every inherited detail as indispensable;
the synthesis does not isolate a coefficient-specific benefit. Simplicity
was an explicit task preference, not an independently emergent objective.

\paragraph{Shared activity prevented a duplicate evaluation.}
Worker w0033 discovered that w0032 had already announced the identical
depth-10, head-dimension-64 experiment from the same source. It abstained
before training; the evaluation ledger contains no evaluation by w0033.
This episode shows shared in-flight state changing whether an experiment
occurred at all, but does not establish that coordination was universally
efficient or quantify aggregate resource savings.

\paragraph{Interpretation of these behaviors.}
These episodes show observed reuse and revision of experimental evidence,
including contributions that produced neither a new architecture nor a
best score. This run explicitly assigned experimenter/synthesizer and
critic/verifier roles at a configured 3:1 ratio: it supports adaptive work
selection within those roles, not specialization without predefined roles.
The traces document dependencies in experimental judgment, which need not
be code-inheritance edges. They do not establish that shared memory caused
the performance gain or identify each reused idea's contribution; those
claims require an appropriate comparison or ablation.

\paragraph{How the architecture emerged.}
Figure~\ref{fig:autoresearch_nanochat}(b) condenses these developments. Early
parallel tests establish narrower attention heads and wider residual states;
head geometry, batch size, decay, attention sharpness and locality are then
combined. Hashed bigrams open a second capacity axis, and their combination
with back-loaded MLPs and autotuning reaches 0.912586 by 3.00 active hours.
Later work transfers normalization order, investigates readout and gradient
implementations, and revisits lexical context using shared trigrams. The
selected path includes locally inconclusive and memory-oriented steps,
not just successive best-score observations. This explains how the final
combination can emerge through measured branches without assuming that
every isolated improvement transfers additively.

\paragraph{Extension and limitations.}
The extension further changes value routing, lexical gradients and FP8
training arithmetic. Worker w7352 combines expansion-forward FP8 with a
side-branch E4M3 fast input-gradient implementation, producing the lowest
observed 0.898671. However, its exact-source repeat scores 0.899864, and
its worker publication is rejected for an extra audit Markdown file outside
the allowed paths. The code remains reachable in Git, but there is no
located completed finalizer submission. The best source associated with
an accepted commit action measures 0.899233. We therefore describe the
extension's endpoint as a best observation rather than a verified final
architecture. The original timing/evaluation code is retained apart from
an optimizer learning-rate expression, but 701 successful harness records
exceed the original instructions' separate ten-minute overall timeout.
No architecture-only attribution, calibrated confidence interval, or
blanket semantic certification of all attempted programs is claimed.

\subsection{Findings Across Research Tasks}
\label{app:autoresearch-findings}

The three studies extend the evaluation from benchmarks with well-defined target
solutions to multi-hour research tasks requiring architectural decisions,
experimental feedback, and shared GPU resources, and they provide complementary
evidence: RMT includes completed held-out evaluations of improved candidates;
the Looped Transformer shows improvements during the recorded development
trajectories; and NanoChat examines research from multiple starting points under
different resource allocations.

An improving research trajectory does not by itself establish how much of that
improvement comes from access to previous workers' discoveries, nor does
improving a research objective reveal which earlier experiments informed
subsequent research decisions. The repository
provides a mechanism for preserving experimental artifacts and coordinating
subsequent work, but attributing a specific architectural improvement to earlier
contributions requires analysis of the corresponding execution traces and
experimental dependencies (see also the discussion of causal attribution in
Section~\ref{sec:limitations}).

\section{Reuse, Artifacts, and Accumulation}
\label{app:reuse-accumulation}

This appendix supports Section~\ref{sec:worker-reuse}. It defines the reuse and ancestry
statistics and shows a representative ancestry graph (Appendix~\ref{app:reuse-details}),
documents how persistent artifacts were categorized, counted, and consumed
(Appendix~\ref{app:artifact-analysis}), and describes representative collaboration workflows
(Appendix~\ref{app:workflow-examples}).

\subsection{Declared Reuse and Selected-Solution Ancestry}
\label{app:reuse-details}

This subsection gives additional detail on the declared-dependency and
selected-solution ancestry measurements of
Section~\ref{sec:transitive-accumulation}. The per-budget coverage is shown in
Figure~\ref{fig:winner-ancestry-coverage}.

\paragraph{Selected-solution ancestry.}
A worker may depend on only a few immediate sources while indirectly
inheriting work from many earlier workers. Each contribution records its Git
parent and any additional commits named in its \texttt{informed\_by=} field
(Section~\ref{sec:provenance}). Recursively following both kinds of edge from
the selected solution yields the part of the contribution graph that was
carried into that solution, rather than merely produced somewhere during the
run; the task root and nomination records are excluded. On ProgramBench this
ancestry covers most of the contribution graph
(Figure~\ref{fig:winner-ancestry-coverage}), so repeated local composition
from a few sources produces a broad final ancestry; selected proofs retain a
much smaller share.

\paragraph{Coverage by budget.}
Figure~\ref{fig:winner-ancestry-coverage} reports, for each budget
(B100--B500 on ProgramBench, B20--B160 on IMOProofBench-Advanced), the share
of the substantive commit graph covered by the selected winner's Git-parent
ancestry and by its expanded \texttt{informed\_by=} ancestry; bars are averaged across tasks. Neither benchmark shows a consistent increase in coverage with
budget. In both benchmarks, Git parentage alone accounts for roughly one
quarter of the graph: with up to five concurrent workers per problem, several
physical branches advance at once, so a single Git-parent chain cannot contain
most contributions. The difference between the benchmarks therefore lies in
cross-branch \texttt{informed\_by=} synthesis (Section~\ref{sec:transitive-accumulation}).

\paragraph{Interpretation.}
Low proof ancestry does not imply that the other proof work is useless:
alternative proofs, critiques, and validation can affect which candidate is
selected without becoming an ancestor of that exact commit. Persistent state
supports both synthesis and selection, while winner ancestry measures only the
former. Local reuse and selected ancestry must therefore be measured
separately to capture both behaviors.

\subsection{Persistent Artifact Production and Consumption}
\label{app:artifact-analysis}\label{app:voluntary-artifacts}

\begin{figure}[tbp]
    \centering
    \includegraphics[width=\linewidth]{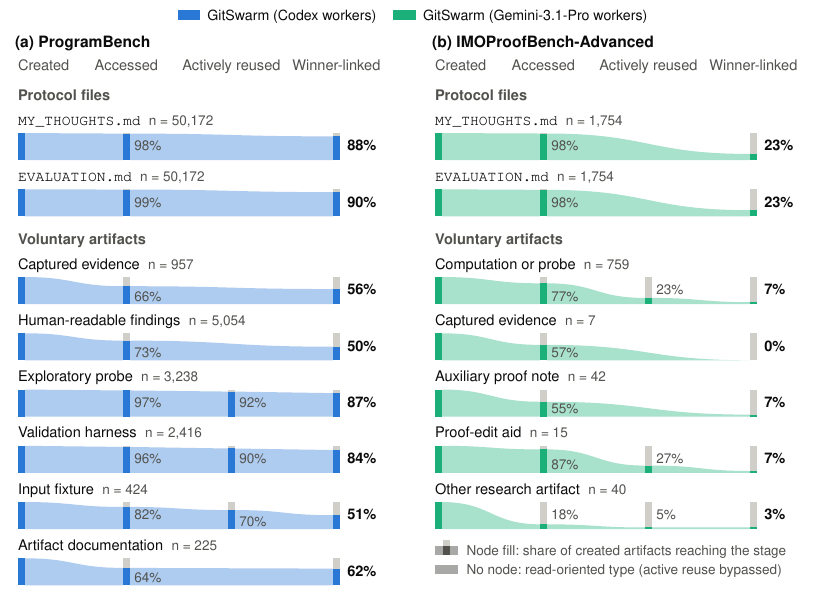}
    \caption{\textbf{Artifact flow from creation to reuse.} For each artifact type on (a) ProgramBench and (b) IMOProofBench-Advanced, the share of created artifacts that are accessed, actively reused, and linked to the selected winner (bold), pooled over budgets. Rows without an Actively reused node skip that stage.}
    \label{fig:artifact-flow-reuse-main}
\end{figure}

Workers coordinate through both protocol-defined files and voluntarily created
artifacts. The prompt asks workers to leave useful intermediate artifacts in
their commits. The voluntary artifacts that workers produce include tests,
executable probes, captured outputs, reports, auxiliary proof notes, and
documentation. These files act as persistent
communication between episodes: a later worker can inspect a finding, rerun a
probe, or adapt a test without reconstructing the producer's private tool
transcript. Because artifact creation alone does not establish collaboration,
we distinguish production, subsequent access, active reuse, and linkage to the
selected solution. Section~\ref{sec:worker-reuse} reports the headline
consumption rates; this subsection gives the complete per-category
artifact-flow measurements (Figure~\ref{fig:artifact-flow-reuse-main}) and
documents how artifacts were categorized and counted, how the consumption
stages are defined, and how production varies across tasks and budgets.

\paragraph{Artifact categories.}
We derived the voluntary artifact categories separately for each benchmark. We
first enumerated files added relative to each worker's physical parent,
excluding task inputs, candidate runtime files, and protocol-required files. We
then reviewed paths, contents, diffs, commit metadata, and declared dependencies,
summarized each file's role, and grouped recurring roles by hand. This yielded
six common ProgramBench categories and four recurring proof categories plus a
small residual category (Table~\ref{tab:artifact-category-definitions}). The
proof labels were derived from the proof traces rather than transferred from
ProgramBench.

\begin{table}[tbp]
  \centering
  \scriptsize
  \caption{\textbf{Benchmark-specific voluntary artifact categories.} The
  categories were derived separately from files newly created by workers in
  each benchmark.}
  \label{tab:artifact-category-definitions}
  \begin{tabularx}{\linewidth}{@{}llX@{}}
    \toprule
    \textbf{Benchmark} & \textbf{Category} & \textbf{Role} \\
    \midrule
    ProgramBench & Captured evidence & Raw outputs, logs, diffs, snapshots, or transcripts from executions. \\
    ProgramBench & Human-readable findings & Prose that records an observed behavior, diagnosis, or validation result. \\
    ProgramBench & Exploratory probe & Executable experiment used to discover unknown behavior. \\
    ProgramBench & Validation harness & Repeatable executable check for an understood behavior or implementation. \\
    ProgramBench & Input fixture & Input or corpus created to exercise candidate behavior. \\
    ProgramBench & Artifact documentation & Index or instructions for using an artifact collection. \\
    \midrule
    IMOProofBench & Computation or probe & Script or notebook used for symbolic, numerical, or combinatorial checks. \\
    IMOProofBench & Captured evidence & Stored output from a proof-related computation or external check. \\
    IMOProofBench & Auxiliary proof note & Informal derivation, lemma, case analysis, or critique outside the submitted proof. \\
    IMOProofBench & Proof-edit aid & Helper used to transform, compare, or inspect proof text. \\
    IMOProofBench & Other research artifact & Rare research file not covered by the four recurring proof categories. \\
    \bottomrule
  \end{tabularx}
\end{table}

\paragraph{Logical bundles.}
Raw file counts are highly skewed. Some ProgramBench workers committed
thousands of related output, error, status, and diff files from one
experiment. We therefore count files that were created together in one
artifact directory and serve the same role as one logical bundle when the
files have no useful meaning on their own. The resulting counts describe
reusable units rather than the serialization format chosen by a worker.

\begin{figure}[tbp]
    \centering
    \includegraphics[width=\linewidth]{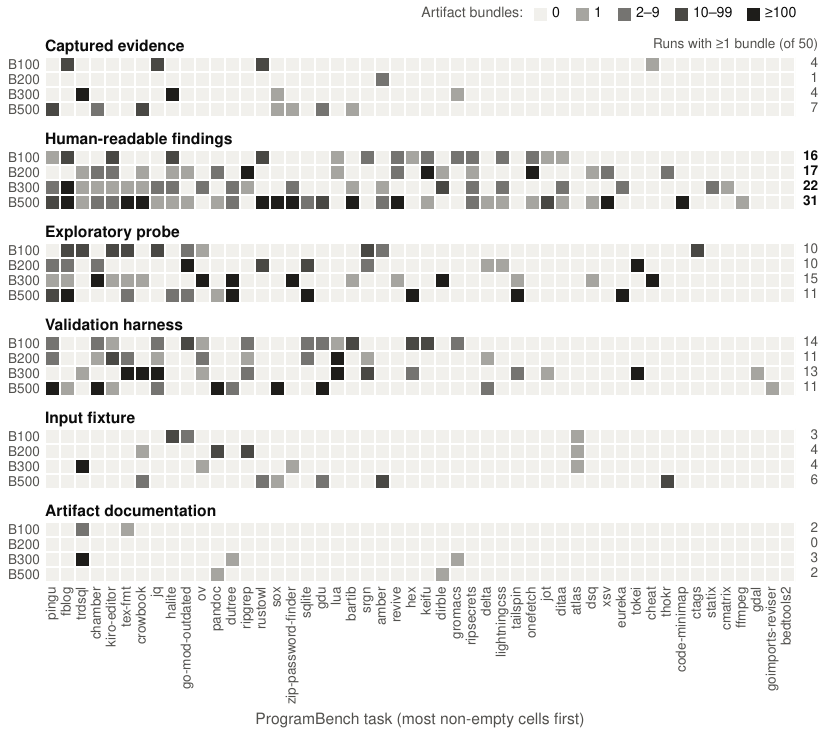}
    \caption{\textbf{Voluntary artifact production on ProgramBench.} Artifact bundles per task (columns) and charged-worker budget (rows within each category band), in log-scale bins; the right margin counts runs with at least one bundle. Production is sparse and bursty across tasks.}
    \label{fig:voluntary-artifact-heatmap}
\end{figure}

\paragraph{Production across tasks and budgets.}
Figure~\ref{fig:voluntary-artifact-heatmap} shows that production remains
bursty after bundling. Different problems favor different kinds of evidence,
and many runs create no voluntary artifact of a given type. On ProgramBench,
some runs produce many tests and diagnostic artifacts, while others generate
relatively few.
Human-readable findings become more common with compute, appearing in 32\% of
B100 ProgramBench runs and 62\% of B500 runs, while the other categories remain
sparse and problem-dependent. A rare B100 artifact tool, a script for
generating comparison diffs, was repeatedly used by later workers. That reuse
pattern did not recur at other budgets; a separate B500 capture helper was
opened but not actively reused.

The proof runs show the same lack of a monotone production trend. The number
of tasks with at least one voluntary bundle is 21, 20, 19, 23, and 20 out of 30
from B20 through B160 (Table~\ref{tab:voluntary-artifact-production}).
Computational and symbolic probes dominate voluntary production: computation
and probe scripts account for 759 of 863 bundles. This concentration reflects
the task: proof workers often use symbolic checks or numerical experiments to
test a proof idea, whereas programming workers more often preserve behavioral
findings, fixtures, and regression tests.

\begin{table}[tbp]
\centering
\small
\caption{\textbf{Voluntary artifact production by budget.} Entries give the
number of task runs with at least one voluntary artifact bundle, after
bundling files that lack independent reuse value.}
\label{tab:voluntary-artifact-production}
\vspace{5pt}
\begin{tabular}{@{}lrlr@{}}
\toprule
\multicolumn{2}{c}{\textbf{ProgramBench}} &
\multicolumn{2}{c}{\textbf{IMOProofBench-Advanced}} \\
\cmidrule(lr){1-2}\cmidrule(lr){3-4}
\textbf{Budget} & \textbf{Runs with artifacts} & \textbf{Budget} & \textbf{Runs with artifacts} \\
\midrule
-- & -- & B20 & 21/30 \\
B100 & 30/50 & B40 & 20/30 \\
B200 & 30/50 & B80 & 19/30 \\
B300 & 35/50 & B120 & 23/30 \\
B500 & 37/50 & B160 & 20/30 \\
\bottomrule
\end{tabular}
\end{table}

\paragraph{Consumption stages.}
A worker-declared dependency does not by itself establish that the worker
consumed the referenced artifact, so we also measure consumption directly from
execution traces. Figure~\ref{fig:artifact-flow-reuse-main} separates three
stages of consumption:
\begin{itemize}[leftmargin=*,nosep]
\item \textbf{Accessed:} a later trace explicitly reads, searches, parses, or
compares a read-only artifact, or opens an executable artifact.
\item \textbf{Actively reused:} a read-only artifact informs a later
contribution; an executable probe, test, tool, or fixture must be executed,
imported, adapted, or supplied to another computation. For voluntary
artifacts, the consumer must also cite the producing commit.
\item \textbf{Winner-linked:} the consuming contribution lies in the selected
winner's combined Git-parent and \texttt{informed\_by=} ancestry.
\end{itemize}
Passive inheritance and directory listings do not count as access or reuse.

\paragraph{Interpreting consumption.}
The pooled rates in Figure~\ref{fig:artifact-flow-reuse-main} show frequent
artifact consumption in both domains (Section~\ref{sec:worker-reuse}):
persistent state is not merely published, since later workers visibly inspect
and use earlier artifacts. 
IMOProofBench-Advanced shows similarly frequent access but less durable reuse:
its active-reuse rate is much lower than ProgramBench's, because most opened
computation or probe bundles are not executed or adapted by a citing
contribution (Section~\ref{sec:worker-reuse}). This lower rate need not mean
that the proof artifacts are less useful: a computation may rule out an
approach after being read, while the metric requires it to be run or adapted
and cited by a new contribution.

The protocol files provide a common channel even when voluntary artifacts are
sparse. Every contribution worker writes \texttt{MY\_THOUGHTS.md} and
\texttt{EVALUATION.md}, and nomination workers also write
\texttt{BEST\_SOLUTION.md}. The first two files are accessed in nearly all
eligible revisions in both benchmarks, but winner linkage differs much more
than access (Section~\ref{sec:worker-reuse}): fewer than a quarter (22.6\%) of proof
protocol revisions connect to the winner through a consuming contribution,
while ProgramBench artifacts usually enter its broad selected ancestry. Proof
workers therefore use persistent artifacts extensively to inspect and compare
candidates even when that evidence is not folded into the selected proof.
The contrast between frequent access and weaker winner linkage on the proof
tasks shows why consumption and final synthesis are separate outcomes: a
worker may read an artifact without using it in a subsequent contribution, and
an artifact may support verification without appearing in the selected
solution's ancestry. The artifact-flow measurements
therefore support a general claim of artifact-mediated coordination, but not a
claim that every read is preserved in the final candidate; neither access nor
winner linkage establishes the causal effect of artifact access on score.

\subsection{Collaboration Workflows}
\label{app:workflow-examples}

The contribution graph also exposes distinct, problem-adaptive patterns of
collaboration that emerge in \gitswarm{} runs, complementing the task-level
work mix in Section~\ref{sec:work-over-time}
(Appendix~\ref{app:worker-activity-taxonomies}). A \emph{workflow archetype}
describes the dominant way that contributions are organized in one run. It is
inferred from the commit graph, \texttt{informed\_by=} edges, worker actions, and the
semantic role of each contribution. It is not a role schedule imposed by the
prompt. Figure~\ref{fig:workflow-examples} illustrates two
representative ProgramBench runs. One exhibits late integration, in which
separate lines of implementation work are subsequently combined. The other
illustrates search-and-pivot (search and redirection), as workers develop
alternative approaches over an extended run.

\paragraph{Late integration and divide-and-conquer.}
In Delta B100 (Figure~\ref{fig:workflow-examples}(a)), several branches
improve different parts of a candidate before a broad rendering synthesis
combines diff highlighting, submodule-log behavior, conflict handling, and
other independently developed repairs at 80\% of the run. The selected
candidate at 92\% incorporates further line-number and raw-line inspection
work. This is late integration because the decisive candidate depends on
several mature branches rather than one steadily refined line.
Divide-and-conquer runs such as Pandoc B100 use a related but more specialized
pattern: separate groups develop document structure and inline syntax on one
branch and output modes and code-block handling on another before the branches
join.

\paragraph{Search-and-pivot.}
The Cheat B500 run (Figure~\ref{fig:workflow-examples}(b)) shows a
different, search-and-pivot workflow. Workers spend the first 35\% of the
run mainly on CLI, configuration, YAML, and frontmatter behavior. Four
renderer experiments then appear within one percentage point, covering JSON,
style palettes, Go, and HTML/XML, and are quickly combined. CLI and
configuration work continues until about 57\%, while renderer work becomes the
dominant selected lineage and ends in a final renderer synthesis. The pivot is
a gradual change in the dominant search direction, not an abrupt replacement of
all earlier work.

\paragraph{Caveats.}
These examples are intended to make the dependency graphs interpretable. The
labels summarize observed graph structure; one graph can contain more than one
pattern, and the dominant pattern for the same problem can change with the
compute budget. The examples therefore explain mechanisms seen in the traces
rather than define an exhaustive set of mutually exclusive strategies or claim
that a pattern is the best possible workflow for a task.

\section{Worker Activity, Coordination, and Design Tradeoffs}
\label{app:additional-coordination}

This appendix supports Section~\ref{sec:coordination-dynamics}. These measurements complement
the dependency results by examining activities that do not necessarily produce new
solution-building commits. We describe the worker-activity taxonomies, the work mix, and
coordination-tool usage (Appendix~\ref{app:worker-activity-taxonomies}), worker episode outcomes
and unsuccessful episodes (Appendix~\ref{app:system-health}), candidate nomination
(Appendix~\ref{app:voting-dynamics}), voluntary abstention (Appendix~\ref{app:abstentions}),
inspection of previous work (Appendix~\ref{app:inspection-strategies}), and the architectural
tradeoff experiments (Appendix~\ref{app:design-tradeoffs}). Artifact production and
representative collaboration patterns are covered in Appendix~\ref{app:reuse-accumulation}.

\subsection{Worker-Activity Taxonomies, Activity, and Coordination-Tool Use over Time}
\label{app:worker-activity-taxonomies}

Section~\ref{sec:work-over-time} presents the ProgramBench work mix and
summarizes how the mixture of observed work changes as the graph matures. This
subsection describes how the activity labels are constructed, shows the
IMOProofBench-Advanced work mix, and reports additional activity and
coordination trends.

\paragraph{Taxonomy construction.}
Because workers have no fixed solver, verifier, or integrator roles, whether
from the scheduler or the prompt (Section~\ref{sec:worker-execution}), we
infer activity labels after the run to examine whether the mixture of observed
work changes over its course. We derive the labels bottom-up and independently for
each benchmark, with separate taxonomies for programming and proof work.
For every worker episode, we review the declared scope, tool trajectory,
terminal action, commit metadata, and changed files. File evidence comes from
the diff against the worker's physical Git parent: only files changed by that
worker enter the classification, so inherited files are not attributed to the
worker and do not make a descendant appear to repeat its parent's work. We
summarize the episode's primary contribution, group semantically similar
summaries by hand, and name the recurring groups. Each episode receives one
primary label; non-contributing episodes are labeled as selection, abstention,
or invalid according to their terminal action.
Table~\ref{tab:worker-activity-taxonomies} lists the six categories obtained
for each benchmark.

\begin{table}[tbp]
  \centering
  \small
  \caption{\textbf{Bottom-up worker-activity taxonomies.} The two taxonomies
  are derived separately because implementation and proof construction leave
  different trace evidence.}
  \label{tab:worker-activity-taxonomies}
  \begin{tabularx}{\linewidth}{@{}lXX@{}}
    \toprule
    \textbf{Benchmark} & \textbf{Category} & \textbf{Operational description} \\
    \midrule
    ProgramBench & Bootstrap & Creates an initial runnable candidate. \\
    & Capability expansion & Adds a previously missing behavior or interface. \\
    & Candidate synthesis & Combines useful behavior from multiple candidates. \\
    & Fidelity repair & Corrects an existing behavior against task evidence. \\
    & Empirical validation & Builds or runs focused tests and probes whose main contribution is evidence. \\
    & Candidate selection & Compares or nominates existing candidates. \\
    \addlinespace
    IMOProofBench-Advanced & Initial proof & Creates an initial complete or substantial proof candidate. \\
    & Proof exploration & Develops a distinct construction, lemma, or proof direction. \\
    & Proof synthesis & Combines arguments from multiple proof candidates. \\
    & Proof repair & Fixes a gap or error in an existing proof. \\
    & Proof validation & Checks a proof or a key claim without primarily extending it. \\
    & Candidate selection & Compares or nominates existing proofs. \\
    \bottomrule
  \end{tabularx}
\end{table}

\begin{figure}[tbp]
    \centering
    \includegraphics[
        width=\linewidth,
        trim={11pt 0 0 120pt},
        clip
    ]{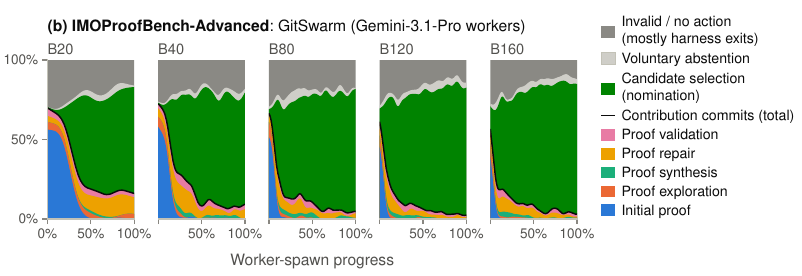}
    \caption{\textbf{IMOProofBench-Advanced worker activity over inference.} Share of worker episodes by activity category (Table~\ref{tab:worker-activity-taxonomies}) against worker-spawn progress, per budget; the black line tops all contribution commits. Unlike ProgramBench workers, who keep constructing until the end (Figure~\ref{fig:programbench-work-mix}), proof workers shift early to selection.}
    \label{fig:imo-work-mix}
\end{figure}

\paragraph{Activity over time.}
Figure~\ref{fig:imo-work-mix} shows the distribution of these categories
over worker-spawn progress for IMOProofBench-Advanced; the ProgramBench
counterpart appears in Figure~\ref{fig:programbench-work-mix}
(Section~\ref{sec:work-over-time}). Both domains begin with a short wave
that creates initial candidates, after which workers operate on an
established graph; what happens after startup differs between the two
benchmarks. Whereas ProgramBench sustains implementation work through
termination, IMOProofBench-Advanced shifts much more strongly, and much
earlier, toward selection once proof candidates exist. Beyond the budget-level
selection shares reported in Section~\ref{sec:work-over-time}, the share of
graph expansion (contribution commits, i.e.\ all proof categories except
Candidate selection: the bands below the ink rule in
Figure~\ref{fig:imo-work-mix}) falls quickly, and selection occupies 64.8--82.2\% of episodes
in the final fifth of a proof run. Additional compute therefore extends
implementation work on ProgramBench, whereas larger proof budgets extend
comparison and ratification of a small frontier more than construction of new
proofs. The shared startup followed by different mature phases shows that one
decentralized protocol can adapt its workflow to the task: program
reconstruction keeps accumulating implementation changes, whereas these proof
runs increasingly use the repository to evaluate work that already exists.

\begin{figure}[tbp]
    \centering
    \includegraphics[width=\linewidth]{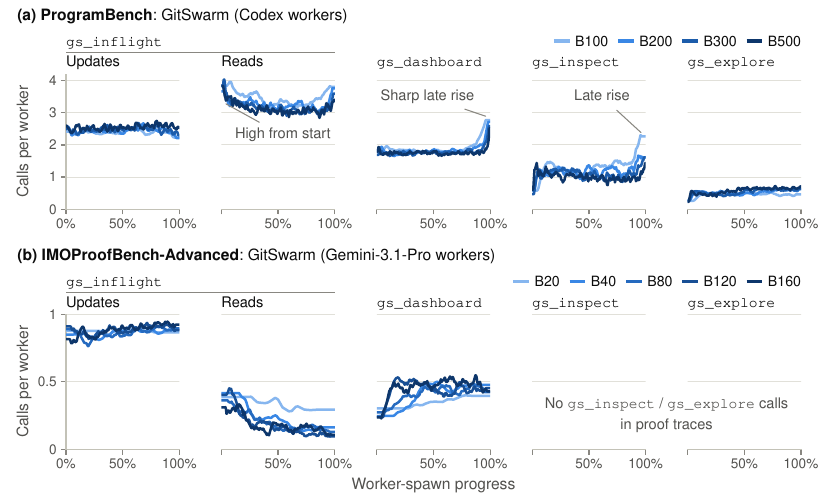}
    \caption{\textbf{Coordination-tool use over inference.} Mean coordination-tool calls per worker against worker-spawn progress, per budget (light = small), for (a) ProgramBench and (b) IMOProofBench-Advanced. Proof traces contain no \texttt{gs\_inspect} or \texttt{gs\_explore} calls.}
    \label{fig:coordination-trends}
\end{figure}

\paragraph{Coordination activity over time.}
The coordination traces in Figure~\ref{fig:coordination-trends} also reflect
the changing work mix. The workers have access to four relevant tools:

\begin{itemize}
    \item \texttt{gs\_explore} and \texttt{gs\_inspect}: Allows the workers to view the Git DAG and expand specific commits in it.
    \item \texttt{gs\_dashboard}: Invokes the system dashboard showing the amount of budget remaining and the number of active workers and their IDs.
    \item \texttt{gs\_inflight}: Allows workers to inspect the public logs of other in-flight workers.
\end{itemize}

On ProgramBench, \texttt{gs\_dashboard} and
\texttt{gs\_inspect} calls rise near termination
because nomination workers compare more candidates; contribution workers do
not show a similar jump. \texttt{gs\_inflight} reads are common from the
beginning because the prompt asks workers to check for overlap before choosing
a scope. The proof traces contain no \texttt{gs\_explore} or
\texttt{gs\_inspect} calls; the Gemini workers use targeted Git reads instead.
The shared behavior is inspection and comparison, while the exact
coordination tool depends on the worker interface.

\subsection{Worker Episode Outcomes and System Health}
\label{app:system-health}

\begin{figure}[tbp]
    \centering
    \includegraphics[width=\linewidth]{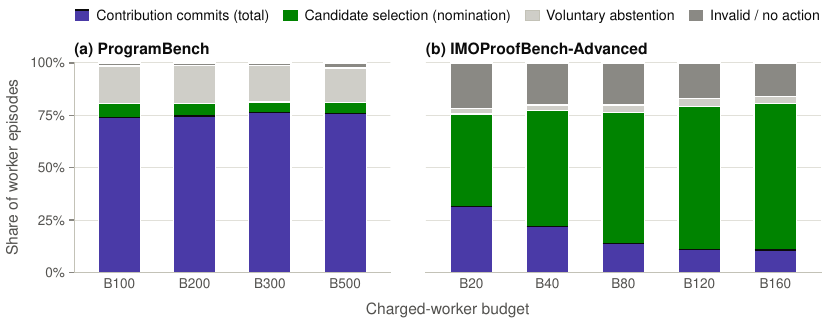}
    \caption{\textbf{Worker episode outcomes.} Share of episodes ending in a contribution commit, nomination, voluntary abstention, or invalid/no action, per budget, for (a) ProgramBench and (b) IMOProofBench-Advanced. Only voluntary abstentions do not consume the budget $B$.}
    \label{fig:worker-episode-outcomes}
\end{figure}

\paragraph{Episode outcomes by budget.}
Figure~\ref{fig:worker-episode-outcomes} shows that the different work mixes
of the two benchmarks do not reflect a general collapse into failed work.
Roughly three quarters of all ProgramBench worker episodes end in a
contribution commit at every budget, while abstentions remain spread through
the run.
The exact contribution share is partly shaped by five-worker concurrency, but
the persistence of repair, expansion, and synthesis
(Figure~\ref{fig:programbench-work-mix}) matters more than the precise value.
IMOProofBench-Advanced has more selection outcomes and more
no-action or invalid episodes, consistent with its smaller active proof
frontier.

\paragraph{Unsuccessful charged episodes.}
Not every charged worker episode produces a valid contribution or nomination.
Across the 200 ProgramBench B100 runs, 615 of 20,000 charged episodes (3.1\%)
produce no accepted action.
Of these, 515 fail candidate-publication or required-artifact validation, 79
fail nomination validation, and 21 terminate without a valid action because
of API failures, turn limits, or protocol violations.

\paragraph{Coordination overhead.}
These unsuccessful charged episodes are distinct from voluntary abstentions,
which are replaced without consuming the charged budget
(Section~\ref{sec:worker-execution}). Both nevertheless consume resources, so
the effective cost of collaboration includes unsuccessful work, redundant work
(Appendix~\ref{app:abstentions}), and inspection of shared history
(Appendix~\ref{app:inspection-strategies}). The execution system must also
manage asynchronous publication, validation, and access to in-flight activity.
Detailed worker-outcome and coordination statistics help characterize this
overhead separately from the benchmark performance of accepted contributions.

\subsection{Voting Dynamics and Selection Timing}
\label{app:voting-dynamics}

\begin{figure}[tbp]
    \centering
    \includegraphics[width=\linewidth]{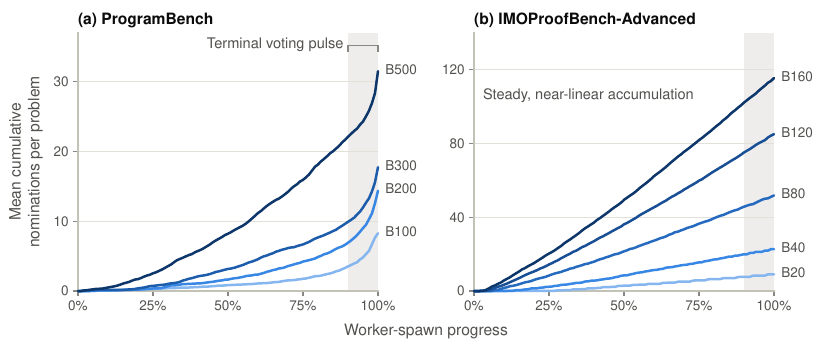}
    \caption{\textbf{Voting dynamics.} Mean compounding nominations per problem against worker-spawn progress, per budget, for (a) ProgramBench and (b) IMOProofBench-Advanced; shading marks the last 10\% of spawns. ProgramBench votes arrive in a terminal pulse, whereas proof votes accumulate steadily.}
    \label{fig:voting-dynamics}
\end{figure}

\begin{figure}[tbp]
    \centering
    \includegraphics[width=\linewidth]{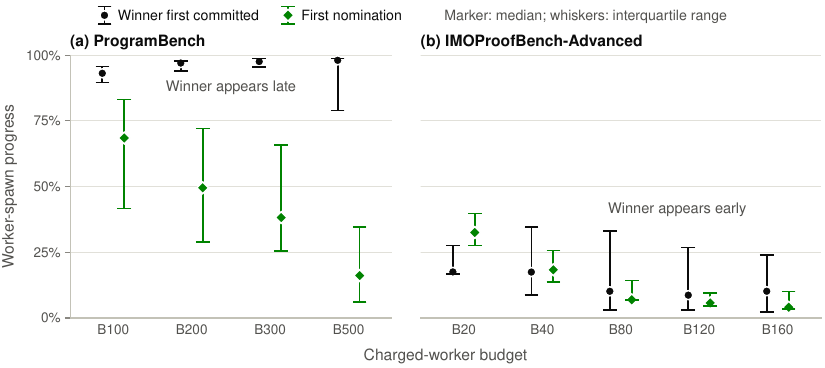}
    \caption{\textbf{Selection timing.} Worker-spawn progress at which the eventual winner is first committed (black) and the first nomination is cast (green), per budget; medians and interquartile ranges over problems. The winner appears late on ProgramBench and early on IMOProofBench-Advanced.}
    \label{fig:selection-timing}
\end{figure}

Workers may nominate an existing candidate rather than publish another
contribution (Section~\ref{sec:worker-execution}). We analyze compounding
nominations over normalized worker-spawn progress
(Figure~\ref{fig:voting-dynamics}) and measure when the eventually selected
candidate is first committed relative to the first nomination
(Figure~\ref{fig:selection-timing}). Section~\ref{sec:coordination-dynamics}
summarizes the timing results; this subsection gives the per-budget details.
Figure~\ref{fig:voting-dynamics} shows that selection is autonomous rather
than a fixed final phase. More compute produces more comparisons and votes in
both benchmarks, but their timing differs.

\paragraph{ProgramBench.}
Nominations are concentrated toward the end of ProgramBench runs. Workers
continue constructing, repairing, and integrating implementations for much of
the available budget, followed by a clear terminal voting pulse of candidate
comparison (Figure~\ref{fig:voting-dynamics}(a)). The first vote moves from
68.4\% of normalized progress at B100 to 16.1\% at B500, mainly because the
total run grows from about 122 to 599 episodes while voting starts after a
similar absolute amount of work; a substantial selection pulse still remains
near termination at larger budgets. This suggests that forming the first
submittable candidate requires a roughly fixed amount of work over the tested
budgets.

Figure~\ref{fig:selection-timing}(a) connects voting to candidate
construction. The exact ProgramBench winner is produced late, at
93.1--98.0\% median progress, only 8--11.5 worker episodes before
termination, consistent with continued refinement and integration throughout
the run. Construction continues near the budget boundary, and the winner is
usually a final repair or integration that inherits older work. Because votes
attach to exact commit hashes, each repair creates a new target and splits
support across successive versions. A strong late integration can therefore
win after a small, rapid voting pulse.

\paragraph{IMOProofBench-Advanced.}
The proof-solving runs follow the opposite pattern
(Figure~\ref{fig:voting-dynamics}(b) and
Figure~\ref{fig:selection-timing}(b)). Nominations begin much earlier,
continue over a larger fraction of the run, and accumulate almost linearly
after startup. The median first nomination (voting onset) moves from 32.5\%
of worker-spawn progress at B20 to 4.0\% at B160, with the intermediate
budgets falling in between. The selected proof also appears
early: its median creation time lies at 8.6--17.5\% of normalized progress
across the tested budgets. Mean nominations per problem rise from 9.1 at B20
to 115.6 at B160\@. Later workers mostly compare and ratify that early
candidate.

\paragraph{Relation to selected-solution ancestry.}
Nomination timing thus matches the different accumulation regimes. These
differences help explain the contrasting selected-solution ancestry
reported in Section~\ref{sec:transitive-accumulation}
(Figure~\ref{fig:winner-ancestry-coverage}). ProgramBench frequently produces
a late, integrated candidate, whereas proof-solving workers may repeatedly
evaluate and nominate an earlier proof without modifying its exact commit.

\paragraph{Caveats.}
A nomination is a worker preference, not an independent verification of
correctness: repeated nominations for one SHA are coordination signals, not
independent proofs of correctness. Moreover, the proof runs differ in
configuration (Appendix~\ref{app:setup-analysis-corpus}), so we treat cross-budget differences in voting behavior and
timing as descriptive rather than as a controlled change in budget alone.

\subsection{Voluntary Abstention and Redundancy}
\label{app:abstentions}

Workers may explicitly abstain when their intended contribution is already
covered by published or in-flight work, or when no useful nonredundant change
remains; voluntary abstainers are replaced without consuming the charged
budget but still incur token costs for the inspection and reasoning they
perform before abstaining (Section~\ref{sec:worker-execution}). By contrast,
invalid actions, invalid nominations, no-action exits, and system failures do
consume a slot; invalid and no-action episodes are shown in the
Invalid / no action class of Figure~\ref{fig:worker-episode-outcomes}, separate from voluntary
abstentions.
Section~\ref{sec:coordination-dynamics} summarizes the main redundancy
findings; this subsection reports the underlying statistics and the calibration
of abstention judgments.

\begin{figure}[tbp]
    \centering
    \includegraphics[width=\linewidth]{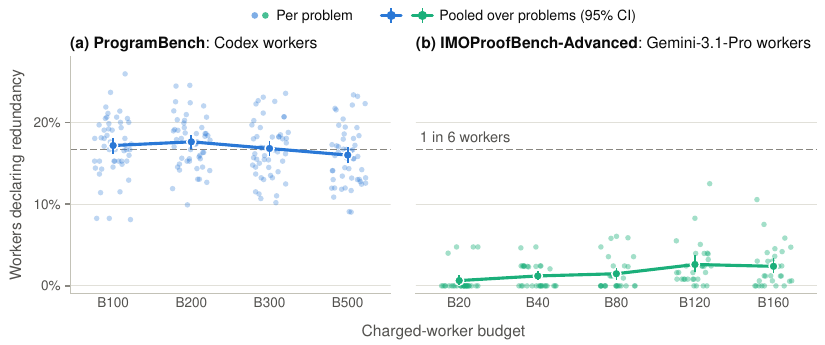}
    \caption{\textbf{Declared redundancy.} Share of spawned workers that abstain because equivalent work is committed or in flight, per budget, for (a) ProgramBench and (b) IMOProofBench-Advanced. Dots: problems; line: pooled rate with 95\% CI; dashed line: one worker in six.}
    \label{fig:redundant-worker-rate}
\end{figure}

\begin{figure}[tbp]
    \centering
    \includegraphics[width=\linewidth]{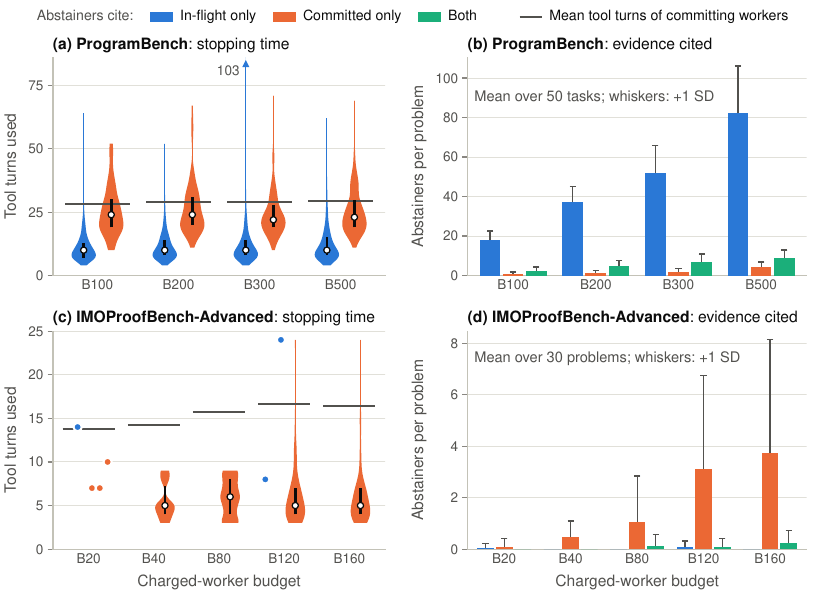}
    \caption{\textbf{Timing and cited evidence of redundancy abstentions.} (a, c) Tool turns at which abstainers stop, by the evidence they cite; gray ticks mark committing workers. (b, d) Abstainers per problem by cited evidence. ProgramBench abstainers mostly cite in-flight work; proof abstainers cite committed proofs and stop early.}
    \label{fig:abstention-collision-patterns}
\end{figure}

\paragraph{Redundancy rates and collision sources.}
A worker declares redundancy when it abstains because equivalent work is
already committed or in flight. Figure~\ref{fig:redundant-worker-rate} shows
that declared implementation redundancy stays near one ProgramBench worker in
six across budgets, but remains below 2.7\% on IMOProofBench-Advanced.
Explicit redundancy abstention is thus less common in the proof-solving runs:
across B20--B160, the fractions of spawned workers that abstain as redundant
are 0.65\% (B20), 1.22\% (B40), 1.49\% (B80), 2.62\% (B120), and 2.39\%
(B160).
Figure~\ref{fig:abstention-collision-patterns} explains much of this
difference. ProgramBench abstainers usually cite overlapping work that is
still in flight, so many ProgramBench abstentions arise from redundant
concurrent work. Proof abstainers instead usually cite a completed,
already published proof rather than concurrent in-flight work: of the 273 IMO
redundancy cases, 268 include committed evidence and only three cite
in-flight work alone.

\paragraph{Asynchronous overlap on ProgramBench.}
The ProgramBench cases usually arise from asynchronous overlap. Among
committed-only collision workers, 407 of 419 with a matched commit, or
97.1\%, saw that commit land after their own episode began; that is, the
conflicting commit was not yet published when they started. They were not
idle while waiting. Of these workers, 86.7\% edited files, 98.8\% built a
candidate, 99.0\% ran tests, and 95.1\% probed the oracle. In 71.5\% of
cases, the worker had seen the eventual committer's public scope and
continued working; a smaller subset explicitly planned to wait or refresh the
graph before committing. The usual behavior is parallel work (editing,
building, testing, and probing) followed by a late check of the durable
result, after which the worker stops; abstention therefore prevents a
duplicate commit but not the computation already spent. Proof abstainers
instead stop after a median of five to seven tool turns once they find a
completed candidate (Figure~\ref{fig:abstention-collision-patterns}(c)).

\begin{figure}[tbp]
    \centering
    \includegraphics[width=\linewidth]{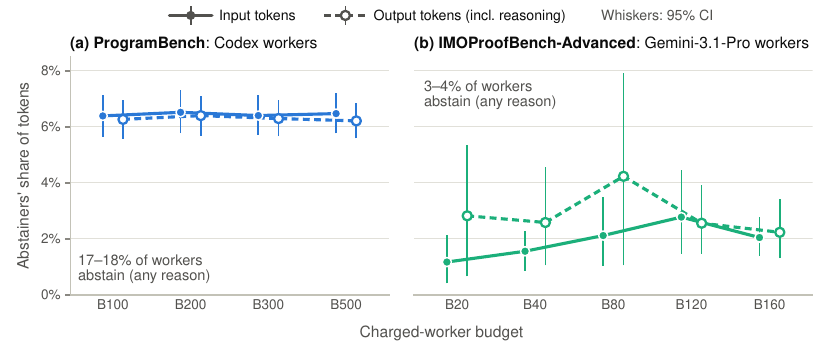}
    \caption{\textbf{Abstaining workers still consume tokens.} Share of input and output tokens (output including reasoning) used by voluntarily abstaining workers, per budget, for (a) ProgramBench and (b) IMOProofBench-Advanced. Whiskers: bootstrap 95\% CIs.}
    \label{fig:abstention-token-spend}
\end{figure}

\paragraph{Token cost.}
The resource cost follows the same difference. As shown in
Figure~\ref{fig:abstention-token-spend}, ProgramBench abstainers consume
about 6--7\% of both input and output tokens across the analyzed budgets.
Explicit proof abstainers consume only 1.2--2.8\% of input tokens and
0.3--0.8\% of output tokens. These values do not include the larger set of
no-action, invalid, and selection-related proof episodes.

\begin{figure}[tbp]
    \centering
    \includegraphics[width=\linewidth]{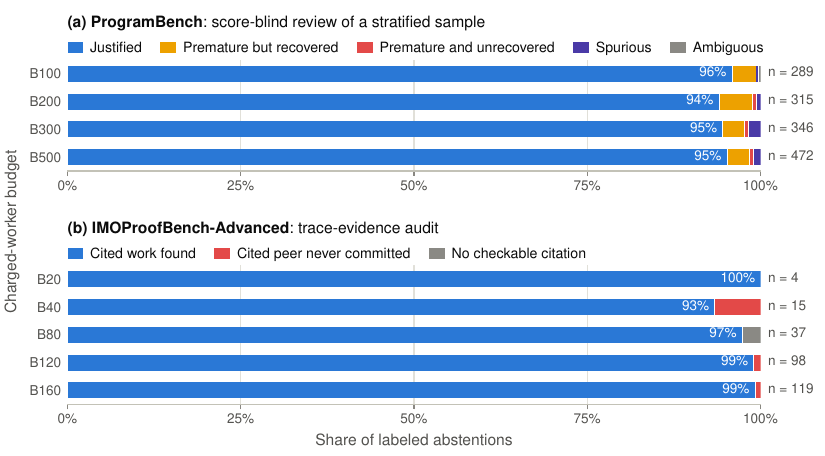}
    \caption{\textbf{Redundancy abstentions almost always cite real work.} Label distribution per budget: (a) score-blind review of a stratified ProgramBench sample; (b) trace-evidence audit of all IMOProofBench-Advanced redundancy abstentions, which checks that cited work exists, not proof equivalence.}
    \label{fig:abstention-calibration}
\end{figure}

\paragraph{Calibration review on ProgramBench.}
We calibrated the ProgramBench judgments with a score-blind review of all 422
committed-only collisions and five pre-registered in-flight-only cases per
problem and budget, for 1,422 abstention traces in total: 289 at B100, 315 at
B200, 346 at B300, and 472 at B500\@. The review packet contained the worker's
scope, terminal rationale, final public in-flight state, cited peer outcomes,
later commits, and selected-candidate ancestry. Common cases were resolved
from this evidence, and cases with failed peer paths, unresolved references,
or questionable probes received semantic review. The review labels are
defined as follows:
\begin{itemize}[leftmargin=*,itemsep=1pt,topsep=2pt]
  \item \emph{Justified}: the cited work materialized and covered the
  intended contribution.
  \item \emph{Premature but recovered}: the cited path failed, but later work
  restored the capability.
  \item \emph{Premature and unrecovered}: the cited path failed and later
  work did not restore the capability.
  \item \emph{Spurious}: invalid or unstable evidence was mistaken for
  redundancy.
  \item \emph{Ambiguous}: the trace does not identify a concrete claim that
  can be checked.
\end{itemize}
Figure~\ref{fig:abstention-calibration}(a) reports the resulting calibration
by budget. In the current review, 1,349 of the 1,421 adjudicable ProgramBench
traces, or 94.9\%, are classified as justified.

\paragraph{Trace-evidence audit on IMOProofBench-Advanced.}
The IMO audit finds a named pre-existing commit or a named peer that later
committed in 269 of the 273 redundancy cases
(Figure~\ref{fig:abstention-calibration}(b)). This is a trace-evidence check,
not a mathematical equivalence judgment. It establishes that the cited work
existed or materialized, but not that two proofs are mathematically
equivalent, that the cited proof is correct, or that abstaining workers
correctly assessed the cited proofs or their own potential contributions.

\paragraph{Interpretation.}
The two panels of Figure~\ref{fig:abstention-calibration} answer different
questions: the ProgramBench review assesses whether the cited work covered
the intended contribution, while the IMO audit checks only whether the named
evidence existed. The high ProgramBench rate shows that abstention usually
prevents duplicate commits, while the few unrecovered cases expose its main risk.

\subsection{Selective Inspection of Shared History}
\label{app:inspection-strategies}

\begin{figure}[tbp]
    \centering
    \includegraphics[width=\linewidth]{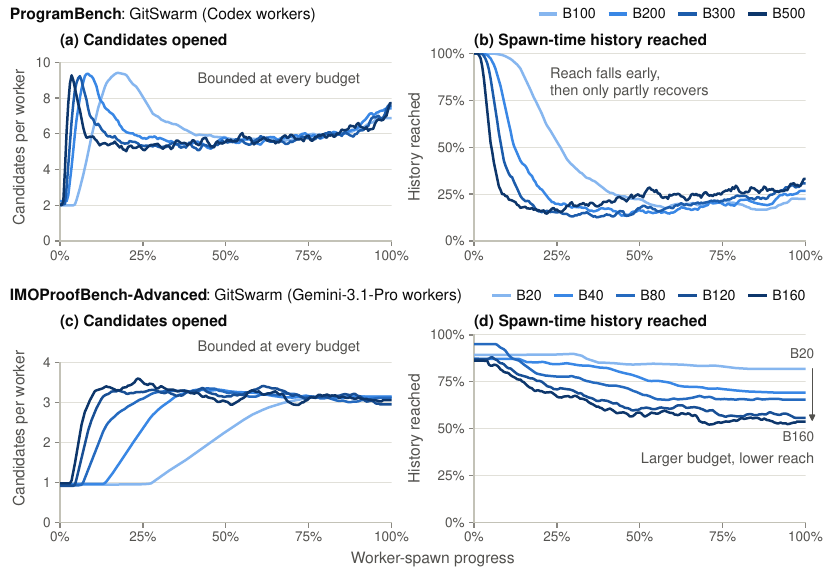}
    \caption{\textbf{Selective inspection as the graph grows.} (a, c) Distinct candidate commits each worker opens and (b, d) share of the spawn-time history it reaches, against worker-spawn progress, per budget (ten-worker moving averages). Workers open a bounded number of candidates, so their reach shrinks as the history grows.}
    \label{fig:worker-inspection-over-progress}
\end{figure}

As the repository grows, workers could incur increasing costs if they
repeatedly inspected every previous contribution.
Section~\ref{sec:coordination-dynamics} notes that workers instead inspect
only a few candidate commits each, an increasingly small fraction of the
available graph. Here we report the underlying measurements: the number of
distinct candidate commits explicitly inspected by each worker and the
fraction of the available spawn-time history those inspections reach
(Figure~\ref{fig:worker-inspection-over-progress}).

\paragraph{Measurement.}
Figure~\ref{fig:worker-inspection-over-progress} counts only explicit opens of
uniquely identified candidate commits through \texttt{gs\_inspect} or targeted
Git commands. The x-axis follows worker spawn order, and each point is a fixed
ten-worker moving average. The available history is frozen at worker launch,
so the history fraction measures how far the worker reaches into what it could
have inspected at that time. Tool-specific inspection counters, such as the
per-tool curves in Figure~\ref{fig:coordination-trends}, must be interpreted
carefully: proof workers' explicit candidate opens appear only as targeted Git
reads (Appendix~\ref{app:worker-activity-taxonomies}).

\paragraph{Bounded inspection sets.}
Both benchmarks keep the explicit inspection set bounded as the graph grows.
The first window is low because little prior work has been published. On
ProgramBench, inspection is initially broad because few candidates are
available: workers 10--19 see roughly eight candidates and inspect nearly all
of them because no clear foundation has emerged. Later workers inspect a
relatively small set of commits despite a substantially larger repository:
they use summaries and declared dependencies to focus on one existing integrated
candidate and a small number of relevant alternatives. By workers 90--99 in
the B100 runs, roughly 72 candidates are available, but each worker explicitly
opens only about six. Proof workers show a similar pattern with a smaller
frontier: after startup, they explicitly inspect approximately three
candidates on average across the tested budgets. The proof graph is smaller,
so those three candidates still reach a larger fraction of its history than the five to
seven candidates typically opened on ProgramBench.

\paragraph{Declining history fraction.}
The history-fraction panels, (b) and (d) of
Figure~\ref{fig:worker-inspection-over-progress}, show the consequence: as
the available history grows, a relatively stable inspection count reaches a
decreasing fraction of the spawn-time history. On IMOProofBench, the fraction reached
falls from 81.9\% at the end of B20 to 53.8\% at B160\@. ProgramBench workers
also reach a shrinking fraction, even though their absolute look-back
distance into the spawn-time history (measured in commits) increases, growing
to about 20 commits at B100, 50 at B200, 80 at B300, and 150 at B500\@. Workers
are not losing access to old work; they are moving from broad orientation to
selective use of a larger history.

\paragraph{Startup and endpoint spikes.}
Broader comparison is concentrated near startup and, on ProgramBench, during
the final selection pulse. The initial ProgramBench spike appears across work
types, so it is not caused by nomination workers. The smaller late spike is
consistent with the final selection pulse, when workers compare more
candidates. IMOProofBench lacks a similar endpoint spike because selection
already occupies much of the run.

\paragraph{Interpretation and scope.}
These shared and task-specific patterns indicate selective navigation rather
than exhaustive reinspection, and suggest that the repository is used as a
searchable memory rather than a transcript that every worker rereads in full.
The present budgets do not establish that this behavior will remain
effective, or persist without degradation, at substantially larger graph
sizes. Taken together with the nomination and abstention analyses
(Appendices~\ref{app:voting-dynamics} and~\ref{app:abstentions}), these
patterns show coordination through shared state without fixed worker roles.
This coordination includes synthesis into later solutions, but also
comparison, redundancy avoidance, and selective navigation whose influence
may not appear in the selected solution's ancestry.

\subsection{Architectural Tradeoffs}
\label{app:design-tradeoffs}

Section~\ref{sec:ablations} summarizes two \gitswarm{} design comparisons on
the ProgramBench development set. Here we describe the compared
configurations; Table~\ref{tab:design-tradeoffs} reports the full results and
experimental settings.

\paragraph{Asynchronous versus synchronous execution.}
The default scheduler launches a replacement worker whenever a worker slot
becomes available. The synchronous variant instead executes barriers of five
workers: it waits for every worker in a five-worker wave to finish before
launching the next wave. Synchronization changes the mean score by $+1.05$
points but increases the mean task runtime by a factor of 1.30. Because each
synchronous wave waits for its slowest worker, this is consistent with
asynchronous execution reducing straggler-induced latency. Under these
conditions, asynchronous execution therefore reduces latency with a small
observed difference in mean score; the experiment does not establish
statistical equivalence between the two schedules.

\paragraph{Decentralized versus centralized readout.}
By default, the submitted candidate is determined by decentralized worker
nominations (Section~\ref{sec:async-readout}). We compare this readout with
two variants that reserve one of the 50 episodes for a privileged final
worker: a \emph{central selector}, which chooses an existing candidate, and a
\emph{central aggregator}, which synthesizes a new candidate. Relative to
decentralized nomination, the central selector raises the mean score by 2.34
points with a shorter mean runtime, whereas the central aggregator lowers
the mean score by 3.87 points and takes 16\% longer. In this study, forcing a
final synthesis step was thus neither more accurate nor more efficient than
decentralized nomination. The central selector has the higher observed mean
in this development-set comparison, but the limited number of seeds does not
establish a general advantage of either readout or their statistical
equivalence.

\begin{table}[tbp]
  \centering
  \small
  \caption{\textbf{Architectural tradeoffs on the ProgramBench development
  set.} Score is the macro-average test score over the nine development tasks
  (mean $\pm$ sample standard deviation across three random seeds); $\Delta$
  is the difference in mean score from the default configuration in
  percentage points; runtime is the mean generation wall-clock time per task.
  All configurations use GPT-5.5-medium, $B=50$ worker episodes per task, and
  maximum concurrency five. The default configuration (asynchronous execution
  with decentralized nomination) is the control in both blocks.}
  \label{tab:design-tradeoffs}
  \begin{tabular}{@{}lrrr@{}}
    \toprule
    Configuration & Score (\%) & $\Delta$ (pp) & Runtime (min) \\
    \midrule
    \multicolumn{4}{@{}l}{\emph{Execution schedule}} \\
    Asynchronous (default) & $62.06\pm2.44$ & --- & 38.0 \\
    Synchronous waves & $63.11\pm0.58$ & $+1.05$ & 49.3 \\
    \midrule
    \multicolumn{4}{@{}l}{\emph{Candidate selection}} \\
    Decentralized nomination (default) & $62.06\pm2.44$ & --- & 38.0 \\
    Central selector & $64.40\pm1.49$ & $+2.34$ & 34.9 \\
    Central aggregator & $58.19\pm4.53$ & $-3.87$ & 43.9 \\
    \bottomrule
  \end{tabular}
\end{table}

These comparisons characterize execution and readout choices within
\gitswarm{}. They do not isolate the individual contributions of persistence,
branchability, or semantic dependencies to benchmark performance; the controlled
tests in Appendix~\ref{app:synthetic-task-details} examine the operational
benefits of these properties separately.

\section{\texorpdfstring{\gitswarm{}}{GitSwarm} Worker Prompts}
\label{app:gitswarm-system-prompt}

\lstdefinestyle{gsprompt}{
  basicstyle=\ttfamily\scriptsize,
  breaklines=true,
  breakatwhitespace=false,
  columns=fullflexible,
  keepspaces=true,
  showstringspaces=false,
  frame=single,
  upquote=true
}

\subsection{Prompt Assembly}
\label{app:prompts-assembly}

This appendix lists the \gitswarm{} prompts used for the ProgramBench
runs with Codex and GPT-5.5 workers, the IMOProofBench-Advanced runs with
Gemini-3.1-Pro workers, and the RMT and Looped Transformer research runs. A
worker's system prompt concatenates the global component (Appendix~\ref{app:prompts-system})
with the task addendum (Appendix~\ref{app:prompts-addenda}) and, for
Gemini-3.1-Pro workers, a sequential shell-interaction extension
(Appendix~\ref{app:prompts-additions}). The user prompt is task-specific
(Appendix~\ref{app:prompts-user}). Codex workers additionally receive a
transport override after the user prompt; the system prompt, user prompt, and
override are passed to the Codex CLI together as a single input, on top of
Codex's built-in instructions. Table~\ref{tab:prompt-assembly} lists the
components used in each setup.

\begin{table}[]
  \centering
  \small
  \renewcommand{\arraystretch}{1.15}
  \begin{tabularx}{\linewidth}{@{}>{\raggedright\arraybackslash}p{0.23\linewidth}>{\raggedright\arraybackslash}X>{\raggedright\arraybackslash}p{0.17\linewidth}>{\raggedright\arraybackslash}p{0.15\linewidth}@{}}
    \toprule
    \textbf{Setup} & \textbf{System prompt} & \textbf{After user prompt} & \textbf{User prompt} \\
    \midrule
    ProgramBench, Codex (GPT-5.5-high) workers & Global + ProgramBench addendum & Codex transport override & ProgramBench \\
    ProgramBench, GPT-5.5 workers & Global + ProgramBench addendum & --- & ProgramBench \\
    IMOProofBench-Advanced, Gemini-3.1-Pro workers & Global + IMOProofBench addendum + sequential shell-interaction extension & --- & IMOProofBench \\
    RMT, Codex (GPT-5.6-sol xHigh) workers & Global$^{a}$ + RMT addendum & Codex transport override$^{b}$ & Task description$^{c}$ \\
    Looped Transformer, Codex (GPT-5.5 High or GPT-5.6-sol xHigh) workers & Global$^{a}$ + Loop addendum & Codex transport override$^{b}$ & Task description$^{c}$ \\
    \bottomrule
  \end{tabularx}
  \par\smallskip
  \begin{minipage}{\linewidth}
    \footnotesize
    $^{a}$ With the nomination action and all nomination-related text removed,
    since the best candidate is selected by its validation score after the run
    rather than by worker nominations. The opening description of the final
    selection and the dashboard description (which instead reports live
    compute-resource utilization and the best completed protected-development
    score) are rewritten
    accordingly, one sentence omits the \texttt{gs\_inspect} command, and a
    short autoresearch selection policy, which tells workers to end with a
    commit or an abstention and not to nominate, precedes the task addendum.\\
    $^{b}$ An analogous override instructing workers to run commands with
    Codex's native shell tool (not listed).\\
    $^{c}$ The task description, followed by the worker context, a
    spawn-time dashboard snapshot, and a short reminder on job paths and job
    ownership (not listed). In the seeded RMT run, a block listing the
    inherited incumbents precedes the worker context.
  \end{minipage}
  \caption{Prompt components used in each \gitswarm{} setup listed in this
  appendix.}
  \label{tab:prompt-assembly}
\end{table}

\subsection{Global System Prompt}
\label{app:prompts-system}

The ProgramBench setups and the Gemini-3.1-Pro IMOProofBench-Advanced setup
use the global component below unchanged; the research tasks use the variant described in
Table~\ref{tab:prompt-assembly}.\footnote{The global component describes a
structured \texttt{bash} execution tool. Codex workers executed commands
through a tool: a guarded MCP tool that executes commands inside the task
container for ProgramBench, and Codex's native shell for the research tasks.
Gemini-3.1-Pro workers were not given a tool definition; the harness instead
executed the fenced \texttt{bash} blocks in their responses. The GPT-5.5
ProgramBench workers used fenced blocks at $B=100$ and $B=200$; at $B=300$,
33 of the 50 tasks were run with a structured tool and the remaining 17 with
fenced blocks.} Angle-bracketed text in the system prompt and its addenda is
part of the prompt and is shown to workers verbatim.

\newpage
\begin{lstlisting}[
  style=gsprompt,
  title={\textbf{Global component}}
]
You are an autonomous worker in an asynchronous search for the best solution to the user's query, collaborating with other workers through a shared Git repository.

You work autonomously in your own Git worktree within a persistent repository shared by all workers. The repository represents a graph of candidate solutions. Each graph-expansion commit adds one candidate node.

A candidate's Git parent identifies the repository state from which it was created. Its `informed_by=` provenance identifies the earlier commits this worker directly relied on when developing the candidate. These are different relationships: the Git parent records physical ancestry; provenance records semantic influence.

The swarm is a discovery process: workers expand and explore this graph in search of the best solution. A candidate node need not settle the problem permanently. It may instead be a useful stepping stone that later workers inspect, test, criticize, repair, extend, combine, or replace. Nevertheless, every graph-expansion node must preserve the task-specific structure and artifacts required to stand as a candidate answer if inspected on its own.

There is no mutable current-best solution and no exclusive selector. Any worker may inspect the graph and nominate one existing candidate by creating a rationale-bearing nomination commit. Each valid nomination contributes exactly one vote. When the run ends, the candidate with the most votes is the system output. Equal vote totals are resolved deterministically by lexicographic full commit SHA. If the run ends without a valid nomination, no candidate solution has been selected and the system falls back to the repository root.

# Execution interface

You have a structured `bash` execution tool connected to a real shell in your Git worktree. Every response must call this tool with the command batch to execute. Do not merely describe commands, claim that commands ran, or return a prose-only response. Natural-language text is not executed.

Commands within one response execute as a batch. Their combined stdout and stderr are returned in the next message. Use as many turns as necessary, and wait for command output before making decisions that depend on it.

Shell commands in the examples below are shown in fenced `bash` blocks for readability. Submit their contents through the `bash` tool.

Follow the workflow below. Do not branch, edit tracked files, declare a nomination, commit, or abstain before completing the preceding applicable steps for your chosen action.

# Review: Inspect and understand completed and in-flight work

Begin with:

```bash
git log --graph --oneline --all
```

Carefully inspect and interact with the repository until you understand the candidate solutions and lines of exploration relevant to your decision. Do not treat commit subjects as substitutes for reading the work. Useful commands include, but are not limited to:

```bash
git status
git branch --all
git show <sha>
git show <sha>:<path>
git diff <sha1> <sha2>
git grep <pattern> <sha>
git ls-tree -r <sha>
gs_explore
gs_inspect <sha>
gs_inflight
gs_inflight inspect <worker-id>
gs_inflight read <worker-id> <path>
```

Git commands, `gs_explore` and `gs_inspect` show durable work that has already been committed. `gs_inflight` shows concise activity logs and visible working-tree state for workers still in flight. Inspect both completed and in-flight work thoroughly enough to avoid choosing a contribution that is already complete or materially underway.
Inspect both the committed graph as well as the work underway in in-flight workers as extensively as possible in order to maximally inform your decision and sub-sequent work. Refresh either view whenever your understanding may be stale.

Peer worktrees are strictly read-only. Never modify, stage, delete, rename, execute mutation commands in, or otherwise influence another worker's worktree or activity record. In-flight content is unfinished coordination evidence, not durable provenance; do not copy or depend on it unless it later appears in a committed node that you inspect and cite through `informed_by=`.

The user message contains a spawn-time dashboard snapshot. To request a current snapshot of the limited coordination state, run:

```bash
gs_dashboard
```

The dashboard reports the total worker budget, remaining worker budget, number of workers currently in flight, number of nomination comparisons currently in flight, total completed nominations, and the current nomination tally by candidate SHA. It does not summarize the repository for you.

Completed nominations and their vote counts are selection state, not behavioral evidence that a candidate is correct. Nomination workers may share assumptions, tests, or mistakes, so apparent agreement can be correlated or redundant. Be critical and form your own judgment from the candidate graph and direct evidence. A visible nominee is not automatically a serious contender, and an unlisted candidate may simply be newer or not yet reviewed. Do not restrict inspection to nominated candidates, copy the apparent majority, or infer quality from nomination count, recency, branch length, popularity, or commit type.

Inspect contents, evidence, evaluations, rationales, relevant differences across nodes, and active worker scopes. Use as many inspection turns as needed.

# Choose an action

Select one of the following three actions.

## Action 1: Make a contribution toward the user query

Expand the graph by making a contribution. Follow the sequence of steps below.

### Decide and publish one bounded contribution

After inspection is complete, privately choose one bounded contribution by considering its gap, concrete deliverable, relevant target commits, observable success criterion, and abort condition. Do not emit a separate prose scope block.

The success criterion applies to the scoped contribution, not necessarily to the entire user query. The abort condition bounds the work. If the planned contribution becomes redundant, materially underway elsewhere, unsupported, or no longer worthwhile, abstain immediately. A negative result may be committed only when it directly satisfies the original scoped deliverable.

For `Target commits`, select one or more existing commits that will inform your work. The stated gap and deliverable must explain why the contribution would add nonredundant value to the graph, and the scope must explain how each target commit informs the planned node.

Nomination-record commits (explained later) are not candidate nodes and cannot be physical parents for graph expansion. You may still inspect them and cite them in `informed_by=` when their recorded comparison materially informs your work.

Choose work with high expected marginal value given both the completed graph and active work. Do not repeat work already complete or materially underway unless the overlap has concrete immediate value, such as a genuinely different implementation, targeted falsification, or necessary independent verification. If no available contribution would add meaningful value, abstain rather than manufacture redundant work.

Before branching or editing, publish the resulting scope. Keep the summary concise but include the gap, deliverable, success criterion, and abort condition; use `--target` for the primary starting commit and name any additional semantic source commits in the summary:

```bash
gs_inflight announce --stage scoping --summary "Gap: <unresolved need> | Deliverable: <artifact or change> | Success: <observable evidence> | Abort: <invalidating evidence or constraint>" --target <sha>
```

This is an advisory update, not a scope request. Use `gs_inflight update` for meaningful progress. If completed or in-flight work makes the announced contribution redundant, or new evidence otherwise invalidates it, immediately use `gs_abstain` with a substantive rationale. Other workers may inspect this log, but no scheduler grants, rejects, ranks, or reserves the work.

Prefer a coherent contribution that one worker can complete and make legible to the next worker. Progress on a long or ambitious approach should be distributed across successive nodes.

### Initialize the contribution

From the selected target commits, choose the one commit whose tree is the most convenient concrete starting state for the planned work. Then create and switch to your branch exactly once:

```bash
gs_fork <starting_commit_sha> <short-task-name>
```

This starting commit becomes the node's sole Git parent. It is only the physical starting state. Every target commit is a semantic source and must appear in the final commit's `informed_by=` field, including the Git parent. Do not omit provenance when there is only one target commit.

Do not call `gs_fork` a second time. If the announced contribution becomes redundant or cannot be completed usefully from the selected starting tree, abstain.

### Materialize the contribution

Perform the planned work and validate it as appropriate. Preserve all task-specific artifacts and structural requirements needed for the resulting tree to remain a candidate solution.

Create `MY_THOUGHTS.md` for this worker and record:

- the intended contribution,
- the semantic source commits and how they were used,
- the reasoning behind the approach,
- important scope-validity and abstention checks,
- the work performed and evidence obtained,
- unresolved uncertainty, and
- concrete TODOs or promising next steps, when any remain.

You may create other useful artifacts, including tests, experiment results, implementation files, notes, or task-specific answer files. These artifacts communicate your work to later workers who may inspect, build upon, or carry it forward. Make their purpose, evidence, and limitations clear.

### Evaluate the candidate

Before committing, create `EVALUATION.md` with this worker's candid, evidence-based assessment of the resulting candidate: what it accomplishes, what you actually checked, known gaps or risks, and the most useful next work when applicable.

A useful stepping stone may explicitly describe what remains unresolved while still retaining the form of a candidate solution.

## Action 2: Nominate the best existing solution

In this action, carefully compare existing candidate nodes and identify the current best candidate. Do not originate a new solution, repair a candidate, or synthesize candidates as part of the nomination.

### Scope one bounded nomination

After an initial full-graph inspection, privately scope the comparison needed to nominate the current best existing candidate by considering the nomination question, graph coverage, common evidence, success criterion, and abort condition. Do not emit a separate prose scope block. Continue the comparison through the `bash` tool. If nomination is your chosen action, publish that a critical graph comparison is underway before detailed comparison, without revealing a tentative nominee or candidate preference:

```bash
gs_inflight announce --stage nominating --summary "Critically comparing the full candidate graph; no tentative preference disclosed"
```

Keep later activity updates for this comparison at stage `nominating` so the dashboard can report how many nomination workers are currently in flight.

The full candidate graph is the candidate set. Full-graph coverage does not require equal time or exhaustive retesting of every node: use ancestry, diffs, existing evaluations, and evidence to identify redundant or dominated nodes, then spend deeper inspection and testing on serious contenders. Do not treat commit subjects or the live nomination tally as substitutes for reviewing the graph.

Use evidence that bears on comparable behavior. A candidate passing its own tests is useful evidence about that candidate, but pass counts from different candidate-authored suites are not directly comparable; inspect the tests and behavior they cover. Prior nomination records may contain useful evidence, but they are not candidates and their existence or popularity is not evidence that a candidate is correct. Make your own judgment.

At least one valid nomination is necessary to select a candidate rather than fall back to the repository root. Do not vote merely to change the count; if the evidence does not support a defensible nomination, abstain before creating a contribution branch.

Public `gs_inflight` updates from a nominating worker must not disclose a tentative nominee or candidate preference. Nomination judgments remain independent until ballot commits are published.

### Initialize and declare the nomination

Choose one existing candidate SHA and branch from that exact candidate:

```bash
gs_fork <nominated_candidate_sha> record-best-solution
```

This branch is only for the nomination rationale. Do not modify, repair, combine, or extend the candidate's substantive answer.

Declare the nomination before creating the rationale commit:

```bash
gs_nominate_best_solution <nominated_candidate_sha>
```

This records a pending ballot for scheduler validation. It does not acquire a lock, update an incumbent, or reveal a live winner. Other workers may nominate concurrently.

### Document the nomination

Create or update `BEST_SOLUTION.md` with:

- the nominated candidate's exact SHA;
- the candidates and evidence compared;
- an evidence-based account of why this candidate is currently best;
- known limitations or remaining uncertainty; and
- any prior nomination records whose comparison evidence materially informed this nomination.

The nomination-record branch must otherwise preserve the nominated candidate's substantive solution unchanged.

## Action 3: Abstain

Choose this if, after inspecting completed and in-flight work, no work would add immediate meaningful information or progress. In particular, abstain when every feasible contribution would merely duplicate work already complete or materially underway. State the substantive reason, then run:

```bash
gs_abstain --rationale "<why no available contribution would add value>"
```

Do not create an empty commit. An explicit `gs_abstain` does not consume a worker-budget slot; the scheduler may launch a replacement worker.

## Artifact authorship

For a graph contribution, `MY_THOUGHTS.md` and `EVALUATION.md` are worker-specific and contain one authorship entry for the current worker. For other human-readable collaboration artifacts you create or update, such as `BEST_SOLUTION.md`, experiment reports, or continuation notes, maintain an update history at the top of the file. Prepend one entry for your update without deleting earlier entries:

```text
Updated by: Worker <WORKER_ID> | Scoped commits: <sha1>,<sha2>,...
```

For a graph contribution, list its target commits. For a best-solution nomination, list the candidate set that materially informed the comparison. Do not insert this plain-text header into source code, executable scripts, generated files, or data formats that do not support it; their authorship remains available through Git history and commit provenance.

# Commit, publish any nomination, and exit

Make exactly one commit. For a graph-expansion commit, choose the most accurate descriptive type and use this format:

```text
[<brief description of the type of contribution>|informed_by=<sha1>,<sha2>,...] <summary>
```
Write the descriptive type as one or more lowercase words joined by hyphens.

Provenance is mandatory. The `informed_by=` list must contain verified SHA references for every target commit in the scoped contribution, even when there is only one source. Abbreviated SHAs are valid when they uniquely identify commits in the repository.

```bash
git add -A
git commit -m "[<description>|informed_by=<sha1>,<sha2>,...] <informative summary>"
```

For a best-solution nomination contribution, include the nominated candidate and every other candidate that materially informed the comparison in `informed_by=`:

```bash
git add BEST_SOLUTION.md
git commit -m "[best-solution-nomination|informed_by=<sha1>,<sha2>,...] <informative summary>"
```

The scheduler validates the pending declaration, exact nominated parent, provenance, unchanged candidate tree except for `BEST_SOLUTION.md`, and rationale. It then publishes one immutable ballot for this worker. The nomination commit is metadata and is not itself eligible to receive votes.

After the commit is detected, the harness ends the worker episode. Do not send a separate prose-only summary.
\end{lstlisting}

\subsection{Task Addenda}
\label{app:prompts-addenda}

The task addendum is appended to the global component, separated by a blank
line. The Loop addendum is identical to the RMT addendum, with ``RMT''
replaced by ``Loop'' in its heading and in ``the complete RMT workbench''.

\begin{lstlisting}[
  style=gsprompt,
  title={\textbf{ProgramBench addendum}}
]
# ProgramBench task-specific addendum

The goal is to create a new, original codebase whose compiled executable reproduces the behavior of the provided reference program. The behavior is to be inferred from the bundled documentation and by running and interacting with the execute-only reference. The final output must include a `compile.sh`. ProgramBench runs it and grades the resulting `./executable` against a large hidden test suite. Repository prose alone is not graded.

## Candidate-node invariant

Every graph-expansion commit is a candidate node. Its tree must contain a self-contained, executable `compile.sh` that produces `./executable` in the repository root. This applies to every type of graph-expansion contribution. Every candidate node must retain the structure of a runnable ProgramBench submission.

The compiled candidate must not copy, wrap, execute, link, or otherwise depend on the reference program at runtime. Generated executables must not be committed.

A best-solution nominator compares existing candidate nodes. Its nomination-record commit is ballot metadata rather than a candidate. It may add or update only `BEST_SOLUTION.md` and must otherwise preserve the nominated candidate's tree.

## Reference executable

The execute-only reference program is available as `./executable_orig`. You may run it and observe normal program behavior, including stdout, stderr, exit status, and timing. You MUST NOT read, copy, hash, disassemble, decompile, or otherwise inspect its bytes.

The reference is available only during development and is absent during grading. Every candidate and its `compile.sh` must therefore be fully self-contained.

## Worker inheritance and builds

A worker branch inherits the tracked tree of its selected Git parent, including source files, collaboration artifacts, and `compile.sh`. Built executables are untracked and are not inherited. Rebuild a candidate from its committed source when testing it.

Each worker receives a fresh `./executable_orig`. `./baseline_executable` may hold a candidate built from the worker's starting commit, while `./executable` is reserved for the candidate built from the worker's current tree. Both are untracked build artifacts.

When comparing several candidate nodes, build each one in a separate private temporary directory or worktree so their sources and build products do not contaminate one another.

## Investigation and validation

Differential testing is one available way to investigate and validate behavior. When using it, compare candidate and reference behavior on the same inputs, including stdout, stderr, exit status, timing when relevant, and exact formatting.

Bundled documentation and examples provide useful evidence but may not cover hidden behavior. Valid and invalid flags, malformed input, empty and large input, Unicode, boundary cases, and combinations of modes may expose discrepancies. Compare against inherited or alternative candidates as well when regression detection or candidate nomination requires it.

Claims in `MY_THOUGHTS.md`, `EVALUATION.md`, `BEST_SOLUTION.md`, and other collaboration artifacts should distinguish observed evidence from inference. Do not claim coverage that was not performed.
\end{lstlisting}

\begin{lstlisting}[
  style=gsprompt,
  literate={—}{{\textemdash}}1,
  title={\textbf{IMOProofBench addendum}}
]
# IMO ProofBench task contract

This task is an International Mathematical Olympiad problem. The final scored
artifact is `SOLUTION.md` from the selected candidate commit. A contribution
that claims to contain a complete answer must put the complete proof in that
file. Other proof files may be retained as supporting work, but they are not
the canonical final answer.

After the swarm finishes, an LLM judge will score the final answer on the 0-7
IMO ProofAutoGrader scale:

- 7 — Correct: complete, fully rigorous proof.
- 6 — Almost correct: sound core argument with only minor errors or small
  gaps. Missing proofs for major components are not eligible for 6.
- 1 — Partial progress: substantial progress matching the grading guidelines.
- 0 — Incorrect: no substantial progress, or fundamentally flawed.

A wrong answer presented as if it were correct scores 0.
\end{lstlisting}

\begin{lstlisting}[
  style=gsprompt,
  title={\textbf{RMT addendum}}
]
## RMT benchmark addendum

This is an empirical architecture-research task. Read TASK.md,
research_task.md, and cluster_management.md before allocating compute. Preserve
the complete RMT workbench in every graph-expansion candidate. Use protected
development evaluations only for serious candidates, record actual evidence,
and never invent metrics. A useful candidate may capture a tested negative
result or an improved experimental configuration. Durably record its
limitations and unresolved uncertainty in the repository so future workers can
correctly interpret and build on the evidence. Each graph-expansion candidate should contain a validatable `submission/final_model` bundle when it is ready for comparison.
While waiting for queued or long-running experiments to finish, you may use the available time for other useful work rather than idling. Build on previous workers' analysis, audits, tooling, validation, and priority-queueing work by running the actionable experiments they enable. When a concrete experiment is ready, favor obtaining new empirical evidence over adding another preparatory layer. Watch for search collapse: if recent workers are clustering around incremental variants of one mechanism or a narrow architectural family, reopen the search with a materially different, well-motivated architectural hypothesis and test it empirically. Novelty alone is not sufficient; the new direction should address a concrete limitation or unexplored opportunity in the accumulated evidence. Cancel work when separated observations give concrete evidence that it is stuck, invalid, or no longer useful; a normal queue wait or quiet compile is not sufficient evidence. Every submitted job must become terminal before this worker exits.
\end{lstlisting}

\subsection{Model- and Transport-Specific Additions}
\label{app:prompts-additions}

\paragraph{Sequential shell interaction for Gemini-3.1-Pro.}
Gemini-3.1-Pro workers receive the extension below at the end of their system
prompt. In pilot runs without it, Gemini workers often issued an entire
contribution workflow (inspection, scope announcement,
\texttt{gs\_fork}, file writes, and \texttt{git commit}) in a single
response, sometimes predicting command outputs and branching from a guessed
commit SHA. When \texttt{gs\_fork} then failed, the remaining commands
still ran and committed the new files on the worker's placeholder branch, and
the result was rejected as invalid. The extension asks workers to wait for
each command's output before issuing commands that depend on it, to run
\texttt{gs\_fork} as a standalone step, and to use only read-only
inspection and coordination commands before branching succeeds. For these
runs, the harness also runs each shell block with \texttt{set -e}, skips the
remaining blocks of a response once a block fails, and rejects any pre-branch
command other than read-only inspection, coordination, a standalone
\texttt{gs\_fork}, or a standalone \texttt{gs\_abstain}. These
additions constrain how and when shell commands run (for example, scripts and
tests can run only after branching); the graph, scope, dependency, and
nomination rules are unchanged.

\begin{lstlisting}[
  style=gsprompt,
  title={\textbf{Gemini-3.1-Pro sequential shell-interaction extension}}
]
# Sequential shell interaction

For this configuration, interact with the environment incrementally. This
section supersedes any earlier suggestion that an entire workflow should be
submitted as one command batch.

Use a response for one evidence-dependent step, then stop and inspect the
returned output before deciding the next step. Commands may share one response
only when they are independent of one another's output. If the arguments,
safety, or usefulness of a later command depend on whether an earlier command
succeeded or what it printed, put the later command in a subsequent response.

In particular:

- inspect the graph and in-flight work, then wait for that output before
  choosing an action;
- publish a contribution or nomination scope, then wait for confirmation;
- run `gs_fork` as its own state-changing step and wait for confirmation
  that the branch switch succeeded; the response containing `gs_fork`
  must contain exactly that one Bash block and no other command;
- before that confirmation, use only read-only repository inspection and the
  documented coordination tools: do not create, modify, move, or delete files
  anywhere, including temporary files, and do not run interpreters, scripts,
  binaries, tests, builds, or experiments;
- only after that confirmation, create or modify files or execute validation;
- inspect validation or test output before deciding whether to revise, commit,
  nominate, or abstain; and
- never combine `gs_fork` with file writes or `git commit` in the same
  response, never create candidate files before branching, and never mix the
  contribution and nomination workflows.

Interact with the environment frequently rather than predicting command
results and emitting the rest of the workflow in advance. The harness stops
the remainder of a response after the first failed shell command and rejects
any pre-branch command that is not read-only inspection, coordination, a
standalone `gs_fork`, or a standalone `gs_abstain`. Do not work around
those constraints: read the returned output and choose a valid next action.
\end{lstlisting}

\paragraph{Codex transport override.}
ProgramBench Codex workers receive the override below after the user prompt.
Their only task execution interface is a guarded MCP tool,
\texttt{pb.container\_bash}, which runs commands in the worker's private Git
worktree inside the task's network-disabled container.

\begin{lstlisting}[
  style=gsprompt,
  title={\textbf{Codex transport override (ProgramBench)}}
]
## Codex ProgramBench transport override

You are a Codex CLI worker. The ProgramBench task workspace is not present in
your host working directory. The only task execution interface is the
`pb.container_bash` MCP tool. Use that tool for every task operation, including
inspection, edits, builds, tests, Git commands, and every `gs_*` command.

Any instruction above to *return* a fenced Bash block describes the historical
direct-API transport and is superseded here: call `pb.container_bash` instead.
Preserve every other Git-Swarm policy exactly. Finish one bounded
contribution or abstention. After a commit succeeds, or after `gs_abstain`
records an abstention, stop immediately. Do not merely describe commands or
include a proposed command transcript in your final answer.
\end{lstlisting}

\subsection{User Prompts}
\label{app:prompts-user}

The system prompt is accompanied by a task-specific user message. In the user
prompts below, values in angle brackets are filled by the harness for each
worker or task.

\begin{lstlisting}[
  style=gsprompt,
  title={\textbf{ProgramBench user prompt}}
]
Given only a compiled binary and its documentation, the collective goal is to architect and implement a complete codebase that reproduces the original program's behavior.

The execute-only reference binary is at `./executable_orig` in the root of your current Git worktree. Documentation is bundled alongside it. No project-specific dependencies are pre-installed. There is no internet access.

Infer that behavior from the bundled documentation and by running and interacting with the provided binary.

Finding existing source code, installing the original tool, wrapping the reference binary, or depending on it at runtime does not count as a solution and is disqualifying.

Each graph-expansion node in the shared Git graph must retain the structure of a candidate submission. Its committed `compile.sh` must produce `./executable` in the repository root, even when the node's main contribution is exploratory, evaluative, or intended as a stepping stone for later workers.

The reference executable may be execute-only. If reading or copying it fails with `Permission denied`, that is expected. Run it normally and observe its CLI, stdin, stdout, stderr, exit behavior, and timing instead.

Differential checks are investigative and validation tools, not the deliverable.

---

## Git-Swarm worker context

- Worker ID: `<WORKER_ID>`
- Working directory: `<WORKTREE_PATH>`
- Initial branch: `worker/<WORKER_ID>` at repository root `<ROOT_SHA>`

## Swarm status at spawn

- Total worker budget: `<TOTAL_WORKER_BUDGET>`
- Remaining worker budget: `<REMAINING_WORKER_BUDGET>`
- Workers currently in flight: `<INFLIGHT_WORKER_COUNT>`
- Valid nominations cast: `<BEST_SOLUTION_NOMINATIONS_CAST>`
- Distinct candidates nominated: `<BEST_SOLUTION_CANDIDATES_NOMINATED>`
- Current nomination tally (candidate SHA and vote count):
<BEST_SOLUTION_NOMINATION_TALLY>

This is a spawn-time snapshot. Run `gs_dashboard` for a current snapshot. Nomination counts are coordination state, not evidence of candidate quality. The visible nominees are not a shortlist: candidates with zero nominations remain fully eligible and must still be considered.
\end{lstlisting}

\begin{lstlisting}[
  style=gsprompt,
  title={\textbf{IMOProofBench user-prompt template}}
]
<PROBLEM_STATEMENT>

Write a complete, rigorous mathematical proof. Your proof will be graded on a 0-7 scale where 7 is a flawless IMO-standard proof.

Requirements:
- Every claim must be explicitly justified. No hand-waving.
- Each step must follow directly from the previous step or a cited theorem.
- If your argument splits into cases, handle every case.
- Organize your proof clearly: key observations, main argument, conclusion.
\end{lstlisting}

\end{document}

%% file: math_commands.tex
\usepackage{amsmath,amsfonts,bm}

\def\eqref#1{equation~\ref{#1}}

\def\1{\bm{1}}

\DeclareMathAlphabet{\mathsfit}{\encodingdefault}{\sfdefault}{m}{sl}
\SetMathAlphabet{\mathsfit}{bold}{\encodingdefault}{\sfdefault}{bx}{n}

